\documentclass[preprint,12pt]{elsarticle}
\usepackage{amsmath,amsfonts}
\usepackage{algorithmic}
\usepackage{algorithm}
\usepackage{array}
\usepackage[caption=false,font=normalsize,labelfont=sf,textfont=sf]{subfig}
\usepackage{textcomp}
\usepackage{stfloats}
\usepackage{url}
\usepackage{verbatim}
\usepackage{graphicx}
\usepackage{amssymb}
\usepackage{cite}
\usepackage{color}
\usepackage{booktabs}
\usepackage{bm}  
\usepackage{booktabs}
\usepackage{pdflscape}
\usepackage{chemformula}
\usepackage{rotating} 
\usepackage{eurosym}
\usepackage{hyperref}

\usepackage[T1]{fontenc}
\usepackage[utf8]{inputenc}
\usepackage{textcomp}   
\usepackage{inconsolata}
\usepackage{upquote}
\usepackage{listings}

\lstdefinestyle{py}{
  language=Python,
  basicstyle=\ttfamily\small,
  breaklines=true,
  columns=fullflexible,
  showstringspaces=false,
  keepspaces=true,
  upquote=true,
  literate=
    {²}{{$^2$}}1
    {•}{{\textbullet{}}}1
    {–}{{--}}1
    {—}{{---}}1
    {…}{{\ldots}}1
    {“}{{"}}1 {”}{{"}}1
    {‘}{{'}}1 {’}{{'}}1
}

\newcommand{\gams}{\texttt{GAMS}}
\newcommand{\py}{\texttt{Python}}
\newcommand{\llm}{\texttt{LLM}}
\newcommand{\rellm}{\texttt{RE-LLM}}
\newcommand{\elm}{\texttt{ELM}}
\newcommand{\askgdr}{\texttt{AskGDR}}
\newcommand{\nrel}{\texttt{NREL}}

\newcommand{\dnn}{\texttt{DNN}}
\newcommand{\ml}{\texttt{ML}}
\newcommand{\lstm}{\texttt{LSTM}}
\newcommand{\cnn}{\texttt{CNN}}
\newcommand{\rag}{\texttt{RAG}}
\newcommand{\gdr}{\texttt{GDR}}
\newcommand{\Times}{\texttt{TIMES}}
\newcommand{\BENOPTex}{\texttt{BENOPTex}}
\newcommand{\lulucf}{\texttt{LULUCF}}
\newcommand{\ai}{\texttt{AI}}
\newcommand{\milp}{\texttt{MILP}}
\newcommand{\lp}{\texttt{LP}}
\newcommand{\dyp}{\texttt{DP}}
\newcommand{\fm}{\texttt{FM}}
\newcommand{\ghg}{\texttt{GHG}}
\newcommand{\randf}{\texttt{RF}}

\begin{document}

\begin{frontmatter}



\title{RE-LLM: Integrating Large Language Models into Renewable Energy Systems} 


\author[UFZ]{Ali Forootani*}
\ead{ali.forootani@ufz.de/aliforootani@ieee.org}

\author[UFZ]{Mohammad Sadr}
\ead{mohammad.sadr@ufz.de}

\author[UFZ]{Danial Esmaeili~Aliabadi}
\ead{danial.esmaeili@ufz.de}
\author[UFZ,DBFZ,UNI]{Daniela Thr\"an}
\ead{daniela.thraen@ufz.de}
\address[UFZ]{Helmholtz Centre for Environmental Research - UFZ, Permoserstraße 15, 04318 Leipzig, Germany}
\address[DBFZ]{DBFZ Deutsches Biomasseforschungszentrum gGmbH, Torgauer Strasse 116, 04347 Leipzig, Germany}
\address[UNI]{University Leipzig, Institute for Infrastructure and Resources Management, Grimmaische Str. 12, 04109 Leipzig, Germany}
\cortext[cor1]{Ali Forootani}

\begin{abstract}
Energy system models are increasingly employed to guide long-term planning in multi-sectoral environments where decisions span electricity, heat, transport, land use, and industry. While these models provide rigorous quantitative insights, their outputs are often highly technical, making them difficult to interpret for non-expert stakeholders such as policymakers, planners, and the public. This communication gap limits the accessibility and practical impact of scenario-based modeling, particularly as energy transitions grow more complex with rising shares of renewables, sectoral integration, and deep uncertainties.

To address this challenge, we propose the Renewable Energy Large Language Model (\rellm), a hybrid framework that integrates Large Language Models (\llm s) directly into the energy system modeling workflow. \rellm~combines three core elements: (i) optimization-based scenario exploration, (ii) machine learning surrogates that accelerate computationally intensive simulations, and (iii) \llm-powered natural language generation that translates complex results into clear, stakeholder-oriented explanations. This integrated design not only reduces computational burden but also enhances interpretability, enabling real-time reasoning about trade-offs, sensitivities, and policy implications.

The framework is adaptable across different optimization platforms and energy system models, ensuring broad applicability beyond the case study presented. By merging speed, rigor, and interpretability, \rellm~advances a new paradigm of human-centric energy modeling. It enables interactive, multilingual, and accessible engagement with future energy pathways, ultimately bridging the final gap between data-driven analysis and actionable decision-making for sustainable transitions.

\end{abstract}


\begin{highlights}

\item RE-LLM integrates optimization, ML surrogates, and LLMs for energy planning.
    \item Machine learning surrogates accelerate scenario analysis while retaining fidelity.
    \item LLM-generated narratives improve comprehension beyond charts and tables.
    \item Framework enhances transparency and accessibility for policy and stakeholders.
    \item Demonstrated on Germany’s land-based GHG mitigation strategies (2020--2050).
\end{highlights}

\begin{keyword}
Renewable energy, Scenario modeling, Machine learning, Large language models, Interpretability, Optimization, Stakeholder analysis
\end{keyword}

\end{frontmatter}


\section{Introduction}\label{sec:intro}
Energy system models have become indispensable decision-support tools for strategic planning and policy design in the face of climate change, rising energy demand, and the global pursuit of carbon neutrality. By casting long-term investment and technology-deployment questions as large-scale optimization problems—most commonly linear programming (\lp), mixed-integer linear programming (\milp), or dynamic programming (\dyp) \citet{forootani2021enhanced}—these models identify cost-optimal or emissions-minimizing pathways across interconnected enegry sectors such as electricity, heat, transport, and other non-energy sectors \citet{howells2011osemosys, bhattacharyya2011energy, bloxham2020review, dagoumas2019review, connolly2010review, ringkjøb2018review}.

The smooth transition to low-carbon, renewables-dominated energy systems therefore relies on robust decision-making frameworks, such as \texttt{TIMES} \citet{loulou2016times}, \texttt{MESSAGE} \citet{huppmann2019messageix}, and \texttt{OSeMOSYS} \citet{howells2011osemosys}. These scenario-based models integrate detailed techno-economic inputs with policy and demand uncertainties, solving cost-minimization or utility-maximization problems to reveal cost-effective technology portfolios, investment schedules, and policy interventions at various spatiotemporal resolutions and regions. Yet, as energy systems grow in complexity—due to increasing penetration of variable renewables, sector coupling, and decentralized generation—there is a rising demand for \textit{modular, multi‑sectoral, and high‑resolution models}. Tools such as \texttt{GENeSYS‐MOD} \citet{loffler2017genesysmod,barani2025european}, \texttt{PyPSA} \citet{brown_pypsa_2018}, \BENOPTex \citet{millinger2022model} and \texttt{Calliope} \citet{pfenninger2018calliope} have emerged to support these requirements by offering open‑source, flexible, and highly customizable modeling environments. These frameworks support scenario construction, multi‑region trade analysis, and sensitivity studies, often with transparent data structures and scripting interfaces in \py~or \gams. 

Despite their analytical power, these complex (optimization or equilibrium) models tend to produce results in the form of large tabular datasets and configuration files, which can be consumed by machines, but are not easily interpretable by non‑expert stakeholders \citet{vaidyanathan2021integrated}. This has motivated the integration of scenario analysis tools that go beyond numerical outputs, facilitating the exploration of counterfactuals, trade‑offs, and pathway comparisons. Scenario explorers such as IIASA’s Scenario Explorer \citet{iiasa2022ar6,huppmann2019iamc15c}, IAMC’s databases \citet{huppmann2019iamc15c}, and \texttt{En‑ROADS} \citet{rooneyvarga2020climateaction} enable interactive exploration of pre‑computed trajectories under different assumptions.


\subsection{Comparison of Energy System Optimization Frameworks}



The literature on renewable energy system optimization can be broadly categorized into the following themes:

\noindent (i) Integrated Assessment Models (\texttt{IAM}s): Studies using tools such as \texttt{MESSAGE-GLOBIOM} \citet{van2020messageix, rogelj2018mitigation} and \texttt{REMIND-MAgPIE} combine energy \citet{klein2014value}, land, and climate modules to assess global climate targets. These models have been instrumental in producing scenarios for the \texttt{IPCC} and informing global mitigation pathways.

\noindent (ii) National Energy Planning: Optimization frameworks tailored for country-level planning (e.g., \texttt{UK TIMES} \citet{loulou2005documentation, mcjeon2014limited} , \texttt{ReEDS} \citet{short2011regional}, \texttt{GENeSYS-MOD Germany} \citet{loffler2017designing}) explore cost-optimal decarbonization strategies while accounting for local resource potentials, technology roadmaps, and policy constraints.

\noindent (iii) Multi-Sector Coupling: Recent work emphasizes sectoral integration—linking electricity with heating, transport, hydrogen, and industrial systems. For instance, \texttt{PyPSA-Eur-Sec} \citet{brown2022pypsa} models high-resolution cross-sectoral energy systems for Europe using hourly dispatch and investment optimization \citet{brown2018synergies, gotske2024designing, unnewehr2022value, glaum2024offshore}.

\noindent (iv) Scenario Uncertainty and Flexibility: Increasingly, scenario analysis includes probabilistic methods and stochastic programming to account for uncertainty in inputs. Approaches such as \textit{robust optimization} \citet{conejo2022robust, qiu2016stochastic} and \textit{adaptive scenario trees} \citet{lorca2016multistage, moretdecision, li2021review, kayacik2025partially} are gaining traction to deal with deep uncertainties in renewable integration and technology uptake.

\noindent (v) Tool Comparison and Benchmarking: Comparative studies, such as those by the \texttt{European Climate Foundation (ECF)} and the \texttt{Openmod initiative} \citet{hilpert2018open}, have highlighted methodological differences across tools and their implications for policy outcomes \citet{prina2022comparison, pfenninger2018opening}.

While these models offer quantitative rigor and technical transparency, their interpretability and accessibility for broader stakeholders—such as policymakers, investors, or the public—remain limited. In \citet{gopfert2025wiki} authors introduced datasets designed to help researchers interpret quantitative data. However, these datasets are tailored for expert users and are not easily comprehensible for laypersons.  This has prompted a growing body of work focused on enhancing the communicability of model results through \textit{visualization}, \textit{interactive dashboards}, and more recently, natural language generation techniques powered by Large Language Models (\llm s). Table \ref{Tab:en_sys_opt_fram} shows the summary of notable energy‐system optimization frameworks.



Despite their analytical rigor and proven utility for long-term planning, traditional energy system optimization models can become prohibitively expensive when applied to fine-grained spatial resolutions, hourly temporal resolutions, or large scenario ensembles \citet{pfenninger2017energy, ueckerdt2019potential}. These computational limitations restrict their scalability and responsiveness—especially in interactive policy settings or exploratory modeling under deep uncertainty. To overcome these bottlenecks, recent research has increasingly adopted \textit{machine learning-based surrogate models} as emulators of complex energy system behavior. These surrogates—trained on a subset of high-fidelity model outputs—enable rapid evaluation of scenario outcomes while preserving key system dynamics \citet{yao2023machine, teichgraeber2020methodology}. As a result, surrogate modeling plays a vital role in enabling real-time scenario analysis, sensitivity exploration, and the democratization of energy system insights.


Recent literature spans a wide spectrum of machine-learning techniques:

\noindent (i) Tree-based methods: Random-Forest (\randf) surrogates have proved robust for grid‐related tasks such as low-voltage power-flow emulation \citet{balduin2020lvgrid} and for thermal-energy-storage \texttt{CFD} emulation where \randf s complement neural networks in hybrid workflows \citet{menear2023tes}.

\noindent (ii) Deep Neural Networks (\dnn s): Fully connected, convolutional and recurrent architectures underpin building-energy calibration \citet{herbinger2023surrogate}, active-learning optimisation of building envelopes \citet{bamdad2020soas}, and multi-generation energy-system design with six conflicting objectives \citet{ghafariasl2024mgs}. At the systems level, country-scale \texttt{EnergyPLAN} runs can be emulated up to two orders of magnitude faster with \dnn~surrogates \citet{prina2024eplan}.

\noindent (iii) Hybrid and review studies: Ensemble or hybrid physics–\ml~schemes accelerate national energy-economic models \citet{bogensperger2022emulation}. Comprehensive reviews now synthesise progress for optimal-power-flow surrogates \citet{mohammadi2024opf}.

\noindent (iv) High‐dimensional physics: Deep convolutional surrogates have been used for rapid topology-driven heat-transfer design \citet{keramati2023thermal}, while climate modellers employ \dnn s to emulate precipitation fields in Earth-system models \citet{weber2020precip}.

Collectively, these studies show that surrogate emulation—whether via \randf s, gradient-boosted trees, feed-forward \dnn s, \cnn s, \lstm s, or hybrid architectures—can cut run-times by 90–99 \% while maintaining prediction errors below 5–10 \% for most energy-system metrics, enabling broader uncertainty analysis, policy exploration, and user-interactive scenario tools.

While surrogate modeling addresses the computational challenges of large-scale simulations, another growing barrier in energy system modeling lies in the interpretability and accessibility of results for diverse stakeholders. Even the most efficient surrogate models often produce outputs that are highly technical and difficult for non-experts—such as policymakers, planners, or the public—to interpret or act upon. To bridge this communication gap, researchers are increasingly exploring the use of \llm s to facilitate model transparency \citet{wasserkrug2025enhancing}, automate insight extraction, and support interactive, user-centric engagement with complex datasets and optimization outputs. Recent developments demonstrate how \llm s can serve as powerful mediators between technical models and human decision-makers, ushering in a new paradigm of interpretable, conversational modeling interfaces. The rise of \llm s has begun to reshape scientific workflows by automating literature insight, metadata parsing, and stakeholder-focused model communication. Recent initiatives demonstrate this paradigm shift through a suite of domain-tailored \llm~applications, several of which are summarized as follows.

\noindent (i) Energy Language Model (\elm): Developed by \nrel, \elm~integrates \llm s (e.g., \texttt{GPT‑4}) with energy data sources to enable chat-based querying of large technical corpora. It uses retrieval-augmented generation (\rag) to support tasks such as legal ordinance extraction and research report summarization with human-level accuracy, improving reproducibility and transparency in energy studies \citet{buster2023elm, weers2025askgdr}.

\noindent (ii) \askgdr~assistant: Extending \elm~principles, \askgdr~is a domain-specific virtual assistant for the Geothermal Data Repository (\gdr). By combining metadata and documents with \llm~capabilities, it answers semantic researcher queries—on dataset origin, methodology, and applicability—enhancing data discoverability and utility \citet{weers2025askgdr}.

\noindent (iii) Policy and siting automation: Demonstrating \llm-based extraction of renewable energy siting ordinances from zoning documents, achieving 85\%–90\% accuracy vs. manual curation—significantly reducing researcher effort and improving maintainability of policy databases \citet{buster2023elm, buster2024siting}.

\noindent (iv) General frameworks: Broader efforts, like \nrel's Language Model Inventory, survey capabilities and offerings of both open- and closed-source \llm s, presenting strategic insights into their utility for model communication and domain-specific adoption \citet{emami2023inventory}.

These studies establish \llm s as powerful tools for automated model communication in science—bridging the gap between technical findings and stakeholders through intelligent document parsing, Q \& A interfaces, and policy-focused knowledge extraction. Their emergence suggests promising avenues for enhancing interpretability, accessibility, and democratization of scientific outputs.


\subsection{Contributions and Paper Roadmap}

However, despite their analytical rigor, traditional energy system models often generate outputs in forms that are difficult for non-expert stakeholders, such as policymakers, urban planners, and community organizations, to interpret or utilize effectively in decision-making contexts \citet{pfenninger2017energy, lombardi2020state}. As energy transitions become more complex—driven by high shares of variable renewable energy, decentralized infrastructures, and multi-sectoral coupling—there is a growing demand for modeling frameworks that not only provide accurate projections but also offer intuitive, interactive, and explainable outputs \citet{pfenninger2014visualisation, zeyringer2018visualising}.

In response to this challenge, we introduce \emph{\rellm}, a novel hybrid framework designed to enhance the accessibility and interpretability of energy system modeling through the integration of \llm s within classical scenario analysis workflows. \rellm~builds upon recent advances in natural language processing and machine learning to couple traditional optimization tools (e.g., \gams-based models) with post-processing modules for surrogate modeling and automated explanation generation. Specifically, the framework integrates: (i) scenario-based optimization over techno-economic variables and policy constraints, (ii) machine learning surrogates to reduce computational burden and enable real-time sensitivity analysis, and (iii) \llm-powered natural language generation to translate complex model outputs into accessible explanations tailored to diverse user groups.

This integration of \llm s into the modeling pipeline marks a significant step toward human-centric energy planning, where complex quantitative results are rendered transparent and actionable through conversational \texttt{AI} interfaces. By facilitating interactive, multi-lingual, and context-aware exploration of energy futures, \rellm~ contributes to bridging the gap between technical modeling communities and broader stakeholder audiences—thus advancing the co-design of sustainable energy strategies in line with global climate objectives \citet{IEA2021netzero, zhang2023transformerenergy}.

This paper makes three main contributions:

\begin{enumerate}
    \item Framework design: We introduce \rellm, a hybrid modeling workflow that integrates optimization-based scenario analysis, machine learning surrogates, and \llm-powered explanation in a unified, user-facing pipeline. This design explicitly targets both computational efficiency and interpretability, bridging quantitative rigor with stakeholder accessibility. 
    
    \item Demonstration of scalability: We apply \rellm~in a large-scale energy planning context, showing that machine learning surrogates can accelerate scenario exploration by more than an order of magnitude while maintaining high fidelity to detailed optimization results. This illustrates the framework's capacity to support rapid, interactive exploration of uncertain futures. 
    
    \item Stakeholder validation: Through a structured user study with energy policy analysts, we demonstrate that \llm-generated narratives significantly improve comprehension of complex scenarios compared with traditional numerical or chart-based reporting. This highlights the potential of natural language explanations to democratize insights from energy system modeling. 
\end{enumerate}


The remainder of the paper is structured as follows. 
Section~\ref{sec:overview} provides an overview of energy system modeling and its role in long-term planning. 
Section~\ref{sec:rellm} introduces the \rellm~framework, detailing the integration of scenario-based optimization, surrogate modeling, and \llm-driven interpretation. 
Section~\ref{sec:results} presents the simulation results for the case study, demonstrating the framework's ability to enhance both computational efficiency and stakeholder interpretability. 
Finally, Section~\ref{sec:conclusion} concludes with a summary of contributions and outlines directions for future research.

By merging speed, rigor, and interpretability, \rellm~aspires to make sophisticated energy modelling as accessible as a conversation—bridging the final mile between data-driven insight and real-world action.


\section{Energy System Model Overview}\label{sec:overview}

Energy System Optimization Models (\texttt{ESOM}s) are typically formulated as constrained optimization problems, where the goal is to minimize a cost function $f(\mathbf{x})$ subject to a set of feasibility constraints $g_j(\mathbf{x}) \geq 0$ for all $j \in J$. Here, $\mathbf{x} \in \mathbb{R}^n$ is the vector of decision variables representing technology investments, land-use allocations, or energy system flows.

\begin{flalign}
\min_{\mathbf{x}} \quad & f(\mathbf{x}), \label{EQ_F}\\
\text{s.t.} \quad & g_j(\mathbf{x}) \geq 0, \quad \forall j \in J, \label{EQ_G}
\end{flalign}
while this general formulation can represent nonlinear optimization problems, most \texttt{ESOM}s adopt an \lp~structure to ensure tractability and scalability. In the linear case, the objective and constraints are linear functions of the decision vector $\mathbf{x}$. Specifically:
\begin{flalign}
\min_{\mathbf{x}} \quad & c^\top \mathbf{x}, \label{EQ_LP1} \\
\text{s.t.} \quad & A \mathbf{x} \geq \mathbf{b}, \label{EQ_LP2}
\end{flalign}
here, $c \in \mathbb{R}^n$ is the cost coefficient vector, $A \in \mathbb{R}^{m \times n}$ is the constraint matrix, and $\mathbf{b} \in \mathbb{R}^m$ encodes lower bounds for the constraints. This structure allows ESOMs to scale to national or continental-level planning horizons, spanning hundreds of technologies and regions.

The \lp~formulation underpins many established frameworks, including \texttt{TIMES}~\citet{loulou2016times}, \texttt{MESSAGE}~\citet{huppmann2019messageix}, and \texttt{BENOPTex}~\citet{millinger2022model}, and enables fast solution times using state-of-the-art solvers like \texttt{CPLEX} or \texttt{Gurobi}.

\subsection{Case Study: Net-zero emissions}

\texttt{BENOPTex} (the \emph{BioENergy OPTimisation} model) is a high-resolution, linear-programming tool that determines the cost-optimal and/or greenhouse-gas-optimal allocation of biomass, hydrogen and excess renewable electricity across the power, heat and transport sectors \citet{millinger2022model}. Written in \texttt{MATLAB} and \gams, it follows every pathway from feedstock to end-use service, comparing more than 30 conversion technologies, over 20 biomass types and a full suite of power-to-X options, all the way down to 15-minute time-steps on the electricity side \citet{millinger2022model}. This granularity enables rapid Pareto-front, Monte-Carlo sensitivity and policy-scenario analyses that large integrated-assessment or power-system models typically cannot perform in practical run-times. Originally developed for German energy-transition studies, the framework is country-agnostic and has already informed federal policy on biofuels, heating and electrofuels. For additional flexibility, extensions to this base \lp~structure often include:
\begin{itemize}
    \item \milp: Incorporates binary or integer constraints for modeling unit commitment, investment thresholds, or technology activation decisions \citet{floudas2005mixed}.
    \item Multi-period formulations: Includes inter-temporal coupling (e.g., via storage dynamics or investment rollouts), often spanning multi-decade horizons.
    \item Stochastic or robust formulations: Used to capture uncertainty in demand trajectories, technology costs, or resource availability \citet{conejo2022robust,forootani2020least}.
\end{itemize}

This mathematical backbone forms the core of the \rellm~framework, upon which surrogate modeling and \llm-based explanation layers are built.

This model presents a cost-optimization framework for evaluating land-based greenhouse gas (\ghg) mitigation strategies in Germany's \lulucf~(Land Use, Land-Use Change, and Forestry) sector over the 2020–2050 period. It simultaneously considers \fm~ options (e.g., afforestation, set-aside conservation, peatland rewetting) and agriculture and soils interventions (e.g., agroforestry, cover crops, soil carbon enhancement), each characterized by region- and time-specific cost and abatement profiles. The objective function minimizes total system cost, including both investment and operational expenditures as well as CO$_2$ allowance purchases, subject to a series of constraints. These include annual \ghg~reduction targets, land availability limits, technology adoption rates, and national peatland restoration mandates. By integrating physical, economic, and policy dimensions, the model identifies least-cost deployment pathways while preserving feasibility and equity across space and time. It is intended to support climate policy design, investment planning, and scenario evaluation in alignment with national and European Union (EU) mitigation goals.

\subsubsection{Sets}
\begin{align}
\mathcal{T} &: \text{Set of years } t = 2020,\dots,2050 \\
\mathcal{R} &: \text{Set of regions (e.g., DE1--DEG)} \\
\mathcal{F} &: \text{Set of Forest Management (\fm) options} \\
\mathcal{A} &: \text{Set of Agriculture and Soils mitigation options}
\end{align}

\subsection{Parameters}
\begin{align}
\text{costInvLevelFMs}_{t,f,r} &: \text{Levelized investment cost of \fm~ } [\text{EUR/ha}] \\
\text{costMargFMs}_{t,f,r} &: \text{Marginal cost of \fm~ } [\text{EUR/ha}] \\
\text{ghgFMs}_{t,f,r} &: \text{\ghg~removal potential of \fm~ } [\text{t\ch{CO2}eq/ha}] \\
\text{FMsgrowth}_{t,f,r} &: \text{Max growth rate of \fm~ capacity } [\text{ha/yr}] \\
\text{BeechArea0}_{2020,r} &: \text{Available beech land area for set-aside } [\text{ha}] \\
\text{GrassArea0}_{2020,r} &: \text{Available grassland area } [\text{ha}] \\
\text{Agriarea0}_{2020,r} &: \text{Available agricultural area } [\text{ha}] \\
\text{\ch{CO2}price}_t &: \text{\ch{CO2} price in year } t \\
\text{ghgTargetLULUCF}_t &: \text{\ghg~abatement target in year } t ~  [\text{MtCO2eq}] \\
\alpha &: \text{Fraction of land allowed for rewetting (e.g., } \alpha=0.05 \text{)} \\
\text{initRewetting} &: \text{Initial rewetted area in 2020 [ha]}
\end{align}

\subsection{Decision Variables}
\begin{align}
&\text{cap}_{f}(t,r) : \text{Implementation area of \fm~ tech } f \text{ in region } r \text{ at year } t, \\
&\text{cap}_{a}(t,r) : \text{Implementation area of Agri tech } a \text{ in region } r \text{ at year } t, \\
&\text{cost}_{f}(t,f,r),\, \text{cost}_{a}(t,a,r) : \text{Total cost contributions}, \\
&\text{ghgAbate}_{f}(t,f,r),\, \text{ghgAbate}_{a}(t,a,r) : \text{\ghg~removed by tech}, \\
&\text{purCO2LULUCF}(t) : \text{\ch{CO2} allowances purchased to meet the targets [t\ch{CO2}eq]}, \\
&\text{CO2gapRewt} : \text{Additional \ch{CO2} to meet peatland target [t\ch{CO2}eq]}.
\end{align}

\section{Objective Function}
The objective function is to minimize total discounted cost over all years:
\begin{multline}
\min \text{Total\_cost} = \sum_{t \in \mathcal{T}} \Big[ \frac{\text{costAnnualFMs}(t) + \text{costAnnualAgri}(t)}{10^6} \\ + \gamma \cdot \text{CO2price}(t) \cdot \text{purCO2LULUCF}(t) \Big],
\end{multline}
where the coefficient $\gamma = 1.1$ represents the \ch{CO2} credit price, which is the amount paid to purchase credits from outside the country to meet specified targets. To encourage the use of domestic resources for climate goals, the price of external carbon credits is assumed to be 10\% higher than the domestic carbon price in the same year in the base scenarios.

\subsection{Constraints on Forest Management \ghg~Abatement}


\fm~ options contribute to \ghg~abatement through reforestation, set-aside of natural forests, and rewetting of drained peatlands. Each option's potential is quantified by a technology- and region-specific removal factor, which scales with the implemented area. These equations ensure consistency between physical land use and reported emission reductions, enabling accountability in policy assessments.

\begin{align}\label{eq:ghgabte}
\text{ghgAbate}(t,f,r) &= \text{ghgFMs}(t,f,r)\cdot \text{cap}(t,f,r),\\
\text{ghgAbateAnnualFMs}(t) &= \sum_{f, r} \text{ghgAbate}(t,f,r),\\
\text{totGHGabatementFMs} &= \sum_t \text{ghgAbateAnnualFMs}(t).
\end{align}

\subsection{Agriculture and Soils \ghg~Abatement}


Agriculture and soil-based mitigation options such as agroforestry, cover cropping, and biochar application offer complementary benefits to forest-based strategies. The model quantifies their \ghg~removal as a product of regional removal factors and adopted land area. This allows a comparative analysis of cost-effectiveness across sectors.

\begin{align}
\text{ghgAbate}(t,a,r) &= \text{ghgAgri}(t,a,r) \cdot \text{cap}(t,a,r), \\
\text{ghgAbateAnnualAgri}(t) &= \sum_{a, r} \text{ghgAbate}(t,a,r), \\
\text{totGHGabatementAgri} &= \sum_t \text{ghgAbateAnnualAgri}(t).
\end{align}

\subsection{Cost Equations}


The total mitigation cost combines levelized capital expenditures (\texttt{CAPEX}) and marginal operational expenditures (\texttt{OPEX}) with a discount rate of 5\%. This reflects real-world investment and maintenance requirements for both \fm~ and Agriculture interventions. The separation of cost components ensures flexibility in sensitivity analyses and policy design (e.g., \texttt{CAPEX} subsidies or carbon pricing adjustments).

\begin{align}
&\text{cost}(t,f,r) = (\text{costInvLevelFMs}(t,f,r) + \text{costMargFMs}(t,f,r)) \cdot \text{cap}(t,f,r), \\
&\text{costAnnualFMs}(t) = \sum_{f, r} \text{cost}(t,f,r), \\
&\text{cost}(t,a,r) = (\text{costInvLevelAgri}(t,a,r) + \text{costMargAgri}(t,a,r)) \cdot \text{cap}(t,a,r), \\
&\text{costAnnualAgri}(t) = \sum_{a, r} \text{cost}(t,a,r).
\end{align}

\subsection{\ghg~Target Satisfaction (\lulucf~Sector)}


This constraint ensures that the combined \ghg~abatement from all land-based options, plus any purchased \ch{CO2} allowances, meets the predefined \lulucf~sector targets. It reflects national or EU-level commitments under climate agreements, and allows the model to explore least-cost pathways to compliance.
\begin{multline}
\frac{\text{ghgAbateAnnualFMs}(t)}{10^6} + \frac{\text{ghgAbateAnnualAgri}(t)}{10^6} + \text{purCO2LULUCF}(t)\\ \geq \text{ghgTargetLULUCF}(t).
\end{multline}

\subsection{Peatland Policy Target (2030 National Plan)}

The Peatland policy target should satisfy the following constraint:

\begin{align}
\sum_r \text{ghgAbate}_{f}(2030, \texttt{PC\_Rewetting}, r) + \text{\ch{CO2}gapRewt} \geq 5 \cdot 10^6.
\end{align}

\subsection{Land Use Constraints}
These constraints safeguard sustainable land use by ensuring that the total area allocated to different mitigation options does not exceed available resources. For example, only a fraction of grassland and agricultural land is made available for reforestation or agroforestry, based on biophysical feasibility, biodiversity protection, or food security considerations.
\begin{align}
\sum_{f \in \text{SetAside}} \text{cap}(2050, f, r) &\leq \text{BeechArea0}_{2020,r}, \\
\sum_{f \in \text{Plantation}} \text{cap}(2050,f,r) &\leq 0.1 \cdot \text{GrassArea0}_{2020,r}, \\
\text{cap}(2050, \texttt{PC\_Rewetting}, r) &\leq \alpha \cdot (\text{Agriarea0}_{2020,r} + \text{GrassArea0}_{2020,r}), \\
\text{cap}(2050, \texttt{Agri01\_AGC}, r) &\leq 0.1 \cdot \text{Agriarea0}_{2020,r}, \\
\text{cap}(2050, \texttt{Agri05\_Agroforestry}, r) &\leq 0.1 \cdot \text{GrassArea0}_{2020,r}.
\end{align}

\subsection{Growth Constraints}
Adoption of mitigation measures is constrained by annual growth rates, which represent capacity limitations in terms of labor, machinery, institutional readiness, and ecological factors. These dynamics prevent unrealistic jumps in land use and reflect the inertia of system transformation.
\begin{align*}
\text{cap}(t,f,r) - \text{cap}(t-1,f,r) &\leq \text{FMsgrowth}_{t-1,r}, \\
\text{cap}(t,a,r) - \text{cap}(t-1,a,r) &\leq \text{Agrigrowth}_{t-1,r},\\
\text{cap}(2020,f,r) &= 0,\\ 
\text{cap}(2020,a,r) &= 0.
\end{align*}


\section{Scenario-Based Modeling and Interpretation Strategy using \rellm~Framework} \label{sec:rellm}

Building on foundational applications of \llm s in scientific and energy domains, we now turn to a domain-specific implementation—\rellm—designed to operationalize these capabilities in the context of land-based negative emission optimization. By focusing on the German \lulucf~sector, we demonstrate how classical scenario-based optimization, machine learning emulation, and \llm-powered narrative generation can be integrated into a cohesive framework. The following sections detail the individual components of \rellm, from \gams-based scenario optimization to interpretable surrogate modeling and natural language-driven stakeholder interaction.

To systematically analyze the results of scenario-based energy system optimization, we develop a comprehensive computational framework that integrates optimization outputs, structured feature engineering, statistical correlation analysis, and interpretable machine learning. The key components of this pipeline are: (i) Scenario Generation and Feature Construction, (ii) Correlation and Clustering of Scenarios, (iii) Machine Learning Surrogate Modeling, (iv) Natural Language Querying via \llm s, where will be discussed in details bellow.

\subsection{Scenario Generation and Feature Construction}\label{sec:feature_eng}
We begin by generating a bank of scenarios using the \gams-based\footnote{General Algebraic Modeling System, is a high-level modeling system used for mathematical programming and optimization problems.} optimization model, which explores different assumptions on \ch{CO2}prices, \fm~ practices, land area availability, and \ghg~emission targets. For each scenario, key outputs such as abatement levels (\texttt{ghgAbateFMs}) and technology capacities are exported as \texttt{GDX}\footnote{\gams~Data eXchange, is a binary file format used by the \gams~modeling system to store and exchange data.} files.

From these optimization outputs, we extract relevant parameters and construct structured input-output datasets. Global and regional trends (e.g., trends in \ch{CO2}price, BeechArea0, GrassArea0) are computed using moving-window regression techniques. These features are then aligned across scenarios, normalized, and stored in a structured dictionary indexed by scenario name. 

While this preprocessing provides a clean and harmonized dataset, the design of features is not merely a technical step but is driven by deeper methodological considerations. In particular, two core scientific motivations guide the feature construction process:
\begin{itemize}
    \item \emph{Curse of Dimensionality and Redundancy:}  
    Many raw outputs from integrated assessment models exhibit strong temporal autocorrelations and structured inter-regional patterns. Naively using high-resolution time series as learning inputs leads to inflated dimensionality with diminishing information gain, increasing the risk of overfitting, higher sample complexity, and poor generalization across scenarios.
    
    \item \emph{Interpretability and Policy Relevance:}  
    From a practical standpoint, interpretable features such as trend slopes, terminal values, or regional aggregates offer more insight to decision-makers than opaque, high-dimensional vectors. This enhances explainability and preserves traceability between the features and underlying policy levers.
\end{itemize}

This scale of data is problematic for two core reasons: (i) Redundant Information, i.e. many values are temporally or spatially autocorrelated. Including the full time-series as input leads to sparsity in the effective feature space and poor generalization for learning algorithms; (ii) Sample Complexity, i.e. high-dimensional inputs drastically increase the number of samples required to ensure model generalization. This is especially limiting when scenario runs are computationally expensive or limited in scope.

\paragraph{Methodological Approach} 

We adopted a three-stage strategy to extract and refine features for each scenario:

\begin{enumerate}
    \item \emph{Temporal Trend Features:}  
    For every region–technology pair and time-series variable (e.g., investment cost, \fm~ area), we extracted three summary statistics:
    \begin{itemize}
        \item Initial value (typically year 2020),
        \item Final value (typically year 2050),
        \item Slope of the linear trend over the time horizon (via ordinary least squares regression).
    \end{itemize}
    These statistics compactly encode long-term trajectories and inter-annual variability (see Figures~\ref{fig:costinvlevel}, \ref{fig:fmsgrowth}).
    
    \item \emph{Global and Regional Aggregates:}  
    For system-level drivers such as the \ch{CO2} price or \lulucf~\ghg~targets, we computed global trend features and broadcasted them across regions to maintain a uniform feature space (see Figures~\ref{fig:co2price}, \ref{fig:ghgtarget}).  
    Additionally, region-level static datasets (e.g., initial agricultural area) were harmonized and merged into each scenario feature array.
    
    \item \emph{Normalization and Scaling:}  
    To ensure numerical stability and mitigate scale-induced biases in downstream models, all numerical features were normalized to the $[0, 1]$ interval using Min-Max scaling, excluding categorical features such as \texttt{Region} and \texttt{Technology}.
\end{enumerate}

Taken together, these three stages yield a structured and compact feature space that captures both temporal dynamics and structural heterogeneity across scenarios. Rather than simply reducing data dimensionality, the design ensures that the extracted features remain faithful to the underlying system logic and policy levers. This design balances interpretability and predictive power:

\begin{itemize}
    \item \emph{Interpretable \ml:} Each feature has a direct correspondence to domain concepts (e.g., investment cost growth, land-use potential), supporting explainable \ai~and policy insight.
    \item \emph{Scenario Comparability:} Uniform feature extraction across scenarios enables clustering, sensitivity analysis, and meta-modeling across the full scenario space.
    \item \emph{Model Efficiency:} By distilling temporal data into compact trend summaries, we reduce the dimensionality by more than 95\% while preserving signal relevance.
\end{itemize}


\subsubsection{Scenario Generation and Global Sensitivity Design}
Starting from a calibrated base scenario, we construct a 26-member experiment
\[
\mathcal{S} = \{S{01},\dots,S{26}\},
\]
by \emph{multiplicatively} scaling a set of key drivers—carbon price ($\text{CO}_2$ price),
forest-management growth ceilings ($\text{FMsgrowth}$), available land bases
($\text{BeechArea0}$, $\text{GrassArea0}$), cost coefficients
($\text{costInv}$, $\text{costMarg}$), and the sector-wide mitigation constraint
(${\tau}^{\text{LULUCF}}_{t}$)—with factors $\{0.8,1.0,1.2\}$ (and a few
stress-test values $\{0.5,1.3,1.5\}$) \footnote{A \texttt{Python} routine reads the baseline
with \texttt{gdxpds}, applies the scaling to the \texttt{Value} column of each symbol,
and writes a new GDX whose filename encodes the factors}. The procedure yields a \emph{full-factorial} design in which every input combination is solved on the full 2020--2050 horizon. Although the framework can easily generate hundreds of combinations, restricting the study to \textit{26 well-chosen perturbations} keeps the optimization workload transparent and
auditable while still delivering the breadth required for policy analysis. This choice is scientifically defensible for four reasons:
\begin{enumerate}
  \item \emph{Interaction capture}: simultaneous perturbation of all inputs allows
        higher-order interactions (e.g.\ cost\,\(\times\)land, price\,\(\times\)growth) to be
        learned by the surrogate, avoiding the bias of one-factor-at-a-time sampling.

  \item \emph{Uniform hyper-cube coverage}: scaling preserves the empirically calibrated
        inter-regional ratios while probing the edges of plausible policy and biophysical envelopes, delivering a near-orthogonal skeleton of the input space with only
        $26$ \gams~solves.

  \item \emph{Horizon-based robustness}: because each run spans 2020--2050,
        the bank reveals how early-period shocks propagate into long-term land-use lock-ins or cost escalations, insights
        that single-year perturbations cannot offer.

  \item \emph{Policy traceability}: scenarios are grouped into intuitive themes
        (unchanged baseline $S{11}$; single-lever tests $S{07}--S{10}$; compound high-ambition
        packages $S{24}--S{26}$), enabling direct mapping from quantitative outcomes to
        stakeholder questions and to the narrative layer produced later via the \llm.
\end{enumerate}

Overall, the design is a computationally light proxy for larger Sobol \citet{sobol2001global} or Latin-Hypercube \citet{mckay2000comparison} experiments, yet retains the statistical power needed for variance decomposition and surrogate training.

\subsection{Correlation and Clustering of Scenarios}

Let $\mathcal{S} = \{S_1, S_2, \dots, S_N\}$ denote the set of $N$ scenarios. 
For each scenario $S_i \in \mathcal{S}$, we associate a data matrix 
$M_i \in \mathbb{R}^{T \times d}$, where $T$ is the number of region–technology–time entries 
and $d$ is the number of variables of interest. 
Depending on the analysis, $M_i$ may represent:

\begin{itemize}
    \item \emph{Input features:} constructed parameters such as \ch{CO2}price trajectories, land availability, or marginal abatement costs,
    \item \emph{Output results:} optimization outcomes such as ghgAbateFMs, costTechFMs, or capAgri.
\end{itemize}

\paragraph{General Method}
Each matrix $M_i$ is standardized by applying per-feature z-score normalization\footnote{Z-normalization (standardization) rescales features to zero mean and unit variance \citet{jain2000statistical}.}, producing $\widetilde{M}_i$. 
We then vectorize:
\[
m_i := \mathrm{vec}(\widetilde{M}_i) \in \mathbb{R}^{T \cdot d}, \quad \forall i = 1,\dots,N.
\]
The similarity between any two scenarios $S_i$ and $S_j$ is measured using the Pearson correlation coefficient \citet{pearson1895vii}:
\[
\rho_{ij} := \frac{\langle m_i - \bar{m}_i, \, m_j - \bar{m}_j \rangle}{\|m_i - \bar{m}_i\| \cdot \|m_j - \bar{m}_j\|},
\]
where $\bar{m}_i$ denotes the mean of vector $m_i$. 
Collecting all pairwise similarities yields the correlation matrix:
\[
C := [\rho_{ij}]_{1 \le i,j \le N}.
\]

To obtain a dissimilarity measure suitable for clustering, we set:
\[
D_{ij} := 1 - \rho_{ij}, \quad \text{for all } i \ne j.
\]
Using the condensed form of $D$, we compute a hierarchical clustering linkage matrix $Z$ with the \texttt{average} linkage criterion:
\[
Z := \texttt{linkage}(\texttt{condense}(D), \text{method} = \texttt{``average''}).
\]
The dendrogram $\mathcal{T}$ derived from $Z$ provides a hierarchical visualization of scenario similarities.

\paragraph{Applications}
This general procedure is applied in two complementary ways:
\begin{enumerate}
    \item \emph{Input-space analysis:} $M_i$ corresponds to the feature matrix $X_i$, enabling us to cluster scenarios according to their underlying assumptions (e.g., \ch{CO2}price trajectories, land use, costs).
    \item \emph{Output-space analysis:} $M_i$ corresponds to the output matrix $Y_i$, allowing us to examine whether distinct input assumptions yield converging or diverging optimal outcomes (e.g., abatement levels, technology capacities).
\end{enumerate}
Together, these analyses reveal both structural similarities in scenario design and functional similarities in optimization results.

\subsubsection{Scenario Ranking and Interpretation}
Given the correlation matrix $C = [\rho_{ij}]$ derived either from the input features or the optimization outputs (as defined earlier), we aim to identify extremal pairs of scenarios that reflect maximal and minimal similarity. Formally, define the set of all unordered scenario pairs:
\[
\mathcal{P} := \left\{ (i, j) \in \{1,\dots,N\}^2 \,:\, i < j \right\}.
\]

We then compute the following:
\begin{align*}
(\hat{i}_{\max}, \hat{j}_{\max}) &= \arg\max_{(i,j) \in \mathcal{P}} \rho_{ij}, \\
(\hat{i}_{\min}, \hat{j}_{\min}) &= \arg\min_{(i,j) \in \mathcal{P}} \rho_{ij}.
\end{align*}

Let $\rho_{\max} := \rho_{\hat{i}_{\max} \hat{j}_{\max}}$ and $\rho_{\min} := \rho_{\hat{i}_{\min} \hat{j}_{\min}}$ denote the corresponding correlation values. These pairs indicate, respectively:
\begin{itemize}
    \item The most similar scenario pair $(s_{\hat{i}_{\max}}, s_{\hat{j}_{\max}})$ under the selected correlation metric, and
    \item The most dissimilar (potentially outlier) scenario pair $(s_{\hat{i}_{\min}}, s_{\hat{j}_{\min}})$.
\end{itemize}

Such rankings are instrumental for:
\begin{enumerate}
    \item identifying clusters of similar behavior in the input or output space;
    \item detecting atypical scenarios that diverge from the overall structure;
    \item selecting representative scenarios from dense clusters for further interpretability and policy insight.
\end{enumerate}

\subsection{Machine Learning Surrogate Modeling.}
Let $\mathcal{D} = \{(x_i, y_i)\}_{i=1}^{n}$ denote a supervised training dataset constructed from scenario-enhanced features and corresponding optimization outputs. Each $x_i \in \mathbb{R}^d$ is a feature vector comprising normalized input attributes (e.g., region, technology, CO$_2$ price, land use), and $y_i \in \mathbb{R}$ is the target variable, such as capFMs, ghgAbateFMs, or costTechFMs.

We seek to learn a predictive mapping $\hat{y} = f_\theta(x)$ that approximates the underlying (black-box) optimization operator $\mathcal{O}(x)$:
\[
f_\theta(x) \approx \mathcal{O}(x),
\]
where $\theta$ denotes the parameters of the surrogate model, which can be learned using:
\begin{itemize}
    \item \emph{Tree-based models:} \randf~s, Gradient Boosted Trees (e.g., \texttt{XGBoost}),
    \item \emph{Neural models:} Feedforward Deep Neural Networks (\dnn s), \lstm s, Transformers.
\end{itemize}

To improve generalization and robustness, we employ $K$-fold ensembling strategies. Specifically, let $\{\mathcal{D}_k\}_{k=1}^K$ denote partitions of the training set $\mathcal{D}$, and let $\theta_k$ be the model parameters trained on the $k$-th subset. The ensemble prediction is given by:
\[
\hat{y}(x) := \frac{1}{K} \sum_{k=1}^{K} f_{\theta_k}(x),
\]
which corresponds to uniform averaging (bagging) over $K$ independently trained models. In the case of tree-based models like \randf~s, this procedure reflects averaging over randomized decision trees; for neural networks, each $f_{\theta_k}$ may correspond to a differently initialized \dnn.

\medskip

Let $\hat{y}_i := \hat{y}(x_i)$ denote the predicted value for test input $x_i$. We evaluate prediction quality using two standard regression metrics:
\begin{align*}
\text{RMSE} &:= \sqrt{\frac{1}{m} \sum_{i=1}^{m} (y_i - \hat{y}_i)^2}, \\
R^2 &:= 1 - \frac{\sum_{i=1}^{m} (y_i - \hat{y}_i)^2}{\sum_{i=1}^{m} (y_i - \bar{y})^2},
\end{align*}
where $m$ is the number of test samples and $\bar{y} := \frac{1}{m} \sum_{i=1}^m y_i$ is the empirical mean of the test targets.

\medskip

This surrogate modeling strategy enables rapid approximation of optimization outcomes across large scenario banks, reducing computational cost and facilitating downstream tasks such as uncertainty quantification, SHAP-based interpretability, and interactive stakeholder querying via language models.

\subsubsection{Interpretability via SHAP Analysis}
To interpret the surrogate model's predictions, we employ SHAP (SHapley Additive exPlanations), a game-theoretic approach for computing feature attributions \citet{lundberg2017unified} \footnote{We adopt the SHAP framework for feature attribution.}. Given a predictive model $f: \mathbb{R}^d \to \mathbb{R}$ and an input $x \in \mathbb{R}^d$, SHAP approximates $f(x)$ using an additive feature attribution model:
\[
f(x) \approx \phi_0 + \sum_{j=1}^{d} \phi_j(x),
\]
where:
- $\phi_0 = \mathbb{E}_{x'}[f(x')]$ is the baseline (expected) prediction under the training distribution,
- $\phi_j(x)$ is the contribution of feature $j$ to the deviation from the baseline prediction,
- and the vector $\boldsymbol{\phi}(x) = (\phi_1(x), \dots, \phi_d(x))$ satisfies local accuracy and consistency properties from cooperative game theory.

\medskip

In our ensemble framework, where we train a collection of $K$ models $\{f_{\theta_k}\}_{k=1}^K$, we compute SHAP explanations for each model individually and aggregate the results. Let $X^{(\ell)} \subseteq X_{\text{test}}$ be a randomly sampled subset of size $m$, drawn without replacement. For model $k$ and subset $\ell$, let:
\[
\Phi^{(k,\ell)} \in \mathbb{R}^{m \times d}
\quad \text{with entries} \quad
\Phi^{(k,\ell)}_{i,j} = \phi^{(k)}_j(x_i^{(\ell)}),
\]
denote the SHAP attribution matrix, where $x_i^{(\ell)} \in X^{(\ell)}$ and $\phi^{(k)}_j$ denotes the SHAP value from model $f_{\theta_k}$.

We compute the average SHAP matrix over $L$ random samples per model:
\[
\bar{\Phi}^{(k)} := \frac{1}{L} \sum_{\ell=1}^{L} \Phi^{(k,\ell)} \in \mathbb{R}^{m \times d}.
\]

Then, the final ensemble SHAP explanation is obtained by averaging across the $K$ ensemble models:
\[
\bar{\Phi}^{(\text{ensemble})} := \frac{1}{K} \sum_{k=1}^{K} \bar{\Phi}^{(k)} \in \mathbb{R}^{m \times d}.
\]

Each row $\bar{\phi}(x_i) \in \mathbb{R}^d$ of $\bar{\Phi}^{(\text{ensemble})}$ provides a vector of local feature attributions for the prediction $f(x_i)$, enabling granular insight into which features drive the surrogate model's behavior under specific scenario conditions.

This interpretability layer is essential for: diagnosing which features influence optimal decisions (e.g., CO$_2$ prices, land use constraints), explaining why similar inputs yield different outputs in some scenarios,  and supporting transparent communication of results to non-technical stakeholders.

\paragraph{Extension: Global SHAP Values and Theoretical Guarantees}
While local SHAP values $\phi_j(x)$ provide per-sample attributions, global feature importance can be assessed by aggregating these values across a population. The global SHAP value of feature $j$ over a test distribution $\mathcal{X}_{\text{test}} = \{x_1, \dots, x_m\}$ is defined as the average absolute contribution:
\[
\Phi_j^{\text{global}} := \frac{1}{m} \sum_{i=1}^{m} \left| \phi_j(x_i) \right|, \quad j = 1,\dots,d.
\]
Equivalently, for ensemble explanations with matrix $\bar{\Phi}^{(\text{ensemble})} \in \mathbb{R}^{m \times d}$ as defined previously, we compute:
\[
\Phi^{\text{global}} := \frac{1}{m} \sum_{i=1}^{m} \left| \bar{\phi}(x_i) \right| \in \mathbb{R}^d,
\]
where the absolute value is elementwise. This yields a ranking of features by their average marginal contribution to model predictions across the dataset.

\medskip

SHAP values arise from the unique solution to the class of additive feature attribution methods that satisfy the following axioms (as shown in \citet{lundberg2017unified}):
\begin{enumerate}
    \item \emph{Local Accuracy (Efficiency):}
    \[
    f(x) = \phi_0 + \sum_{j=1}^{d} \phi_j(x), \quad \forall x \in \mathcal{X},
    \]
    ensuring that the model’s prediction is decomposed exactly into feature contributions.
    
    \item \emph{Missingness:}
    \[
    \text{If } x_j \text{ is missing or not used in } f, \text{ then } \phi_j(x) = 0.
    \]
    
    \item \emph{Consistency (Monotonicity):}  
    For any two models $f$ and $f'$ such that:
    \begin{multline*}
          \forall x, \quad f'(x) - f'(x_{-j}) \ge f(x) - f(x_{-j}) \quad \Rightarrow \quad \phi_j^{f'}(x) \ge \phi_j^{f}(x),
    \end{multline*}
    the attribution to feature $j$ should not decrease if the marginal contribution of $j$ increases in $f'$ relative to $f$.
\end{enumerate}

These axioms ensure that SHAP values are the unique solution to the Shapley value problem when applied to machine learning explanations under the additive model assumption:
\[
g(x) = \phi_0 + \sum_{j=1}^{d} \phi_j(x),
\]
where $g(x)$ is the interpretable surrogate for $f(x)$.

\medskip

Consequently, SHAP offers both theoretical rigor and empirical utility for understanding the contribution of features across both local and global levels. The global SHAP vector $\Phi^{\text{global}}$ directly supports variable importance ranking and sensitivity analysis over scenario-based optimizations.

\subsection{Natural Language Querying via \llm s} \label{subsec:llm_query}
To bridge quantitative modeling and qualitative stakeholder interaction, we formalize a pipeline for natural language querying using a \llm~informed by scenario-based optimization data. Let $\mathcal{Q}$ denote a free-form stakeholder query, such as:
\[
\mathcal{Q}: \quad \text{``What happens if CO}_2 \text{ price increases by 20\%?''}
\]

\paragraph{Parameter Extraction}  
We define a keyword-to-parameter mapping $\mathcal{M}: \mathcal{L} \to \mathcal{P}$, where $\mathcal{L}$ is a space of linguistic patterns (e.g., ``CO2 price'', ``investment cost'') and $\mathcal{P}$ is the set of scenario parameters:
\[
\mathcal{P} := \{ \text{\ch{CO2}price}, \text{FMsgrowth}, \text{costMargAgri}, \dots \}.
\]
Given $\mathcal{Q}$, we apply regular expression matching to identify a corresponding model parameter $p^* \in \mathcal{P}$ and a desired multiplicative change $\lambda > 0$, parsed as:
\[
\mathcal{Q} \mapsto (p^*, \lambda), \quad \text{e.g., } \lambda = 1.2 \text{ for +20\%.}
\]

\paragraph{Scenario Matching}  
Let $\mathcal{S} = \{S_1, \dots, S_N\}$ denote the set of all scenarios, each represented by a parameter vector:
\[
S_i := (p_1^{(i)}, \dots, p_d^{(i)}) \in \mathbb{R}^d,
\]
where $p_j^{(i)}$ is the value of parameter $p_j$ in scenario $s_i$. Define:
\[
\mathcal{S}_{\text{match}} := \left\{ s_i \in \mathcal{S} : \left| \frac{p^*_i}{\bar{p^*}} - \lambda \right| < \epsilon \right\}
\]
as the set of matching or nearest scenarios with respect to $p^*$, within tolerance $\epsilon > 0$.

\paragraph{Scenario Clustering}  
Let $C \in \mathbb{R}^{N \times N}$ be the correlation matrix of scenario outputs (or inputs), and $Z \in \mathbb{R}^{(N-1) \times 4}$ the linkage matrix from hierarchical clustering. Using a fixed threshold $t > 0$, we assign each scenario $S_i$ a cluster label:
\begin{multline*}
\ell_i := \texttt{fcluster}(Z, t),
\quad \\ \text{with cluster membership } \mathcal{C}_k := \{ S_i : \ell_i = k \}.
\end{multline*}

The matching scenario set $\mathcal{S}_{\text{match}}$ is then assigned to a dominant cluster $\mathcal{C}_{k^*}$, from which we extract the cluster-wide scenario descriptions and average correlation:
\[
\bar{\rho}_{k^*} := \frac{1}{|\mathcal{C}_{k^*}|^2} \sum_{i,j \in \mathcal{C}_{k^*}} \rho(S_i, S_j).
\]

\paragraph{Prompt Construction}  
We construct a structured prompt $\Pi(\mathcal{Q}, p^*, \lambda, \mathcal{C}_{k^*}, \bar{\rho})$ for the \llm, encoding:
\begin{itemize}
    \item the original query $\mathcal{Q}$,
    \item matched parameter $p^*$ and requested change $\lambda$,
    \item representative scenarios from $\mathcal{C}_{k^*}$,
    \item cluster summary statistics (size, average correlation).
\end{itemize}

\paragraph{\llm~Querying}  
Finally, the prompt $\Pi$ is submitted to an \llm:
\[
\texttt{LLM}(\Pi) \rightsquigarrow \text{Explanation},
\]
which generates a textual response describing expected changes in output quantities (e.g., cost, abatement, land use), derived from trends within the matched cluster. This approach enables a seamless integration of model-driven scenario analytics with human-centered decision support, grounded in optimization results, similarity structure, and interpretable language.


\paragraph{Outlook}
This integrated pipeline opens the door to horizon-based optimization and multi-factor sensitivity analysis, allowing researchers and decision-makers to explore a wide range of assumptions and assess their systemic implications in a transparent and data-driven manner. While 26 scenarios are considered in this study for tractability and interpretability, the framework is scalable to a larger scenario bank.



\section{Simulation results for the Case Study}
\label{sec:results}
As mentioned, in complex systems modeling—such as land-use, forestry, and climate mitigation planning—raw scenario outputs are often high-dimensional, heterogeneous, and temporally rich. A single scenario can yield thousands of data points across multiple indicators, regions, technologies, and time steps. While this level of detail is essential for representing systemic complexity, it presents a fundamental challenge for data-driven analysis: the \emph{curse of dimensionality}.

To address this, we constructed a compact, interpretable, and generalizable set of features via a carefully designed feature engineering pipeline. The objective is not only to reduce the dimensionality of the scenario data but also to capture domain-relevant dynamics (e.g., temporal trends, regional disparities, and cross-technology interactions) that preserve signal fidelity while enhancing learning efficiency.


\subsection{Feature Illustration}

To illustrate the nature and diversity of the raw scenario inputs before feature engineering, we show a representative subset of time-series and static parameters from the base scenario run of the Zero Negative Emission optimization model in Section \ref{sec:feature_eng}. These variables form the foundational drivers of land-use allocation and mitigation cost outcomes in subsequent analysis.

\begin{figure}
    \centering
    \includegraphics[width=0.65\linewidth]{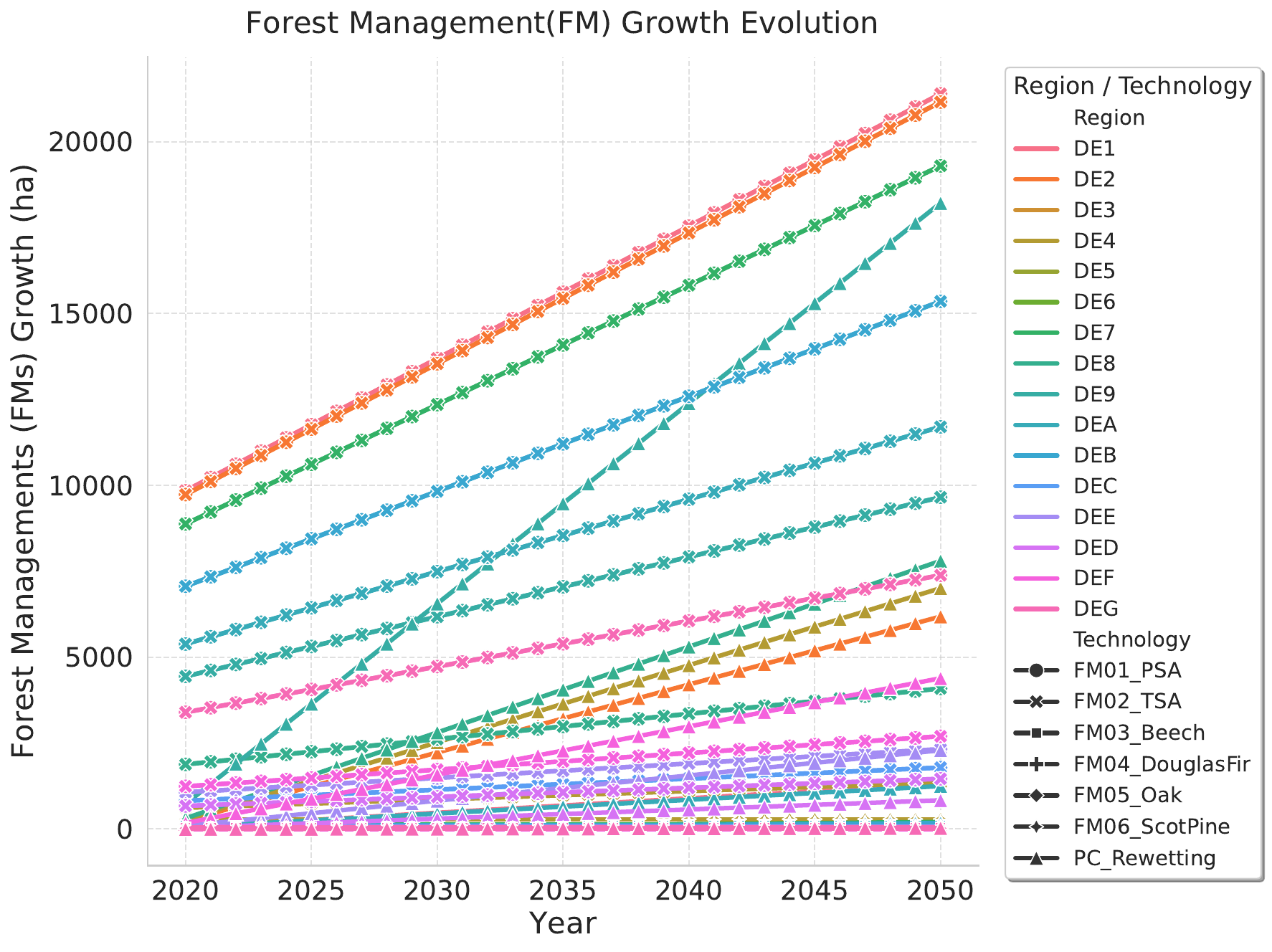}
    \caption{\fm~ capacity growth (ha) by technology and region over 2020--2050. Expansion is constrained by annual adoption ceilings, reflecting practical limits on labor, infrastructure, and ecological readiness.}
    \label{fig:fmsgrowth}
\end{figure}

\begin{figure}
    \centering
    \includegraphics[width=0.65\linewidth]{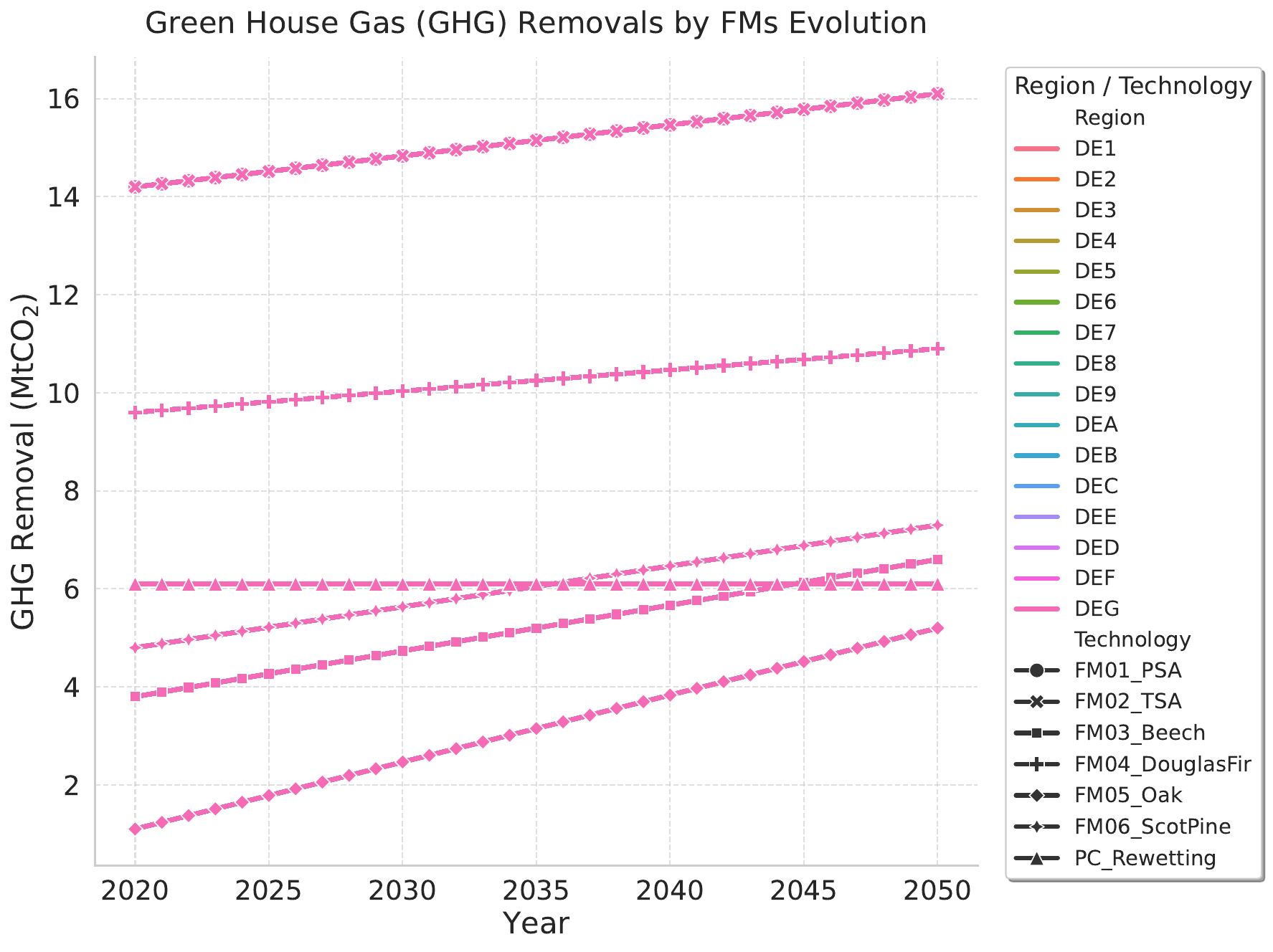}
    \caption{GHG removals by \fm~ technologies and regions over 2020--2050. Growth patterns reflect technology-specific removal factors and capacity adoption limits.}
    \label{fig:ghgfms}
\end{figure}

Figures~\ref{fig:fmsgrowth}-- \ref{fig:ghgfms} illustrate \emph{\ghg~removal, marginal cost, and capacity growth trajectories for \fm~technologies}. Removal potentials differ both by technology (e.g., rewetting vs.\ reforestation) and region, with linear growth patterns constrained by annual adoption ceilings. Marginal costs are largely stable for most \fm~s but escalate significantly for peatland rewetting due to increasing marginal difficulty of restoration. Growth patterns (Figure~\ref{fig:fmsgrowth}) reflect resource constraints and technology-specific saturation points.

\begin{figure}
    \centering
    \includegraphics[width=0.65\linewidth]{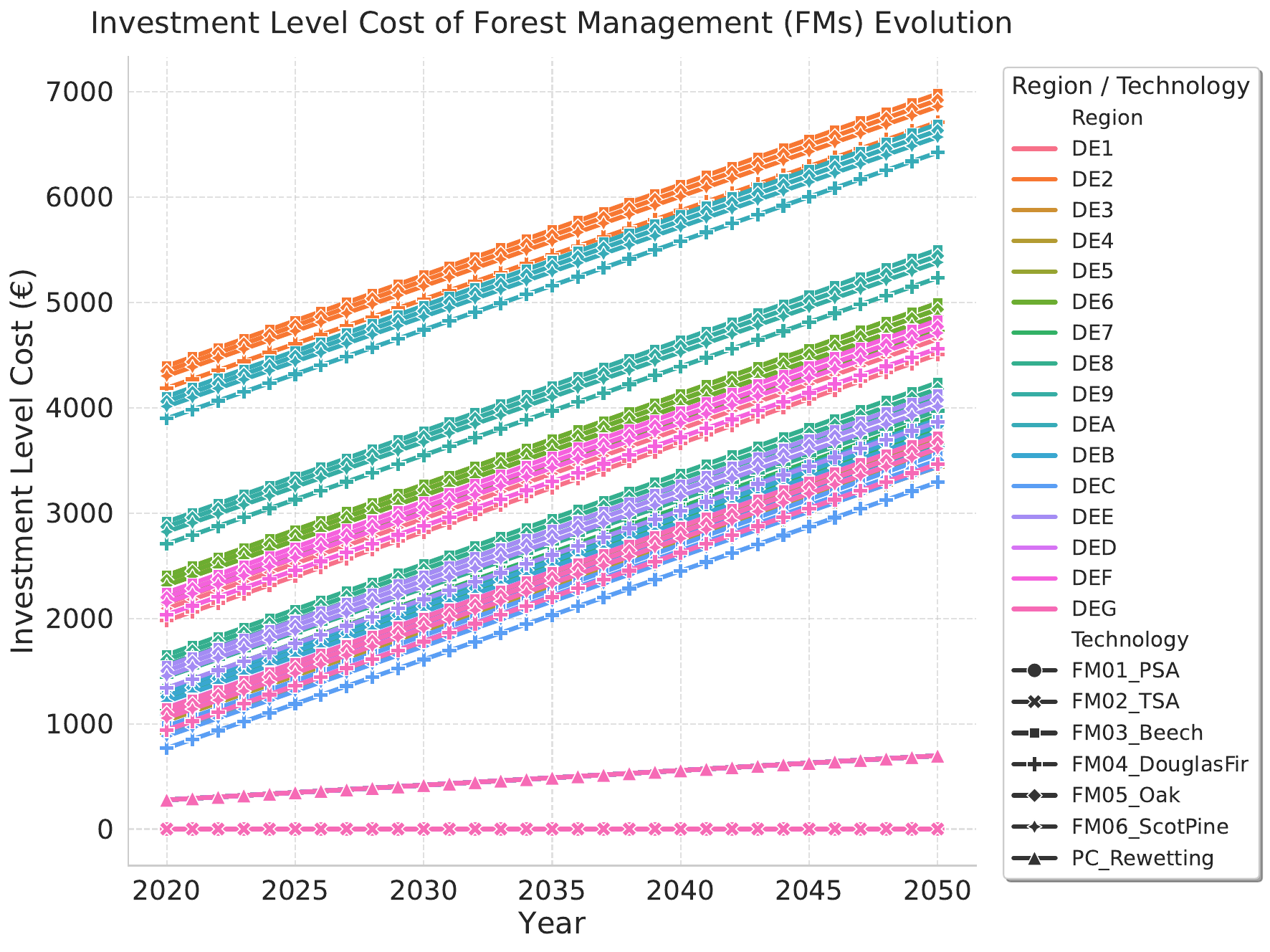}
    \caption{Investment level cost (\,\euro/ha) for \fm~technologies. Assumed constant in real terms for most options, reflecting stable CAPEX assumptions across the planning horizon.}
    \label{fig:costinvlevel}
\end{figure}

\begin{figure}
    \centering
    \includegraphics[width=0.65\linewidth]{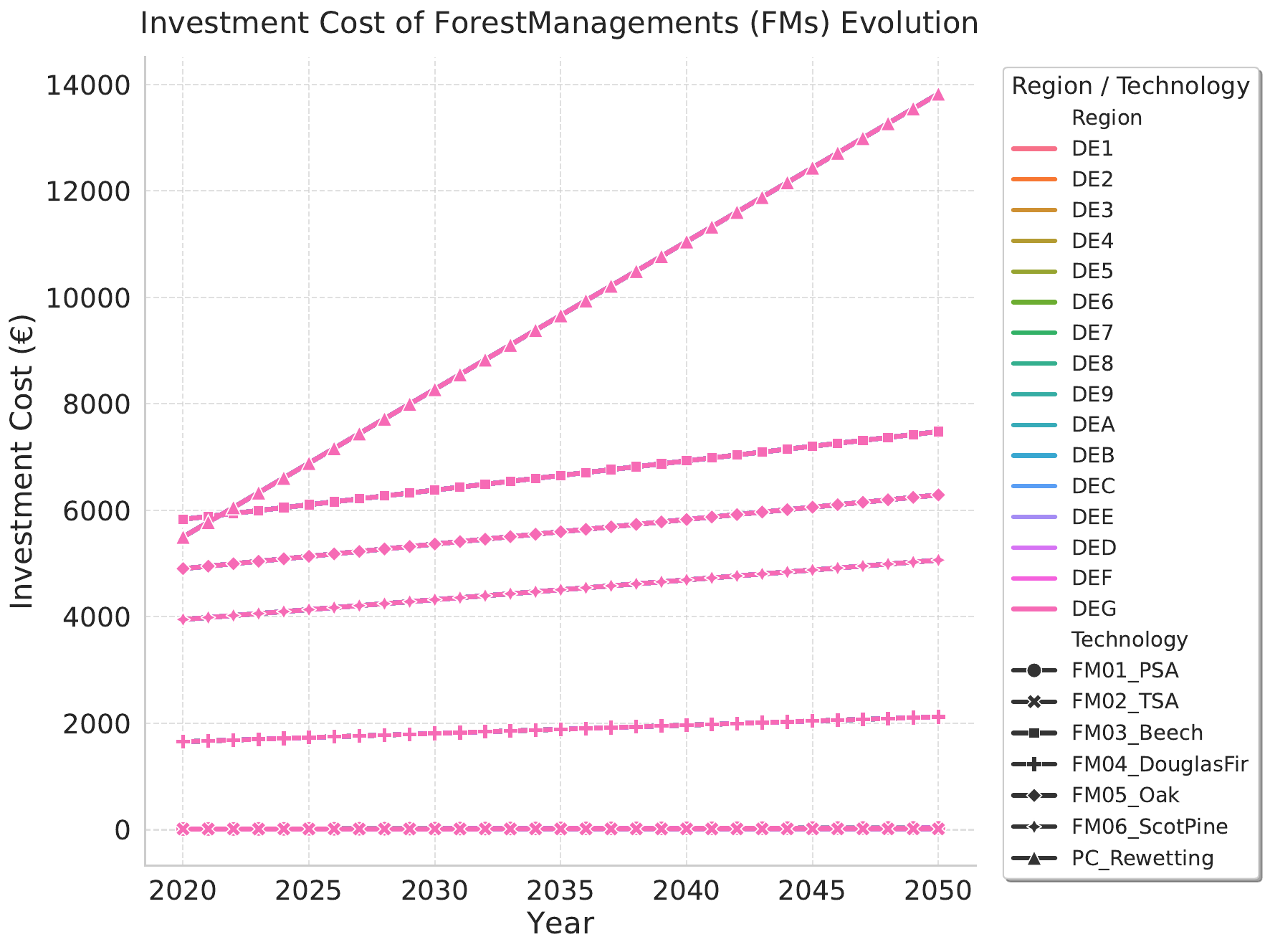}
    \caption{Annual investment costs (M \euro) for \fm~technologies. Growth reflects the combined effect of cumulative capacity additions and constant per-unit investment levels.}
    \label{fig:costinv}
\end{figure}

Figures~\ref{fig:costinvlevel} and~\ref{fig:costinv} show the \emph{investment-related cost metrics for \fm~s}. The \emph{investment level cost} (\,\euro/ha) remains relatively stable over time, reflecting technology-specific \texttt{CAPEX} assumptions. In contrast, \emph{total investment costs} (M\euro/yr) grow as a function of both unit costs and cumulative area expansion.

\begin{figure}
    \centering
    \includegraphics[width=0.85\linewidth]{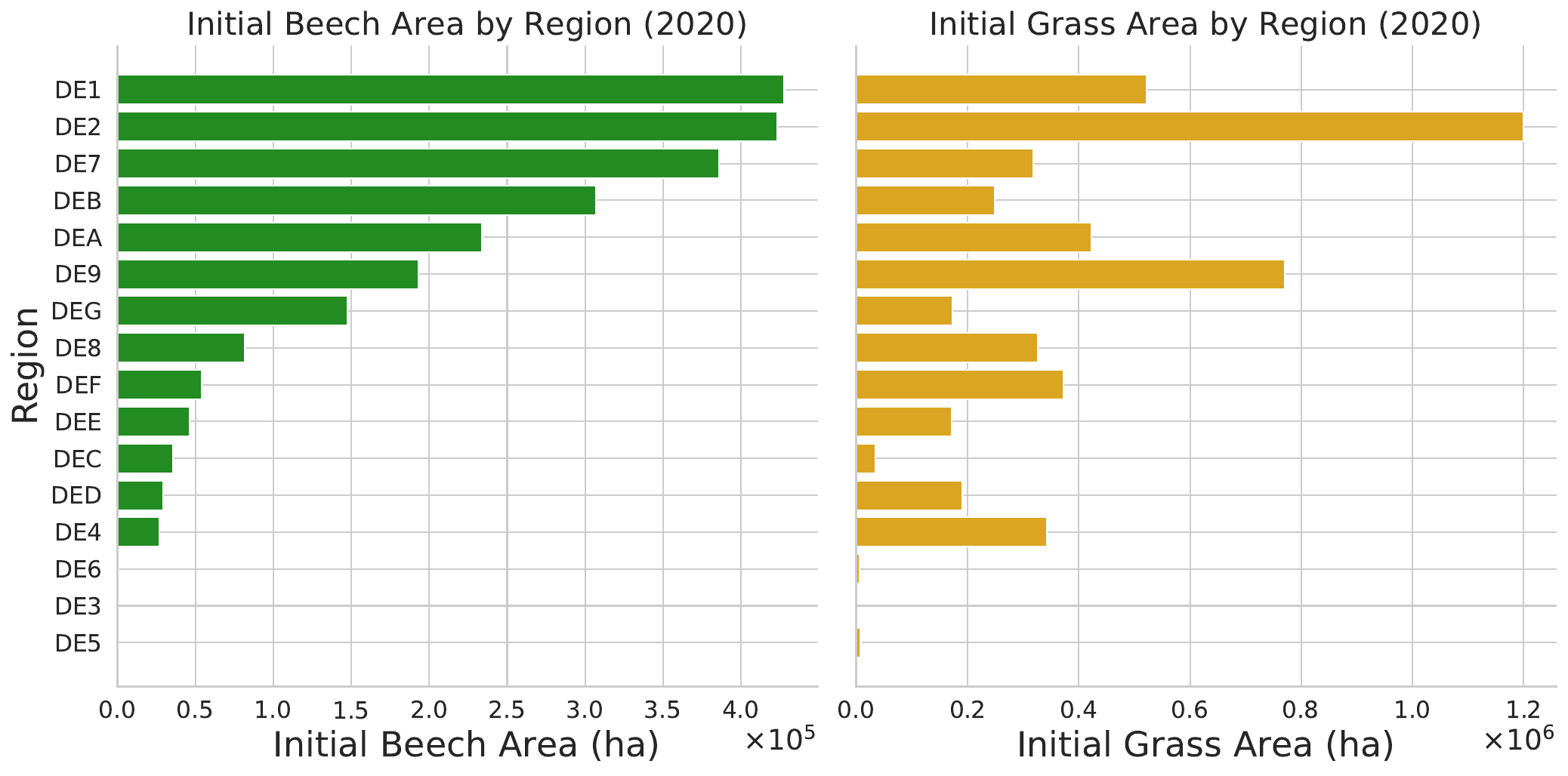}
    \caption{Initial beech and grassland area by region (2020). Static land-use endowments serve as constraints for \fm~deployment pathways.}
    \label{fig:beechgrass}
\end{figure}

Figure~\ref{fig:beechgrass} displays the \emph{initial beech and grassland areas by region} in~2020, which define the upper bounds for set-aside conservation and reforestation measures. Substantial regional disparities are evident, with DE1--DE2 hosting the largest beech areas, while grassland is concentrated in DE2, DE9, and DEA. These static land-endowment constraints critically shape feasible deployment paths for \fm~ options.

\begin{figure}
    \centering
    \includegraphics[width=0.65\linewidth]{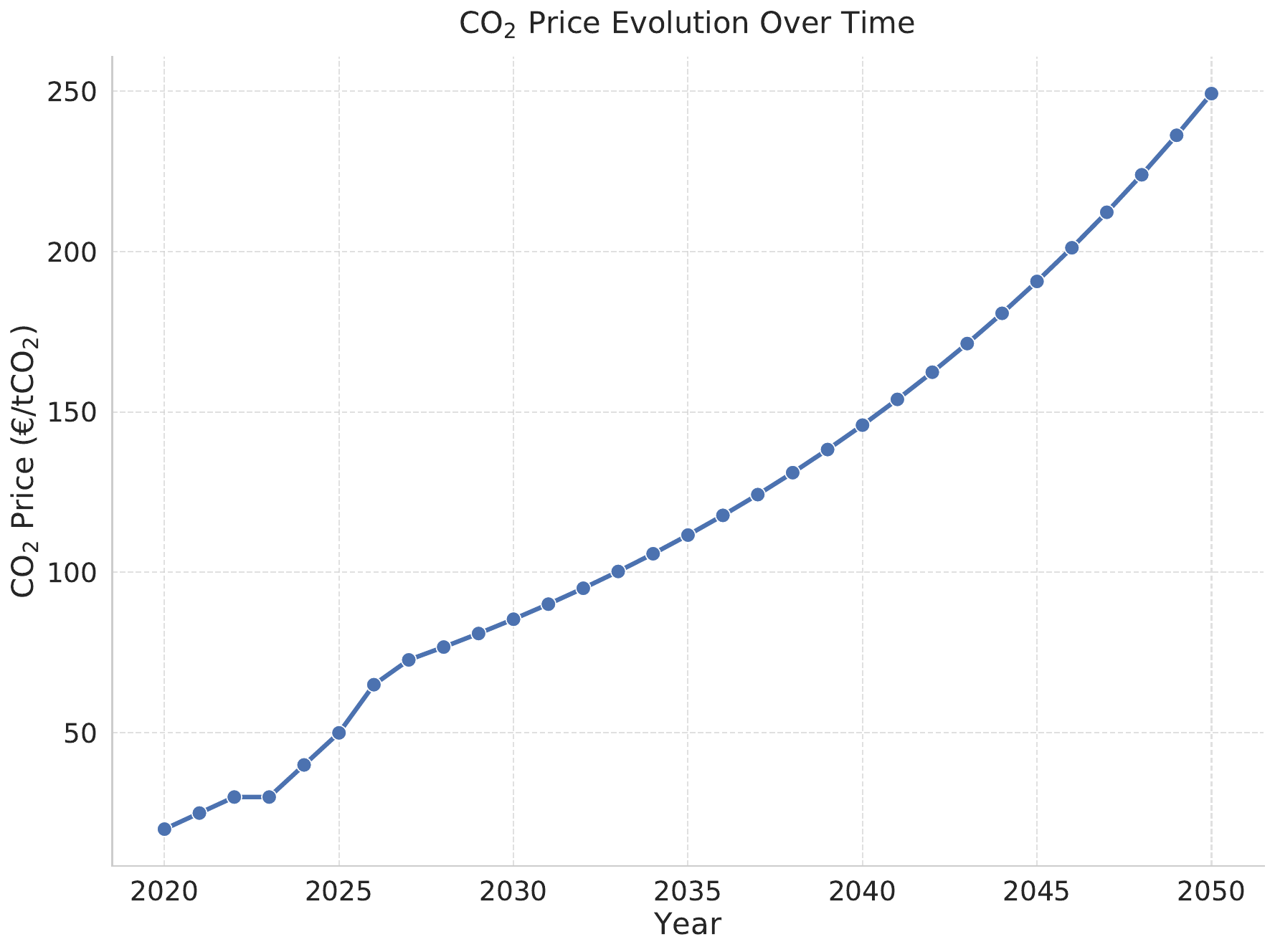}
    \caption{ \ch{CO2} price trajectory over 2020--2050. Rising prices strengthen the economic incentive for land-based mitigation measures.}
    \label{fig:co2price}
\end{figure}

\begin{figure}
    \centering
    \includegraphics[width=0.65\linewidth]{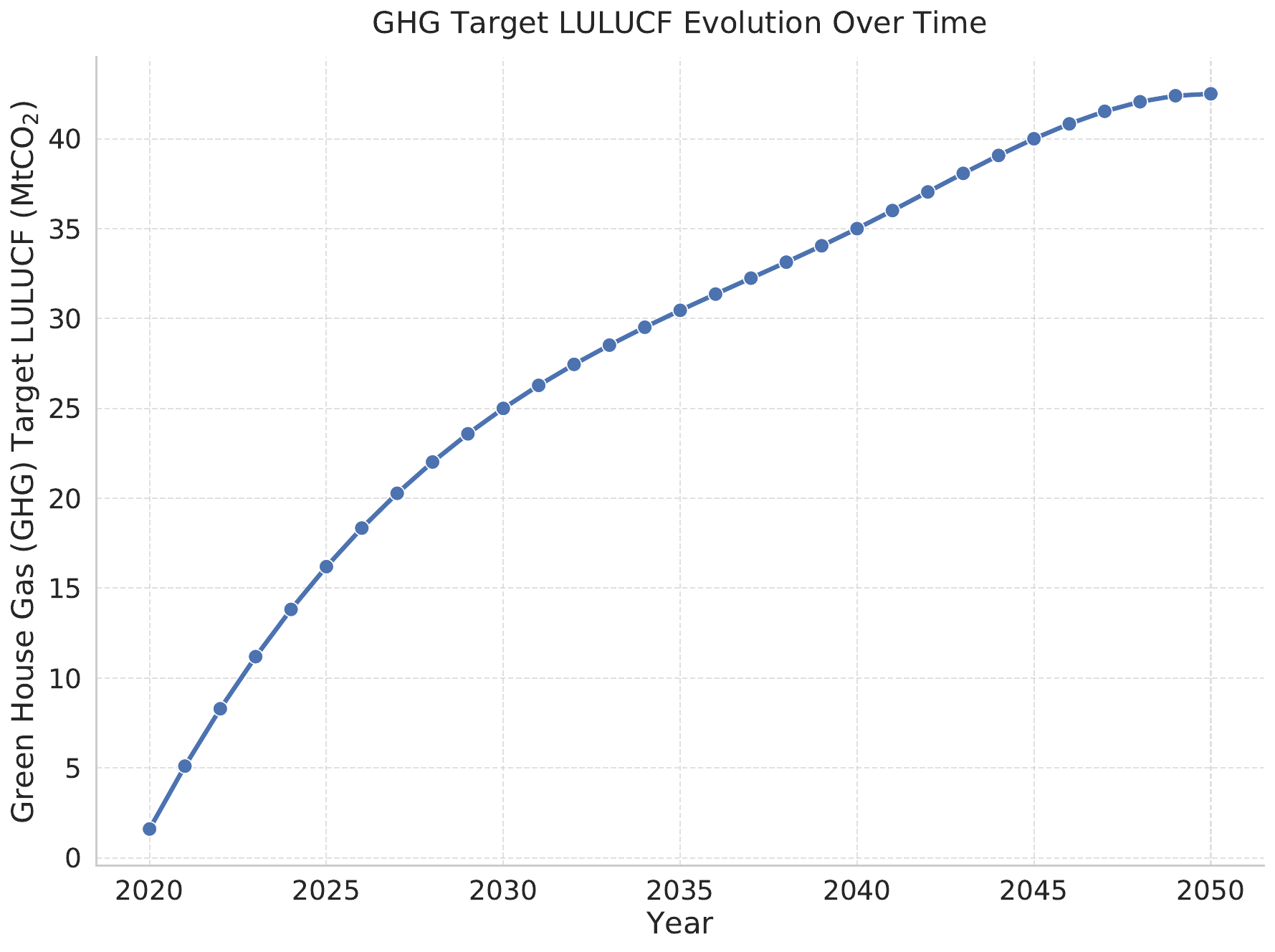}
    \caption{\ghg~target trajectory for the \lulucf~sector over 2020--2050. Increasing targets reflect progressive national and EU climate policy commitments.}
    \label{fig:ghgtarget}
\end{figure}

Figures~\ref{fig:ghgtarget} and~\ref{fig:co2price} present the \emph{\ghg~target trajectory for the \lulucf~sector} and the \emph{CO$_2$ price path} over~2020--2050, respectively. The targets increase steadily toward $42$~MtCO$_2$eq by~2050, while the \ch{CO2} price rises from below~\euro20/t\ch{CO2} to approximately~\euro250/t\ch{CO2} over the same horizon. Together, these time-series define the policy ambition level and economic incentives for mitigation measures.

These raw variables, along with other scenario parameters (e.g., agricultural mitigation options), are directly extracted from the \gams~model outputs. While each parameter has clear sectoral relevance, taken together they form a high-dimensional dataset. Feature engineering (Section~\ref{sec:feature_eng}) condenses these into compact temporal and aggregated descriptors for efficient learning and scenario clustering, without losing interpretability.


\begin{figure}
    \centering
    \includegraphics[width=0.65\linewidth]{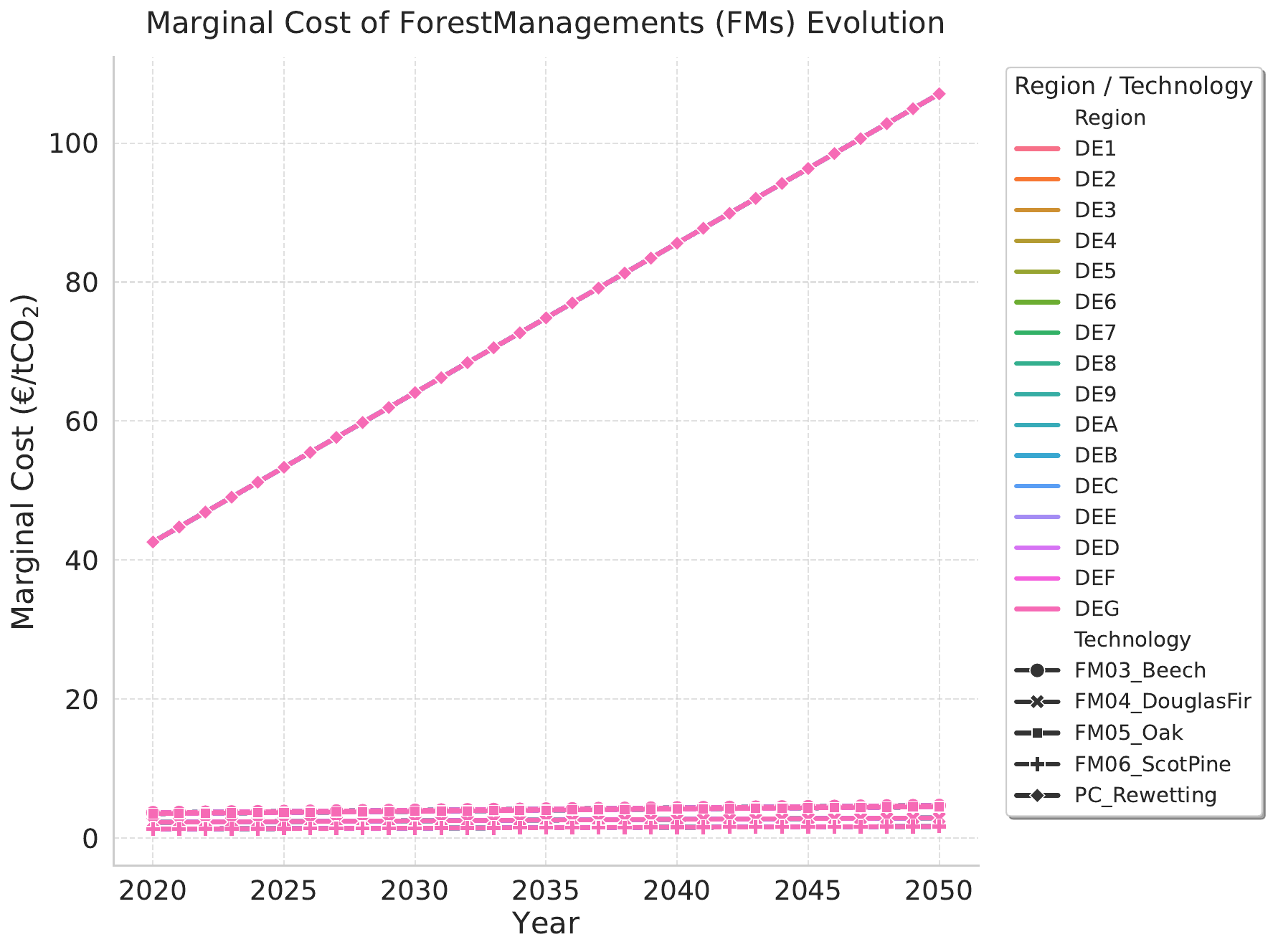}
    \caption{Marginal cost evolution of \fm~technologies. Peatland rewetting exhibits significant cost escalation over time due to increasing marginal restoration difficulty.}
    \label{fig:costmarg}
\end{figure}

Tables~\ref{tab:feature_vector_s10} and \ref{tab:feature_vector_agri_s24} show example feature vectors for Scenario $S{10}$ from the \fm~ scenario bank in Table \ref{tab:scenario_relative_changes_fm} and Scenario $S{24}$ from the Agriculture (\texttt{Agri}) scenario bank Table \ref{tab:scenario_relative_changes_agri}, detailing the complete set of quantitative attributes for a specific region-technology pair under the respective scenario settings. It is worth to highlight that each scenario in \fm~ and \texttt{Agri} has a feature matrix of the size $112\times 25$ and $96 \times 21$, respectively. This shows how feature engineering helps for order reduction in the input dataset.

\begin{table}[ht]
\centering
\caption{An example of feature vector for Scenario $S{10}$ in the bank of scenarios for \fm~ optimization
}
\scriptsize
\begin{tabular}{ll}
\toprule
\textbf{Feature} & \textbf{Value} \\
\midrule
Region & DE2 \\
Technology & FM04\_DouglasFir \\
CostMarg\_2020 & 2.272816 \\
CostMarg\_2050 & 2.914732 \\
CostMarg\_Slope & 0.021397 \\
CostInv\_2020 & 1654.288462 \\
CostInv\_2050 & 2121.512452 \\
CostInv\_Slope & 15.574133 \\
CostInvLevel\_2020 & 5023.172313 \\
CostInvLevel\_2050 & 8052.6193 \\
CostInvLevel\_Slope & 100.981566 \\
GHG\_2020 & 11.52 \\
GHG\_2050 & 13.08 \\
GHG\_Slope & 0.052 \\
ForestGrowth\_2020 & 10.461851 \\
ForestGrowth\_2050 & 10.461851 \\
ForestGrowth\_Slope & 0.0 \\
CO2\_2020 & 20.0 \\
CO2\_2050 & 249.197564 \\
CO2\_Slope & 7.182224 \\
GHGTarget\_2020 & 1.92 \\
GHGTarget\_2050 & 51.0 \\
GHGTarget\_Slope & 1.495102 \\
InitialBeechArea & 423046.85 \\
InitialGrassArea & 1199109.02 \\
\bottomrule
\end{tabular}
\label{tab:feature_vector_s10}
\end{table}

\begin{table}[ht]
\centering
\caption{An example of feature vector for Scenario $S{24}$ in the bank of scenarios for Agriculture optimization
}
\scriptsize
\begin{tabular}{ll}
\toprule
\textbf{Feature} & \textbf{Value} \\
\midrule
Region & DE3 \\
Technology & Agri01\_AGC \\
CostMargAgri\_2020 & 17.25154 \\
CostMargAgri\_2050 & 0.24988 \\
CostMargAgri\_Slope & -0.566722 \\
CostInvAgri\_2020 & 2476.190476 \\
CostInvAgri\_2050 & 35.866426 \\
CostInvAgri\_Slope & -81.344135 \\
CostInvLevAgri\_2020 & 158.803346 \\
CostInvLevAgri\_2050 & 2.30019 \\
CostInvLevAgri\_Slope & -5.216772 \\
GHGAgri\_2020 & 1.8 \\
GHGAgri\_2050 & 1.8 \\
GHGAgri\_Slope & 0.0 \\
AgriGrowth\_2020 & 20.62545 \\
AgriGrowth\_2050 & 21047.182701 \\
AgriGrowth\_Slope & 700.885242 \\
Peat\_Extraction\_2020 & 0.03 \\
Peat\_Extraction\_2050 & 0.3 \\
Peat\_Extraction\_Slope & 0.009 \\
Agriarea0 & 4836.495 \\
\bottomrule
\end{tabular}
\label{tab:feature_vector_agri_s24}
\end{table}



\subsection{Input–Output feature correlation (Forest Management (\fm) vs. Agriculture)}
Both the \fm~ and Agriculture input–feature correlation matrices exhibit \emph{near-perfect} scenario–scenario similarity: almost all off-diagonal Pearson coefficients are \(r \approx 0.999\text{–}1.000\), and hierarchical clusters merge at essentially zero height. This pattern implies that scenarios differ mainly by proportional rescalings of the same underlying signals across regions/technologies/years. The combination of multiplicative scenario levers (e.g., uniform scaling of prices, areas, or growth caps) and subsequent feature normalization further compresses variation, inflating collinearity and rendering the scenario space effectively low-dimensional. Practically, both \fm~ and Agriculture inputs form a single dense cluster, so unsupervised grouping provides limited separation and feature-importance attribution may be fragile unless the perturbation design is diversified (e.g., introducing orthogonal shocks by feature group, region-specific shifts, or non-proportional changes).

\begin{figure}
    \centering
    \includegraphics[width=0.9\linewidth]{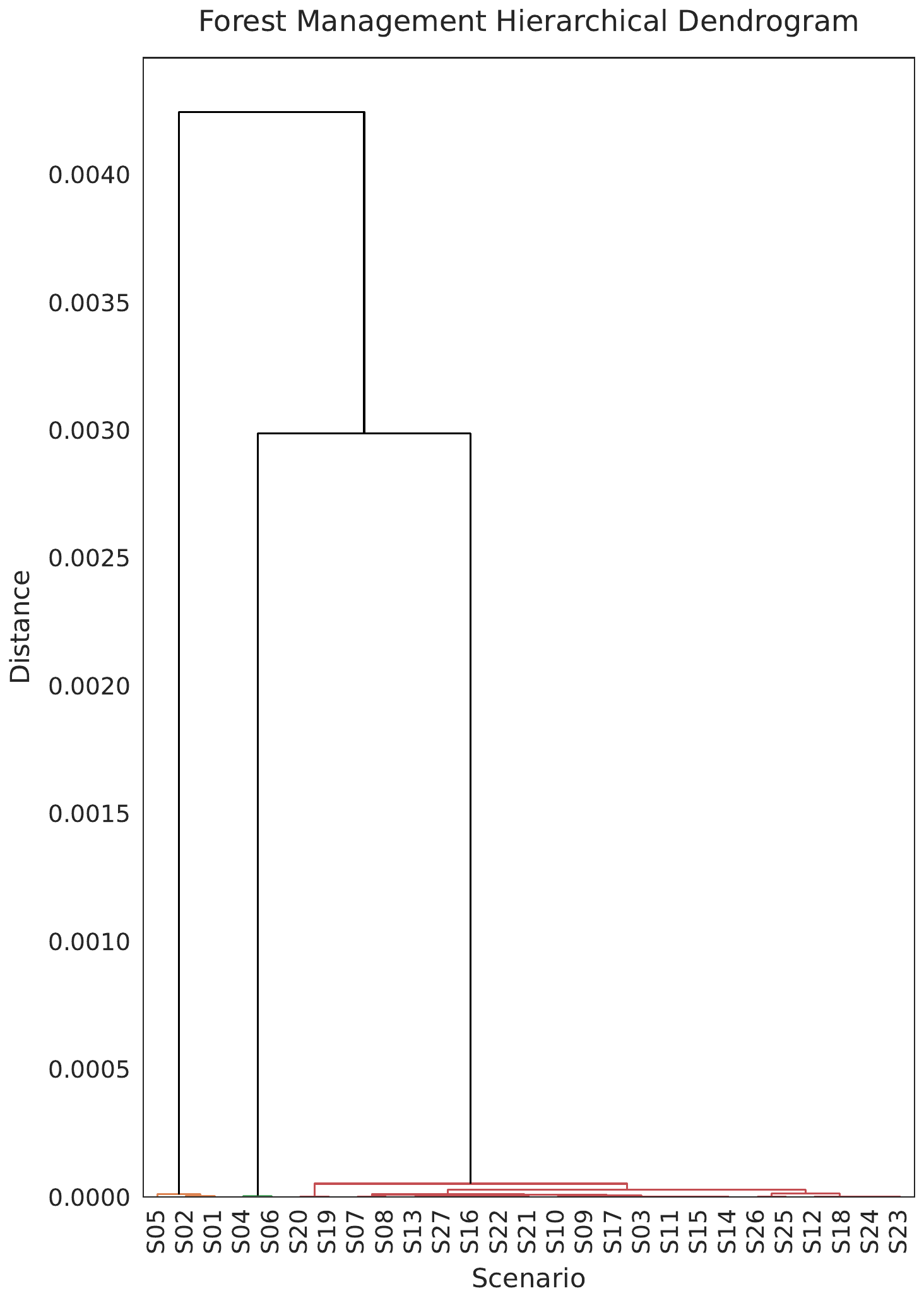}
    \caption{Hierarchical clustering dendrogram from the correlation matrix corresponding to bank of scenarios for the \fm~.}
    \label{fig:fm_input_dendrogram_scenarios}
\end{figure}

\begin{figure}
    \centering
    \includegraphics[width=0.9\linewidth]{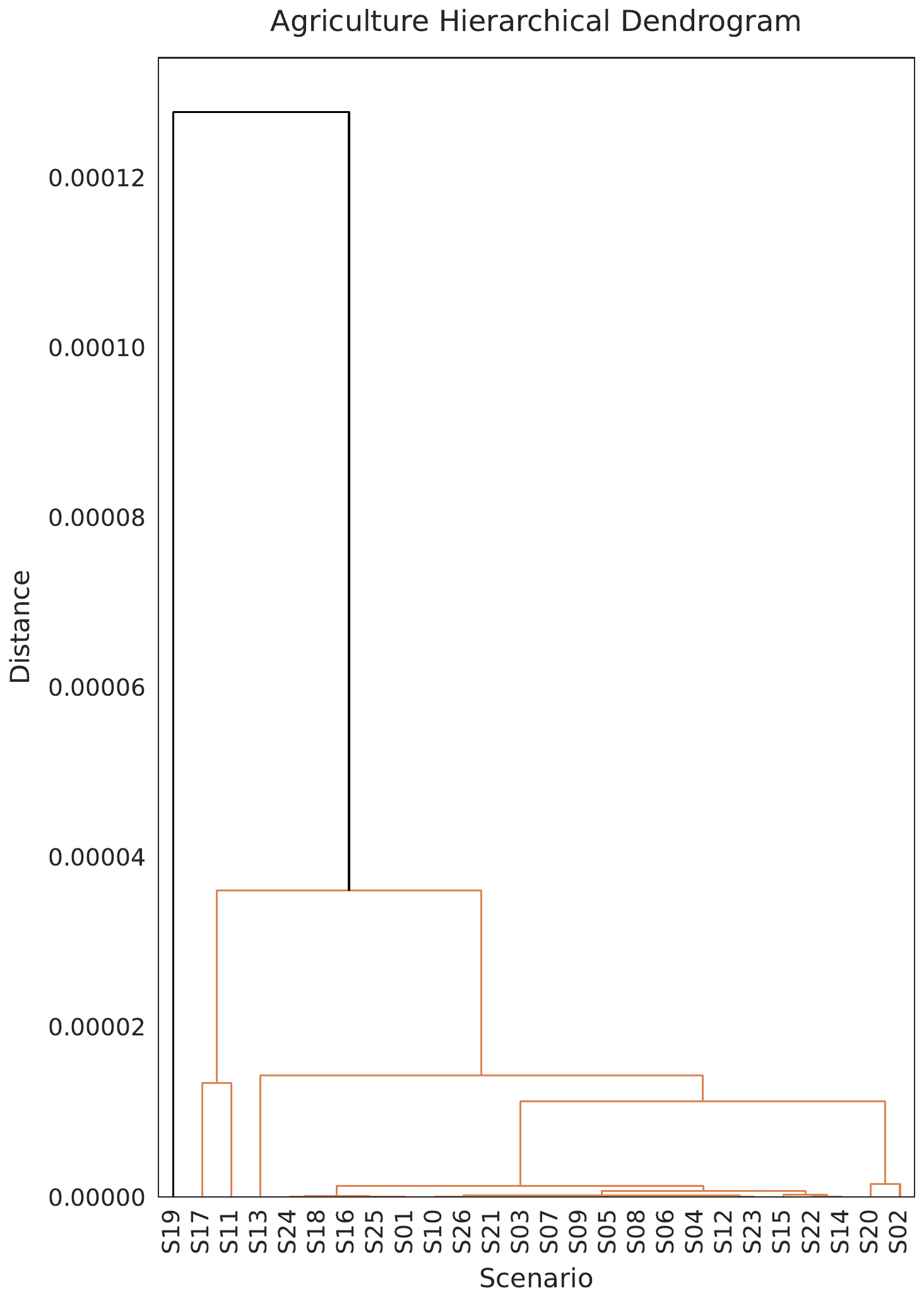}
    \caption{Hierarchical clustering dendrogram from the correlation matrix corresponding to bank of scenarios for the Agriculture optimization.}
    \label{fig:Agriculture_input_scenario_dendrogram}
\end{figure}

\paragraph{Intput–feature correlation}
The correlation matrices for the input features (Figures~\ref{fig:fm_input_dendrogram_scenarios} and \ref{fig:Agriculture_input_scenario_dendrogram}) reveal extremely high pairwise correlations across the 26 scenarios for both \fm~ and Agriculture inputs. Specifically, the \fm~ input matrix exhibits correlation coefficients in the range $[0.987,\,1.000]$ with an average of $0.999$, while the Agriculture input matrix is even more tightly clustered, ranging from $0.9997$ to $1.000$ with an average of $0.99998$. This near-perfect correlation structure reflects the fact that the scenario generation process perturbed key parameters in a highly structured and proportional manner, preserving strong co-movement among input features across scenarios.

\paragraph{Output–feature correlation (costTechAgri and ghgAbateFMs)}
Unlike the inputs—which form a single, nearly collinear cluster—the \emph{outputs} exhibit clear structure and wider dispersion. The costTechAgri matrix shows several well-separated clusters (dendrogram splits up to noticeable heights), with pairwise correlations ranging from high to only moderate values; small, aligned perturbations in inputs are thus \emph{amplified} by the optimization into distinct cost trajectories. This behaviour is consistent with binding constraints and thresholds (e.g., investment limits, land/area and growth ceilings, carbon-price–driven switching of technologies) that create nonlinear responses: once a constraint activates, marginal costs jump and scenarios diverge.

\begin{figure}
    \centering
    \includegraphics[width=0.9\linewidth]{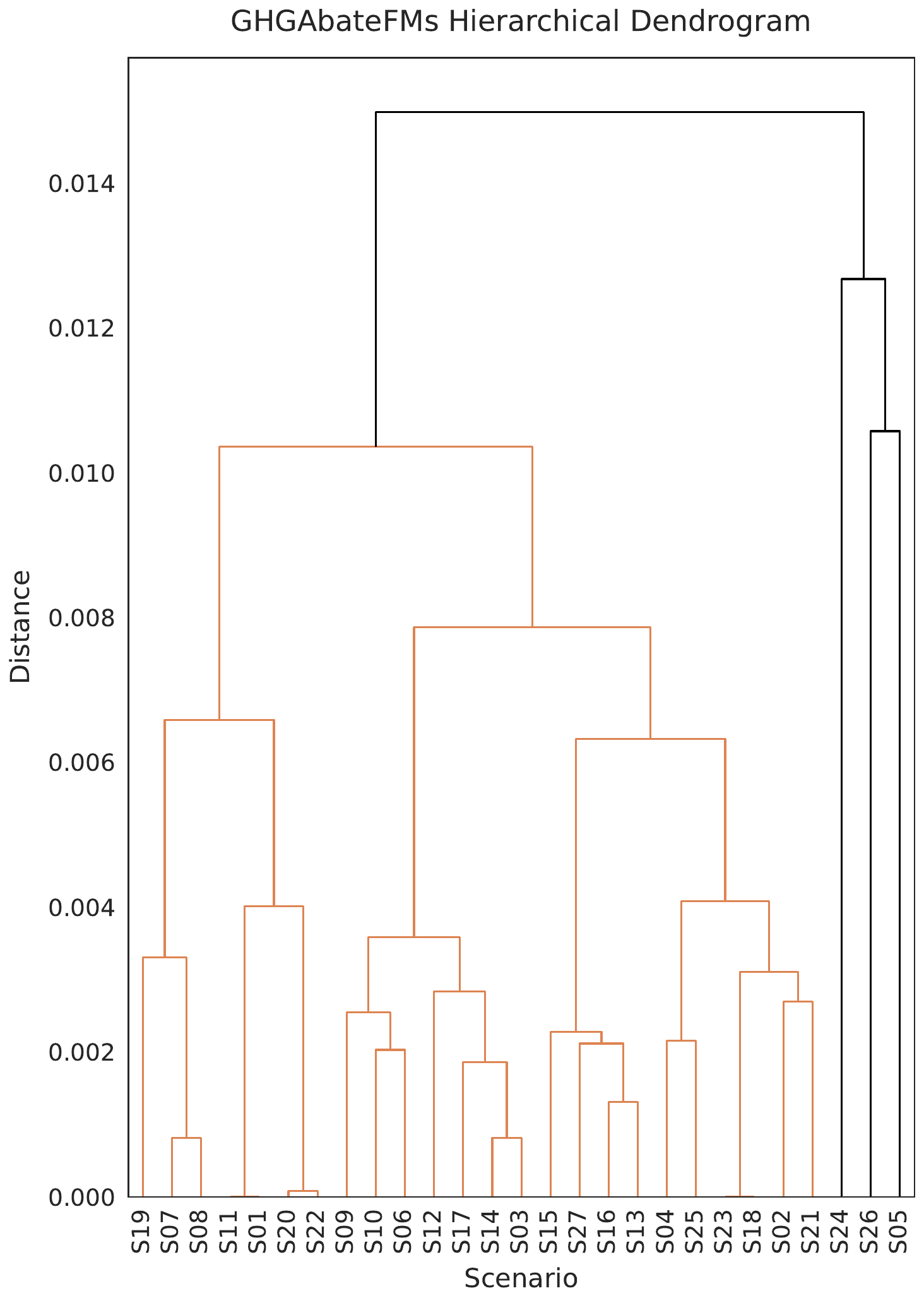}
    \caption{Hierarchical clustering dendrogram from the correlation matrix corresponding to the results of bank of scenarios in the \fm~ for GHGAbateFMs in \eqref{eq:ghgabte}.}
    \label{fig:ghgAbateFms_output_scenario_dendrogram}
\end{figure}

\begin{figure}
    \centering
    \includegraphics[width=0.9\linewidth]{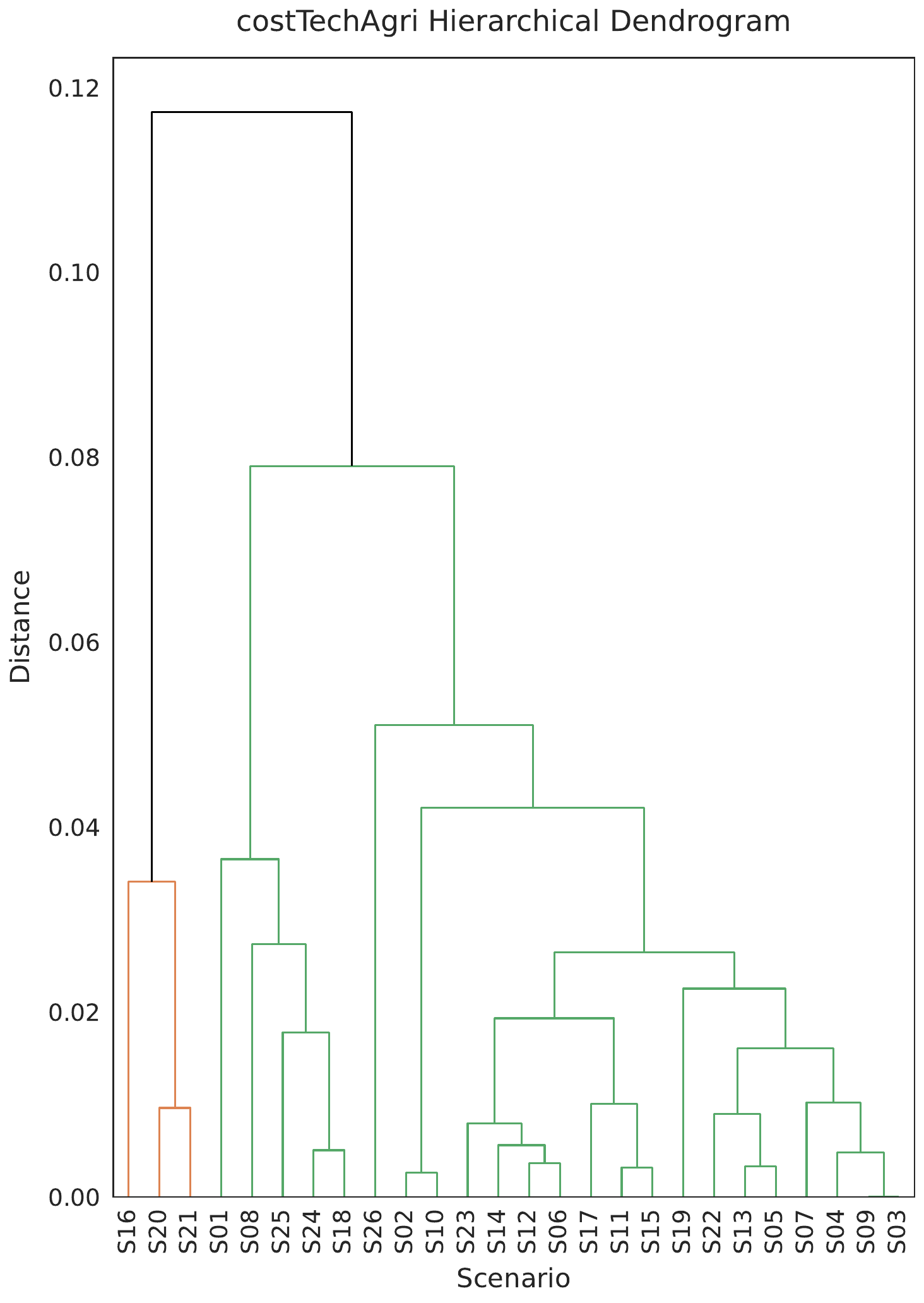}
    \caption{Hierarchical clustering dendrogram from the correlation matrix corresponding to the results of bank of scenarios in the Agriculture optimization for costTechAgri variable.}
    \label{fig:costTechAgri_output_scenario_dendrogram}
\end{figure}

In contrast, the output correlations (Figures~\ref{fig:ghgAbateFms_output_scenario_dendrogram} and \ref{fig:costTechAgri_output_scenario_dendrogram}) display greater variability, highlighting the system's sensitivity to input perturbations. For \fm~ outputs, represented by ghgAbateFMs, correlations remain high but show a wider spread ($[0.973,\,1.000]$, mean $0.991$), indicating that while most \fm~ scenarios respond similarly to input changes, certain combinations of cost, growth constraints, and land availability induce discernible deviations in \ghg~abatement trajectories. The effect is more pronounced for Agriculture outputs, represented by costTechAgri, where correlations span a much broader range ($[0.743,\,1.000]$, mean $0.941$). This suggests that agricultural technology costs are far more sensitive to input parameter variations, likely due to non-linear cost responses, cross-sectoral interactions, and binding constraints in specific scenarios.

Overall, the comparison between input and output correlation structures underscores two key points: (i) the input space is highly coherent across scenarios by design, ensuring controlled and interpretable perturbations, and (ii) the output space, especially for cost-related agricultural metrics, exhibits amplified differentiation, revealing which parts of the system are most responsive to policy or techno-economic shifts. This divergence between input uniformity and output variability is precisely what enables the \rellm~framework to identify and narratively explain the most policy-relevant sensitivities in the \lulucf~sector.

\subsection{Illustration of \ml~surrogate Modeling for \fm~ capacity}

In order to evaluate the predictive capacity of the proposed feature engineering framework, we construct a supervised learning dataset by combining the scenario outputs from the \gams-based optimization model with the engineered input features. The raw optimization results are first processed to extract key variables, including \ch{CO2}price, FMsgrowth, BeechArea0, and GrassArea0, along with the target variable capFMs. Global and regional trend indicators are subsequently derived using moving-window regressions and integrated with technology--region identifiers to form a structured feature space. All continuous predictors are scaled using min--max normalization, while categorical variables (Region, Technology) are represented through one-hot encoding. The resulting feature arrays are cross-joined with the corresponding years and aligned with observed capFMs values, yielding a consistent input--output dataset for model training.  

We employ an ensemble of \randf~regressors \citet{breiman2001random, pedregosa2011scikit}, trained on folded sub-samples of the dataset to ensure robustness against overfitting and to capture non-linear dependencies between features and forest management capacities. Model performance was assessed using out-of-sample predictions, with metrics reported both in normalized units and in original hectares.

In the ensemble training of \fm~ capacity, we employ $n_{\text{folds}} = 10$ (cross-validation) to generate diverse sub-models. Each base learner is a \randf~Regressor with \texttt{n\_estimators = 50} trees and varying random seeds across folds to promote model diversity. No explicit constraints on tree depth, minimum split size, or leaf size were introduced. Target values were normalized using Min--Max scaling before training and subsequently inverse-transformed for evaluation. This setup provides a balanced ensemble that emphasizes variance reduction while maintaining computational efficiency.

\begin{figure}
    \centering
    \includegraphics[width=0.65\linewidth]{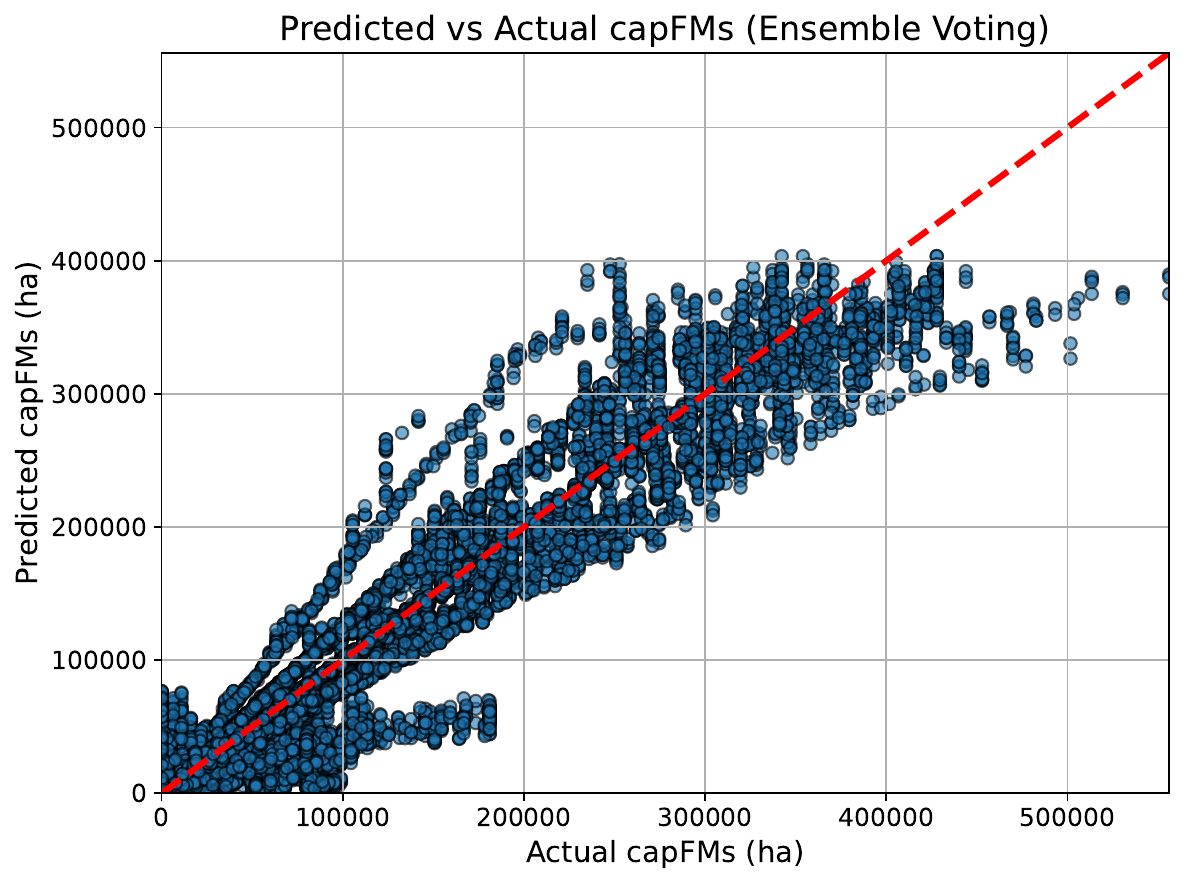}
    \caption{Predicted versus actual values of the \fm~ capacity variable (\texttt{capFMs}) obtained using the \randf~ensemble model}
    \label{fig:capfms_rf}
\end{figure}

The ensemble achieved high explanatory power, with \(R^2\) scores exceeding $0.9440$ and root mean squared errors (RMSE) on the order of $10150.07$ hectares. As illustrated in Figure ~\ref{fig:capfms_rf}, the predicted values exhibit strong alignment with observed capacities along the 45$^\circ$ reference line. These findings confirm that the engineered scenario features effectively encapsulate the main drivers of \fm~capacity, and that the ensemble learning approach yields stable generalization performance across heterogeneous scenario settings.  

\begin{figure}
    \centering
    \includegraphics[width=0.99\linewidth]{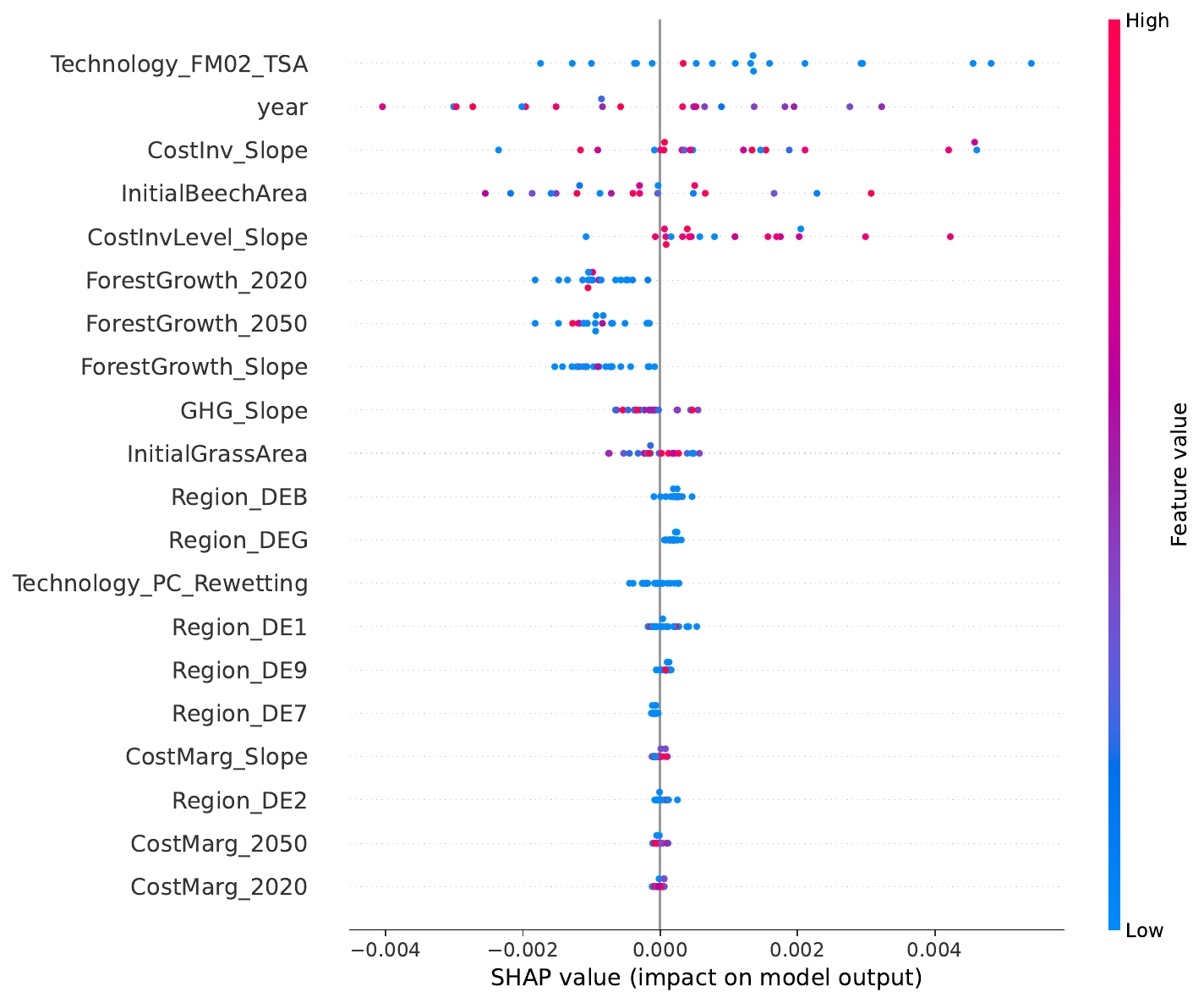}
    \caption{SHAP (SHapley Additive exPlanations) summary plot for the ensemble \randf~surrogate model 
predicting \fm~ capacity.}
    \label{fig:shap_capFMs}
\end{figure}

To interpret the ensemble predictions, we apply SHAP (SHapley Additive exPlanations) using a TreeSHAP approach tailored to the \randf~regressors. The procedure aggregates explanations across models and repeated subsamples of the test set, ensuring robust feature attribution. Figure~\ref{fig:shap_capFMs} illustrates the resulting SHAP summary plot, where each point represents the marginal contribution of a feature to the predicted Forest Management capacity (capFMs). Each point represents the marginal contribution of a feature to the model output across the test data. The horizontal axis indicates the SHAP value (impact on prediction), while the color encodes the feature value from low (blue) to high (red). The most influential drivers are the technology indicator FM02\_TSA, temporal effects (year), and investment-related slopes (CostInv\_Slope, CostInvLevel\_Slope), alongside regional initial conditions such as InitialBeechArea and InitialGrassArea. Positive SHAP values (right side) increase the predicted capacity, while negative values (left side) decrease it, with color encoding reflecting the underlying feature values. This analysis confirms that both cost trends and land endowments interact strongly with regional and technology-specific factors, shaping the optimization outcomes and validating the surrogate model's interpretability within the \rellm~pipeline.

\paragraph{SHAP-guided \llm~narratives}
To make the language layer faithful to the underlying surrogate, we use SHAP attributions as a structured control signal for prompt construction. Concretely, after computing ensemble SHAP matrices $\bar{\Phi}^{(\text{ensemble})}$ and their global importance $\Phi^{\text{global}}$ (local accuracy, missingness, and consistency hold by construction), we select the top-$k$ drivers (feature, sign, magnitude, representative value) and serialize them into the \llm~prompt used for stakeholder summaries.\footnote{See our SHAP formulation and aggregation across $K$ models and repeated test subsamples, and the definition of $\Phi^{\text{global}}$ for population-level ranking}. In practice, the SHAP summary (e.g., Figure \ref{fig:shap_capFMs} for capFMs) identifies cost trends, land endowments, and technology markers as dominant factors; these are injected into a templated prompt that the \llm~converts into a concise, audience-appropriate explanation. By grounding the \llm~with SHAP-ranked evidence, the narrative remains traceable to the model's feature attributions and reduces the risk of free-form speculation—operationalizing the link between interpretable \ml~and our natural-language interface. In the Appendix Section \ref{app:llm_prompt_shap} a prompt is illustrated to showcase how SHAP analysis can play role in \footnote{For the \llm~query pipeline (parameter parsing, scenario matching/clustering, and prompt construction), see Section~\emph{\ref{subsec:llm_query}}}.


\subsection{\llm-driven Stakeholder Interaction}\label{subsec:stack_holder_queries}
Stakeholders may issue natural language queries, such as \emph{``What happens if \ch{CO2} price increases by 20\%?''} or \emph{``What happens if cost of investment in agriculture i.e. costInvAgri decreases?''}. These are mapped to scenario subsets using string parsing and clustering metadata. Prompts are constructed and sent to an \llm, which returns a human-readable summary.

\paragraph{Query 1 — ``What happens if the CO\textsubscript{2} price increases by 20\%?''}
The \llm~matched this query to scenarios S06–S08, where the carbon-price multiplier is $1.2$ while other levers (e.g., FMsgrowth, BeechArea0) vary within their designed ranges. At the cluster level, these runs fall within a coherent group (high intra-cluster correlation), so the narrative reflects behavior that is robust across nearest neighbors. In short, a higher \ch{CO2} price strengthens incentives to deploy land-based mitigation earlier and at larger scale: cumulative \ghg~abatement rises, reliance on external allowances falls, and technologies with strong abatement per hectare (notably peatland rewetting and set-aside) scale faster subject to land and growth ceilings. Cost trajectories increase when adoption fronts meet binding constraints, but the net system-cost pressure is partially offset by reduced purchases of \ch{CO2} credits (the model penalizes external credits relative to the domestic price). The explanation is grounded in the scenario metadata and the correlation-aware clustering used to construct the prompt, ensuring that the textual summary remains traceable to modeled evidence. In Appendix \ref{lst:stakeholder_response_co2} the output of the \llm~to this query is shown.

\paragraph{Query 2 — ``What happens if the cost of investment in agriculture (\texttt{costInvAgri}) decreases?''}
For this query, the \llm~retrieved the agriculture subset with reduced investment costs (e.g., S10: $-50\%$, S11: $-20\%$, S12: $-20\%$ with coupled levers) and composed a response against the same cluster-aware template. Lower costInvAgri systematically depresses costTechAgri (levelized costs), bringing forward adoption of agricultural mitigation options (agroforestry, cover crops, soil-carbon enhancement) until constrained by land-share and annual growth caps. In turn, the model shifts part of the abatement burden from forest management to agriculture where the marginal abatement cost becomes cheaper, while respecting area constraints and peatland policies. Because agricultural outputs in these experiments exhibit wider dispersion than inputs (optimization non-linearities around binding constraints), the \llm~highlights conditionals---where/when the steepest costTechAgri declines appear, and which regions/technologies benefit first---and links them to the specific cost-lever scenarios surfaced by the matcher. This keeps the narrative faithful to the designed perturbations while remaining stakeholder-friendly. In Appendix \ref{lst:stakeholder_response_costinvagri} the output of the \llm~to this query is shown.

\paragraph{Reproducibility note}
Both paragraphs are generated with the documented pipeline: (i) parse the natural-language query to extract parameter changes; (ii) match to nearest scenarios in the bank; (iii) ground the prompt using output-space correlation structure and cluster representatives; and (iv) return an explanation that cites the representative scenarios/clusters. This preserves traceability from the \llm~text to the underlying scenario set and clustering artifacts.

\section{Conclusion} \label{sec:conclusion}

This paper introduced \rellm, a hybrid framework that integrates optimization-based scenario analysis, machine learning surrogates, and large language models (\llm s) into a unified pipeline for renewable energy and land-use planning. By combining classical \gams-based optimization with structured feature engineering, correlation and clustering analysis, and surrogate modeling, the framework enables rapid exploration of complex scenario spaces. The \llm~layer further enhances interpretability by translating quantitative outputs into accessible narratives tailored to diverse stakeholders.

Through the German \lulucf~case study, we demonstrated that surrogate models (\randf~, \dnn, \texttt{XGBoost}) can reproduce optimization outcomes with high fidelity while reducing computational requirements by more than an order of magnitude. SHAP-based interpretability added transparency to model behavior, highlighting the most influential drivers of mitigation costs and capacities. Finally, the \llm-assisted querying pipeline illustrated how natural language interfaces can bridge the gap between technical outputs and actionable insights, thereby improving stakeholder comprehension and engagement.

Overall, \rellm~advances a human-centric paradigm for energy system modeling, where computational efficiency and interpretability are treated as complementary design goals. By merging speed, rigor, and accessibility, the framework supports interactive, multilingual, and policy-relevant engagement with future energy pathways. Future research will extend this approach toward larger scenario banks, multi-country applications, and deeper integration of \llm s with optimization solvers, ultimately contributing to more transparent and participatory energy transition planning.

\section*{Acknowledgment}
This study is funded by the \href{https://man0euvre.eu/}{\emph{Man0EUvRE}} (100695543), which is co-financed by means of taxation based on the budget adopted by the representatives of the Landtag of Saxony. ``Man0EUvRE – Energy System Modelling for Transition to a net-Zero 2050 for EU via REPowerEU,'' is funded by CETPartnership, the European Partnership under Joint Call 2022 for research proposals, co-funded by the European Commission (GAN°101069750).

\appendix

\subsection{Summary of notable energy‐system optimization frameworks}
\begin{landscape}
\begin{table*}[h]\label{Tab:en_sys_opt_fram}
\centering
\caption{Summary of notable energy‐system optimization frameworks.}
\scriptsize
\begin{tabular}{@{}p{3cm}p{3cm}p{3cm}p{4cm}p{3cm}@{}}
\toprule
\textbf{Framework} & \textbf{Licence / language} & \textbf{Coupled sectors} & \textbf{Typical use case} & \textbf{Notable feature(s)} \\
\midrule
\Times & Proprietary model generator (\gams) & Power, fuels, industry, transport, land & National LT planning, marginal CO$_2$ pricing & Mature IEA-ETSAP community and large global database \citet{loulou2005documentation, etsap:times, wikipedia:times} \\
\texttt{MESSAGE-ix} \citet{huppmann2019messageix, messageix:docs, messageix:openmod}& Open source (\py/\gams) & Full energy, engineering, economy, environment (\texttt{E4}) & IAM‑consistent energy scenarios, SDG co‑analysis & Multiregional, perfect‑foresight, flexible solver back‑end  \\
\texttt{OSeMOSYS} \citet{howells2011osemosys} & MIT‑style licence (\py/\texttt{GNU MathProg}) & Power, fuels; heat \& transport via extensions & Capacity‑expansion in data‑poor regions & Minimal core; ideal for teaching and rapid prototyping  \\
\texttt{PyPSA} / \texttt{PyPSA‑Eur‑Sec} \citet{brown2017pypsa} & GPL (\py / \texttt{Pyomo}) & Electricity, heating, transport, hydrogen & High‑res European sector coupling (hourly) & Integration with open GIS data; \milp~for unit commitment \citet{pypsa:wikipedia} \\
\texttt{GenX} \citet{hopkins2018recently} & Apache‑2.0 (\texttt{Julia} / \texttt{JuMP}) & Power; emerging heat and hydrogen modules & Flexible‑resource valuation, system adequacy & Semi‑continuous unit commitment \& storage clustering \\
\texttt{ReEDS} \citet{ho2021regional} & \nrel~proprietary (\gams) & Power; links to e‑mobility \& fuels via add‑ons & U.S.\ decarbonisation studies, transmission siting & Probabilistic resource‑adequacy and zoning algorithm \\
\texttt{EnergyPLAN} \citet{lund2021energyplan} & Freeware GUI (\texttt{Delphi}) & Electricity, DH, transport fuels & Hour‑level operational feasibility, policy mixes & ``Smart‑energy‑system'' heuristic with excess balancing \\
\texttt{GCAM} \citet{calvin2019gcam} & BSD‑style (\texttt{C++}) & Economy, energy, land‑use, water \& climate & Integrated land‑energy pathways, AFOLU mitigation & Dynamic‑recursive market equilibrium approach \\
\texttt{PLEXOS} \citet{plexos:energyexemplar} & Commercial (\texttt{C\#}/\texttt{SQL}) & Primarily electricity + gas/water co‑simulation & Market dispatch, generation bidding, nodal prices & Sub‑hourly chronological \lp/\milp engine \\
\texttt{urbs} \citet{github:urbs} & MIT licence (\py/\texttt{Pyomo}) & Multi‑commodity (electricity, heat, hydrogen) & Urban \& regional capacity expansion & Straightforward hourly LP, Pandas native I/O \\
\bottomrule
\end{tabular}
\end{table*}
\end{landscape}


\subsection{Bank of \fm~and Agriculture scenarios}

\begin{landscape}
\begin{table*}[ht]\label{tab:scenario_relative_changes_fm}
\centering
\caption{Bank of Scenarios for \fm~ optimization;\\
\footnotesize\emph{Legend:} \textbf{*} = costInvAgri / costMargAgri / costInvLevelAgri / Agriarea0.
}
\tiny
\begin{tabular}{lcccccccccccc}
\toprule
\textbf{Scenario} & \textbf{CO2price} & \textbf{FMsgrowth} & \textbf{BeechArea0} & \textbf{GrassArea0} & ${\tau}^{\text{LULUCF}}_{t}$ & \textbf{cap0FMs} & \textbf{costMargFMs} & \textbf{costInvFMs} & \textbf{costInvLevelFMs} & \textbf{ghgFMs} & \textbf{*} \\
\midrule
S01 & -20\% & -20\% & -20\% & --    & --    & --    & --    & --    & --    & --    & -- \\
S02 & -20\% & 0\%   & -20\% & --    & --    & --    & --    & --    & --    & --    & -- \\
S03 & -20\% & 0\%   & 0\%   & --    & --    & --    & --    & --    & --    & --    & -- \\
S04 & +20\% & 0\%   & +20\% & --    & --    & --    & --    & --    & --    & --    & -- \\
S05 & +20\% & +20\% & -20\% & --    & --    & --    & --    & --    & --    & --    & -- \\
S06 & +20\% & +20\% & +20\% & --    & --    & --    & --    & --    & --    & --    & -- \\
S07 & --    & -20\% & --    & --    & --    & --    & --    & --    & -20\% & --    & -- \\
S08 & --    & -20\% & --    & --    & -20\% & --    & --    & --    & -20\% & -20\% & -- \\
S09 & --    & +20\% & --    & --    & --    & --    & --    & --    & +20\% & --    & -- \\
S10 & --    & +20\% & --    & --    & +20\% & --    & --    & --    & +20\% & +20\% & -- \\
S11 & --    & -20\% & -20\% & -20\% & -20\% & -20\% & --    & --    & --    & --    & -- \\
S12 & --    & +20\% & +20\% & +20\% & +20\% & +20\% & --    & --    & --    & --    & -- \\
S13 & --    & --    & --    & --    & --    & --    & --    & --    & --    & --    & -20\% \\
S14 & --    & --    & --    & --    & --    & --    & --    & --    & --    & --    & +20\% \\
S15 & --    & --    & --    & --    & --    & --    & --    & -20\% & -20\% & --    & -- \\
S16 & --    & --    & --    & --    & --    & --    & --    & -10\% & +10\% & --    & -- \\
S17 & --    & --    & --    & --    & --    & --    & --    & +20\% & +20\% & --    & -- \\
S18 & --    & --    & --    & --    & --    & --    & -50\% & -50\% & -50\% & --    & -- \\
S19 & --    & --    & --    & --    & --    & --    & -50\% & -50\% & -50\% & -50\% & -- \\
S20 & --    & --    & --    & --    & --    & -20\% & -20\% & -20\% & -20\% & --    & -- \\
S21 & --    & --    & --    & --    & --    & --    & -20\% & -20\% & -20\% & -20\% & -- \\
S22 & --    & --    & --    & --    & --    & +20\% & +20\% & +20\% & +20\% & --    & -- \\
S23 & --    & --    & --    & --    & --    & --    & +20\% & +20\% & +20\% & +20\% & -- \\
S24 & --    & --    & --    & --    & --    & --    & +50\% & +50\% & +50\% & --    & -- \\
S25 & --    & --    & --    & --    & --    & --    & +50\% & +50\% & +50\% & +50\% & -- \\
S26 & --    & +30\% & +30\% & +30\% & +30\% & --    & --    & --    & --    & --    & -- \\
\bottomrule
\end{tabular}
\end{table*}
\end{landscape}

\begin{landscape}
\begin{table*}[ht]\label{tab:scenario_relative_changes_agri}
\centering
\caption{Bank of Scenarios for Agriculture Optimization}
\tiny
\begin{tabular}{lccccccccccc}
\toprule
\textbf{Scenario} & \textbf{CO2price} & \textbf{FMsgrowth} & \textbf{BeechArea0} & \textbf{costMargAgri} & \textbf{costInvAgri} & \textbf{costInvLevelAgri} & \textbf{ghgAgri} & \textbf{Agrigrowth} & \textbf{Agriarea0} & \textbf{PeatExtract} \\
\midrule
S01 & --     & --     & --     & --     & --     & --     & --     & +90$\%$  & +90$\%$  & +90$\%$ \\
S02 & --     & --     & --     & --     & --     & --     & --     & -60 $\%$  & -60 $\%$  & -60 $\%$ \\
S03 & -20 $\%$  & -20 $\%$  & -20 $\%$  & --     & --     & --     & --     & --     & --     & --    \\
S04 & -20$\%$  & 0 $\%$    & -20$\%$ & --     & --     & --     & --     & --     & --     & --    \\
S05 & -20$\%$ & 0$\%$    & 0$\%$    & --     & --     & --     & --     & --     & --     & --    \\
S06 & +20$\%$  & 0$\%$    & +20$\%$  & --     & --     & --     & --     & --     & --     & --    \\
S07 & +20$\%$ & +20$\%$  & -20$\%$  & --     & --     & --     & --     & --     & --     & --    \\
S08 & +20 $\%$  & +20$\%$  & +20$\%$  & --     & --     & --     & --     & --     & --     & --    \\
S09 & --     & --     & --     & --     & --     & --     & --     & --     & --     & --    \\
S10 & --     & --     & --     & --     & -50 $\%$  & -50 $\%$  & -50 $\%$  & -50 $\%$  & -50 $\%$  & --    \\
S11 & --     & --     & --     & --     & -20 $\%$  & -20 $\%$  & --     & -20 $\%$  & --     & --    \\
S12 & --     & --     & --     & --     & -20 $\%$  & -20 $\%$  & -20 $\%$  & --     & --     & --    \\
S13 & --     & --     & --     & --     & +20 $\%$  & +20 $\%$  & --     & +20 $\%$  & --     & --    \\
S14 & --     & --     & --     & --     & +20 $\%$  & +20 $\%$  & +20 $\%$  & --     & --     & --    \\
S15 & --     & --     & --     & --     & -30 $\%$  & -30 $\%$  & --     & -30 $\%$  & --     & --    \\
S16 & --     & --     & --     & --     & +70 $\%$  & +70 $\%$  & --     & +70 $\%$  & --     & --    \\
S17 & --     & --     & --     & -40 $\%$  & --     & --     & --     & -40 $\%$  & --     & --    \\
S18 & --     & --     & --     & +50 $\%$  & --     & --     & +50 $\%$  & +50 $\%$  & +50 $\%$  & +50 $\%$ \\
S19 & --     & --     & --     & +60 $\%$  & --     & --     & --     & +60 $\%$  & --     & --    \\
S20 & --     & --     & --     & +100 $\%$ & +100 $\%$ & +100 $\%$ & --     & --     & --     & --    \\
S21 & --     & --     & --     & --     & --     & --     & - 50 $\%$  & --     & --     & - 50 $\%$ \\
S22 & --     & --     & --     & --     & --     & --     & -20 $\%$  & -20 $\%$  & -20 $\%$  & -20 $\%$ \\
S23 & --     & --     & --     & --     & --     & --     & +20 $\%$  & +20 $\%$  & +20 $\%$  & + 20 $\%$ \\
S24 & --     & --     & --     & --     & --     & --     & +50 $\%$  & +50 $\%$  & +50 $\%$  & +50 $\%$ \\
S25 & --     & --     & --     & --     & --     & --     & +80 $\%$  & + 80 $\%$  & + 80 $\%$  & + 80 $\%$ \\
S26 & --     & --     & --     & --     & --     & --     & +80 $\%$  & --     & --     & +80 $\%$ \\
\bottomrule
\end{tabular}
\end{table*}
\end{landscape}


\subsection{Data and code availability}\label{app:code_available_statement}

The full implementation of the \rellm~framework, including scenario generation, feature construction, correlation and clustering analysis, surrogate modeling, and \llm-assisted interpretation, is openly available at \texttt{Github}:  
\url{https://github.com/Ali-Forootani/RELLM} and \texttt{Zenodo} \url{Ali-Forootani. (2025). Ali-Forootani/RELLM: RELLM (v1.0.0). Zenodo. https://doi.org/10.5281/zenodo.17311775}.  

The repository provides a reproducible pipeline that integrates \gams~outputs, structured feature engineering, surrogate modeling (\randf~, \dnn, \texttt{XGBoost}), SHAP-based interpretability, and natural-language querying. All scripts, documentation, and example workflows are included to enable replication and extension of the results.  

All \texttt{.gdx} input files, the \gams~optimization code (\texttt{.gms}), and the associated outputs (\texttt{.gdx}/\texttt{.csv}) used in this study are openly available at Zenodo:  
\href{https://doi.org/10.5281/zenodo.17077632}{10.5281/zenodo.17077632}.

\onecolumn

\subsection{\llm~Prompt Example Grounded in SHAP Values}
\label{app:llm_prompt_shap}

For transparency and reproducibility, we document the exact prompt template used to generate stakeholder-oriented narratives from SHAP analysis. The template is automatically constructed by inserting the top-$k$ SHAP-ranked features, model performance metrics, and scenario-level highlights. An example for the \texttt{capFMs} task is shown below.

\begin{lstlisting}[style=py, caption={Python prompt for stakeholder report}, label={lst:stakeholder_prompt}]
prompt = f"""
You are a sustainability analyst preparing a summary report for stakeholders, based on a machine learning ensemble model and SHAP analysis focused on forest management capacity (`capFMs`).

**Objective**: Predict and understand the key drivers of forest management capacity (`capFMs`)

**Model Performance**:  
• R² Score: {results_ensemble['r2_original']:.4f}  
• RMSE: {results_ensemble['rmse_original']:.2f} hectares

**Top 3 Influential Features (from SHAP analysis across ensemble models)**:  
1. **{top_feature_names[0]}** – SHAP = {top_shap_vals[0]:.3f}, Avg value = {top_feature_vals[0]:.3f}  
2. **{top_feature_names[1]}** – SHAP = {top_shap_vals[1]:.3f}, Avg value = {top_feature_vals[1]:.3f}  
3. **{top_feature_names[2]}** – SHAP = {top_shap_vals[2]:.3f}, Avg value = {top_feature_vals[2]:.3f}

**Regional & Policy Highlights**:  
• Region with highest capFMs potential: **{best_region}**  
• Leading growth technology: **{best_tech}**  

**Task**:  
Craft a clear and professional report that:
- Summarizes the ensemble model's performance in non-technical terms  
- Interprets how the top 3 features influence capFMs outcomes across the ensemble models  
- Highlights regional and technological opportunities  
- Recommends actions that align with long-term decarbonization goals  

The tone should be insight-driven, stakeholder-friendly, and suitable for regional planners, policymakers, and sustainability investors. Avoid equations or technical jargon—focus on actionable insights.
"""
\end{lstlisting}

\subsubsection{\llm~outputs in response to SHAP analysis}

This listing illustrates how SHAP-derived feature attributions were integrated into a structured natural-language prompt, ensuring that generated explanations remain grounded in interpretable model evidence. The following shows the \llm~response to the prompt \ref{lst:stakeholder_prompt}.

\begin{lstlisting}[style=py, caption={Report on Forest Management Capacity Ensemble Model Analysis}, label={lst:stakeholder_response}]
    Introduction:

This report summarizes the findings of an ensemble model analysis focused on forest management capacity (`capFMs`). The goal of the analysis was to predict and understand the key drivers of `capFMs` and identify opportunities for long-term decarbonization.

Model Performance:

The ensemble model achieved a high R² score of 0.9439, indicating a strong relationship between the input features and the outcome of `capFMs`. The RMSE of 10159.99 hectares indicates that the model is able to accurately predict the `capFMs` values.

Top 3 Influential Features:

The SHAP analysis identified the top 3 features that influence `capFMs` outcomes:

1. **year**: The year variable had a positive impact on `capFMs`, which suggests that as time progresses, `capFMs` is expected to increase. This could be due to advancements in technology and improvements in forest management practices.
2. **Technology\_FM02\_TSA**: The Technology\_FM02\_TSA variable had a positive impact on `capFMs`, indicating that the use of this technology could lead to improvements in forest management capacity.
3. **CostInvLevel\_Slope**: The CostInvLevel\_Slope variable had a negative impact on `capFMs`, suggesting that higher investment costs may hinder the development of forest management capacity.

Regional and Policy Highlights:

The analysis identified the region with the highest potential for `capFMs` development as DE1. This highlights the importance of investing in forest management capacity in this region.

The leading growth technology was identified as FM02\_TSA, which could be a valuable area of focus for long-term decarbonization goals.

Recommendations:

Based on the findings of the analysis, we recommend the following actions:

1. Invest in the region with the highest potential for `capFMs` development (DE1) to improve forest management capacity.
2. Focus on implementing the leading growth technology (FM02\_TSA) to improve forest management capacity and support long-term decarbonization goals.
3. Consider reducing investment costs to improve `capFMs` outcomes.

Conclusion:

The ensemble model analysis identified key drivers of forest management capacity and regional and technological opportunities for long-term decarbonization. The recommendations outlined in this report align with these findings and could help improve forest management capacity and support long-term decarbonization goals.
\end{lstlisting}

\subsection{\llm~outputs in response of stack holder's queries in subsection \ref{subsec:stack_holder_queries}} 

\begin{lstlisting}[style=py, caption={Report on Forest Management Capacity Ensemble Model Analysis}, label={lst:stakeholder_response_co2}]

Most similar scenarios: ('S06', 'S04') with correlation 1.0000
Least similar scenarios: ('S19', 'S17') with correlation 0.9997
Most similar scenarios: ('S09', 'S03') with correlation 1.0000
Least similar scenarios: ('S02', 'S01') with correlation 0.8805
Scenario Summary:
Matched parameter **CO2price** altered by **+20 %**.
Matched scenario(s): S06, S07, S08.
Cluster #1 -> contains 26 scenarios (average intra-cluster $\rho$ = 0.970).

OpenAI Response:
### Implications of a 20% Increase in CO2 Price

#### Summary of Change
The stakeholder has requested an analysis of what happens if the CO2 price increases by 20%. This change corresponds to the parameter `CO2price` and is represented in three scenarios: **S06**, **S07**, and **S08**. These scenarios are part of a cluster with a high intra-cluster correlation (0.970), indicating that the outcomes across these scenarios are highly similar.

---

### 1. **Overview of Affected Parameters in the Cluster**

The scenarios in this cluster (Cluster ID: 1) are defined with the following key parameters:
- **CO2price**: Increased by 20% (from base value of 1.0 to 1.2)
- **FMsgrowth**: Varies between 1.0 and 1.2
- **BeechArea0**: Varies between 0.8 and 1.2

This variation in FMsgrowth and BeechArea0 reflects potential changes in forest management and land use practices, which can amplify or moderate the impact of the increased CO2 price.

---

### 2. **Key Impacts of the CO2 Price Increase**

#### a. **Emissions (GHG)**

- **General Trend**: A higher CO2 price generally encourages a shift toward lower emissions, as it increases the cost of emitting CO2.
- **Cluster Implication**: In all three main scenarios (S06, S07, S08), the CO2 price is raised to 1.2, likely leading to a **reduction in greenhouse gas emissions**, particularly from agriculture and land use sectors.
- **Scenarios S06–S08** show slightly different values for **FMsgrowth** and **BeechArea0**, so emissions changes will vary slightly, but all follow the same general trend of **lower emissions** due to the higher carbon cost.

#### b. **Land Use**

- **Forest Expansion or Conservation**: The parameter **BeechArea0** indicates a potential expansion of beech forest areas. An increase in CO2 price may incentivize more forested land, which increases carbon sequestration.
- **Agricultural Land Use**: **FMsgrowth** reflects potential changes in forest management and land use intensity. Higher CO2 prices are likely to reduce intensive agricultural land use in favor of **more sustainable and carbon-friendly land use**.

#### c. **Costs**

- **Increased Cost of Emissions**: The higher CO2 price directly increases the cost of emissions in sectors such as agriculture and forestry.
- **Mitigation Investments**: Scenarios may reflect increased investments in carbon sequestration, afforestation, or reduced emissions practices, which in turn **increase short-term costs** but may reduce long-term environmental or regulatory risks.

#### d. **Agricultural and Forestry Management**

- **Shift in Agricultural Practices**: A higher CO2 price may lead to a **reduction in GHG-intensive agricultural practices**, such as peat extraction or intensive livestock production.
- **Forest Management**: The parameter **FMsgrowth** (forest management growth rate) impacts forest carbon storage; higher values (as in S07 and S08) may imply better forest management, leading to **higher carbon sequestration**.

---

### 3. **Most Closely Aligned Scenarios**

The following scenarios are most closely aligned with S06, S07, and S08 due to their shared parameters:

- **S03** and **S04** (CO2price_0.8) — Lower CO2 price, so less impact, but similar structure
- **S05** (CO2price_0.8, FMsgrowth_1.0, BeechArea0_1.0) — Similar structure but lower CO2 price
- **S06** (CO2price_1.2, FMsgrowth_1.0, BeechArea0_1.2) — Base case for this analysis
- **S07** (CO2price_1.2, FMsgrowth_1.2, BeechArea0_0.8) — Higher forest growth but lower beech area
- **S08** (CO2price_1.2, FMsgrowth_1.2, BeechArea0_1.2) — Highest forest growth and beech area

---

### 4. **Key Trends Across Cluster**

- **Reduction in emissions**: Due to higher costs of emissions.
- **Shift towards sustainable land use**: Expansion of forests (especially beech), and reduced GHG-intensive agriculture.
- **Increased cost of carbon-intensive activities**: Especially in agriculture and land use.
- **Enhanced forest management practices**: Especially in S07 and S08, where FMsgrowth is higher.

---

### 5. **Recommendations for Stakeholders**

- **Policy Makers**: Consider a CO2 price increase as a **cost-effective tool** to reduce emissions and promote sustainable land use.
- **Agricultural Sector**: Be prepared for **higher costs** associated with emissions, potentially necessitating a shift towards low-emission practices or carbon offsetting.
- **Forestry Sector**: Opportunities for **increased carbon sequestration** and forest management investments due to higher CO2 pricing.

---

### Conclusion

A 20% increase in CO2 price leads to **reduced emissions**, **shifts in land use toward forested and sustainable practices**, and **higher costs for carbon-intensive activities**. The most closely aligned scenarios (S06–S08) show strong consistency in outcomes, with **S08** being the most ambitious in terms of both forest growth and land use change. This cluster supports the conclusion that a moderate increase in CO2 price can effectively drive sustainability improvements across land use and emissions sectors.

    
\end{lstlisting}

\begin{lstlisting}[style=py, caption={Report on Forest Management Capacity Ensemble Model Analysis}, label={lst:stakeholder_response_costinvagri}]

Scenario Summary:
Matched parameter **costInvAgri** altered by **-20 %**.
Matched scenario(s): S11, S12.
Cluster #1 -> contains 26 scenarios (average intra-cluster $\rho$ = 0.970).

OpenAI Response:
### **Implications of a 20% Decrease in `costInvAgri`**

This scenario explores the consequences of a **20% reduction in the cost of investment in agriculture (`costInvAgri`)**. This change is modeled in **Scenarios S11 and S12**, which are part of a tightly clustered set of scenarios (Cluster ID 1, intra-cluster correlation = 0.970), indicating that they are highly similar in structure and behavior.

---

##  **Key Implications of Lower `costInvAgri`**

### 1. **Investment and Agricultural Expansion**
- A lower investment cost in agriculture makes it **more financially attractive** for farmers and agricultural investors to expand or modernize farming practices.
- This may lead to:
  - Increased agricultural land use (`Agriarea0`)
  - Greater adoption of high-yield farming techniques
  - Enhanced use of fertilizers or machinery (as captured by `Agrigrowth` and `ghgAgri`)

### 2. **Higher Agricultural Emissions (`ghgAgri`)**
- As agricultural investment becomes cheaper, there is likely an increase in agricultural activity (e.g., more land in cultivation, higher fertilizer use).
- This often leads to a **rise in agricultural greenhouse gas emissions (`ghgAgri`)**, especially in scenarios like **S12**, where `ghgAgri` is also reduced to 0.8 (but still increases from baseline due to increased activity).

### 3. **Shift in Land Use**
- With agricultural expansion, **land use changes** are expected in the agricultural sector.
- This may result in:
  - Increased pressure on natural ecosystems (e.g., peat extraction, forest conversion)
  - Changes in **BeechArea**, **PeatExtract**, and other land-use indicators (as seen in similar scenarios)

### 4. **Economic and Cost Implications**
- Reduced `costInvAgri` implies that **agricultural investment becomes more affordable**, which may:
  - Drive lower overall costs in agricultural systems
  - Influence the **cost of production** and profitability
  - Potentially reduce the **costMargAgri** (marginal cost of agriculture) if the shift is substantial

---

## **Trends Across the Cluster (Cluster ID 1)**

The cluster includes scenarios with **similar sensitivities to `costInvAgri`**, and trends across the scenarios show:

- **Positive feedback** between reduced `costInvAgri` and increased `Agrigrowth`, `Agriarea0`, and `ghgAgri`.
- Scenarios with **lower `costInvAgri`** tend to **increase land-use intensity** and **greenhouse gas emissions** from agriculture.
- Many scenarios in the cluster also adjust `costInvLevelAgri`, which likely reinforces these trends.

###  **Shared Scenario Patterns**
| Scenario | `costInvAgri` | `Agrigrowth` | `Agriarea0` | `ghgAgri` | Notes |
|----------|----------------|---------------|--------------|------------|-------|
| S11      | 0.8 (-20%)     | 0.8           | 0.8          | 0.8        | Balanced changes in agriculture and emissions |
| S12      | 0.8            | 0.8           | 0.8          | 0.8        | Similar to S11 |
| S13      | 1.2 (+20%)     | 1.2           | 1.2          | 1.2        | Opposite direction |
| S10      | 0.5            | 0.5           | 0.5          | 0.5        | Low cost, low investment |
| S14      | 1.2            | 1.2           | 1.2          | 1.2        | High cost, high investment |

This indicates a **strong linear or proportional relationship** between `costInvAgri` and agricultural system intensity.

---

## **Most Closely Aligned Scenarios**

The two main scenarios affected by a decrease in `costInvAgri` are:
- **S11**: `costInvAgri_0.8_costInvLevelAgri_0.8_Agrigrowth_0.8`
- **S12**: `costInvAgri_0.8_costInvLevelAgri_0.8_ghgAgri_0.8`

These are:
- Identical in `costInvAgri` and `costInvLevelAgri`
- Slightly different in `ghgAgri`, but both reflect the **same broad change in agricultural investment cost**

### Other similar scenarios in cluster:
- **S10** (`costInvAgri_0.5`) - even lower cost
- **S01**, **S02**, **S03**, **S04**, **S05** - also show similar agricultural cost and growth patterns
- **S22**, **S23**, **S24**, **S25**, **S26** - show increasing `ghgAgri` with increasing `costInvAgri`, suggesting a **positive correlation** between cost and emissions

---

## **Summary of Key Outcomes**

| Outcome | Impact |
|--------|--------|
| **Agricultural Investment** | Increased due to lower `costInvAgri` |
| **Land Use** | Expansion of `Agriarea0` likely |
| **Emissions** | Rise in `ghgAgri` due to higher agricultural activity |
| **Costs** | Lower marginal and investment costs in agriculture |
| **Correlation** | Strong positive correlation with `Agrigrowth` and `ghgAgri` across scenarios |
| **Aligned Scenarios** | **S11** and **S12** are most directly affected, with **S10** and **S22–S26** showing related trends |

---

## **Recommendation for Stakeholders**

- If `costInvAgri` decreases, expect **higher agricultural productivity**, **increased land use**, and **elevated emissions** unless offset by policy or technology changes.
- Monitor how changes in `costInvAgri` interact with **carbon pricing** or **peat extraction** policies, as these also influence emissions and land use.
- Consider aligning agricultural support policies with **sustainability goals** to mitigate unintended environmental consequences of low investment costs.

Let me know if you'd like a visual interpretation (e.g., chart of emissions vs. cost changes) or a comparison with the baseline scenario (S09).

\end{lstlisting}

\bibliographystyle{elsarticle-num-names}
\bibliography{refs}

\begin{thebibliography}{85}
\expandafter\ifx\csname natexlab\endcsname\relax\def\natexlab#1{#1}\fi
\providecommand{\url}[1]{\texttt{#1}}
\providecommand{\href}[2]{#2}
\providecommand{\path}[1]{#1}
\providecommand{\DOIprefix}{doi:}
\providecommand{\ArXivprefix}{arXiv:}
\providecommand{\URLprefix}{URL: }
\providecommand{\Pubmedprefix}{pmid:}
\providecommand{\doi}[1]{\href{http://dx.doi.org/#1}{\path{#1}}}
\providecommand{\Pubmed}[1]{\href{pmid:#1}{\path{#1}}}
\providecommand{\bibinfo}[2]{#2}
\ifx\xfnm\relax \def\xfnm[#1]{\unskip,\space#1}\fi
\bibitem[{Forootani et~al.(2021)Forootani, Tipaldi, Iervolino, and
  Dey}]{forootani2021enhanced}
\bibinfo{author}{A.~Forootani}, \bibinfo{author}{M.~Tipaldi},
  \bibinfo{author}{R.~Iervolino}, \bibinfo{author}{S.~Dey},
\newblock \bibinfo{title}{Enhanced exploration least-squares methods for
  optimal stopping problems},
\newblock \bibinfo{journal}{IEEE Control Systems Letters} \bibinfo{volume}{6}
  (\bibinfo{year}{2021}) \bibinfo{pages}{271--276}.
\bibitem[{Howells et~al.(2011)Howells, Rogner, Strachan, Heaps, Huntington,
  Kypreos, Hughes, Silveira, DeCarolis, Bazillian et~al.}]{howells2011osemosys}
\bibinfo{author}{M.~Howells}, \bibinfo{author}{H.~Rogner},
  \bibinfo{author}{N.~Strachan}, \bibinfo{author}{C.~Heaps},
  \bibinfo{author}{H.~Huntington}, \bibinfo{author}{S.~Kypreos},
  \bibinfo{author}{A.~Hughes}, \bibinfo{author}{S.~Silveira},
  \bibinfo{author}{J.~DeCarolis}, \bibinfo{author}{M.~Bazillian}, et~al.,
\newblock \bibinfo{title}{{OSeMOSYS}: the open source energy modeling system:
  an introduction to its ethos, structure and development},
\newblock \bibinfo{journal}{Energy Policy} \bibinfo{volume}{39}
  (\bibinfo{year}{2011}) \bibinfo{pages}{5850--5870}.
\bibitem[{Bhattacharyya(2019)}]{bhattacharyya2011energy}
\bibinfo{author}{S.~C. Bhattacharyya}, \bibinfo{title}{Energy economics:
  concepts, issues, markets and governance}, \bibinfo{publisher}{Springer
  Nature}, \bibinfo{year}{2019}.
\bibitem[{Laveneziana et~al.(2023)Laveneziana, Prussi, and
  Chiaramonti}]{bloxham2020review}
\bibinfo{author}{L.~Laveneziana}, \bibinfo{author}{M.~Prussi},
  \bibinfo{author}{D.~Chiaramonti},
\newblock \bibinfo{title}{Critical review of energy planning models for the
  sustainable development at company level},
\newblock \bibinfo{journal}{Energy Strategy Reviews} \bibinfo{volume}{49}
  (\bibinfo{year}{2023}) \bibinfo{pages}{101136}.
\bibitem[{Dagoumas and Koltsaklis(2019)}]{dagoumas2019review}
\bibinfo{author}{A.~S. Dagoumas}, \bibinfo{author}{N.~E. Koltsaklis},
\newblock \bibinfo{title}{Review of models for integrating renewable energy in
  the generation expansion planning},
\newblock \bibinfo{journal}{Applied Energy} \bibinfo{volume}{242}
  (\bibinfo{year}{2019}) \bibinfo{pages}{1573--1587}.
\bibitem[{Connolly et~al.(2010)Connolly, Lund, Mathiesen, and
  Leahy}]{connolly2010review}
\bibinfo{author}{D.~Connolly}, \bibinfo{author}{H.~Lund},
  \bibinfo{author}{B.~V. Mathiesen}, \bibinfo{author}{M.~Leahy},
\newblock \bibinfo{title}{A review of computer tools for analysing the
  integration of renewable energy into various energy systems},
\newblock \bibinfo{journal}{Applied energy} \bibinfo{volume}{87}
  (\bibinfo{year}{2010}) \bibinfo{pages}{1059--1082}.
\bibitem[{Ringkj{\o}b et~al.(2018)Ringkj{\o}b, Haugan, and
  Solbrekke}]{ringkjøb2018review}
\bibinfo{author}{H.-K. Ringkj{\o}b}, \bibinfo{author}{P.~M. Haugan},
  \bibinfo{author}{I.~M. Solbrekke},
\newblock \bibinfo{title}{A review of modelling tools for energy and
  electricity systems with large shares of variable renewables},
\newblock \bibinfo{journal}{Renewable and Sustainable Energy Reviews}
  \bibinfo{volume}{96} (\bibinfo{year}{2018}) \bibinfo{pages}{440--459}.
\bibitem[{Loulou et~al.(2005)Loulou, Remme, Kanudia, Lehtila, and
  Goldstein}]{loulou2016times}
\bibinfo{author}{R.~Loulou}, \bibinfo{author}{U.~Remme},
  \bibinfo{author}{A.~Kanudia}, \bibinfo{author}{A.~Lehtila},
  \bibinfo{author}{G.~Goldstein}, \bibinfo{title}{Documentation for the times
  model part {II}}, \bibinfo{type}{Technical Report}, \bibinfo{year}{2005}.
\bibitem[{Huppmann et~al.(2019)Huppmann, Fricko, Kishimoto, Krey, Riahi
  et~al.}]{huppmann2019messageix}
\bibinfo{author}{D.~Huppmann}, \bibinfo{author}{O.~Fricko},
  \bibinfo{author}{P.~N. Kishimoto}, \bibinfo{author}{V.~Krey},
  \bibinfo{author}{K.~Riahi}, et~al.,
\newblock \bibinfo{title}{The {MESSAGE\lowercase{ix}} integrated assessment
  model and the ix modeling platform (ixmp): An open framework for integrated
  and cross‑cutting analysis of energy, climate, the environment, and
  sustainable development},
\newblock \bibinfo{journal}{Environmental Modelling \& Software}
  \bibinfo{volume}{112} (\bibinfo{year}{2019}) \bibinfo{pages}{143--156}.
  \DOIprefix\doi{10.1016/j.envsoft.2018.11.012}.
\bibitem[{L{\"o}ffler et~al.(2017)L{\"o}ffler, Hainsch, Burandt, Oei, Kemfert,
  and von Hirschhausen}]{loffler2017genesysmod}
\bibinfo{author}{K.~L{\"o}ffler}, \bibinfo{author}{K.~Hainsch},
  \bibinfo{author}{T.~Burandt}, \bibinfo{author}{P.~Oei},
  \bibinfo{author}{C.~Kemfert}, \bibinfo{author}{C.~von Hirschhausen},
\newblock \bibinfo{title}{Designing a model for the global energy
  system—{GENeSYS‑MOD}: An application of the open‑source energy modeling
  system ({OSeMOSYS})},
\newblock \bibinfo{journal}{Energies} \bibinfo{volume}{10}
  (\bibinfo{year}{2017}) \bibinfo{pages}{1468}.
  \DOIprefix\doi{10.3390/en10101468}.
\bibitem[{Barani et~al.(2025)Barani, L{\"o}ffler, del Granado, Moskalenko,
  Panos, Hoffart, von Hirschhausen, Kannavou, Auer, Hainsch
  et~al.}]{barani2025european}
\bibinfo{author}{M.~Barani}, \bibinfo{author}{K.~L{\"o}ffler},
  \bibinfo{author}{P.~C. del Granado}, \bibinfo{author}{N.~Moskalenko},
  \bibinfo{author}{E.~Panos}, \bibinfo{author}{F.~M. Hoffart},
  \bibinfo{author}{C.~von Hirschhausen}, \bibinfo{author}{M.~Kannavou},
  \bibinfo{author}{H.~Auer}, \bibinfo{author}{K.~Hainsch}, et~al.,
\newblock \bibinfo{title}{{European Energy Vision 2060}: Charting diverse
  pathways for {Europe}'s energy transition},
\newblock \bibinfo{journal}{arXiv preprint arXiv:2501.12993}
  (\bibinfo{year}{2025}).
\bibitem[{Brown et~al.(2017)Brown, H{\"o}rsch, and
  Schlachtberger}]{brown_pypsa_2018}
\bibinfo{author}{T.~Brown}, \bibinfo{author}{J.~H{\"o}rsch},
  \bibinfo{author}{D.~Schlachtberger},
\newblock \bibinfo{title}{Pypsa: Python for power system analysis},
\newblock \bibinfo{journal}{arXiv preprint arXiv:1707.09913}
  (\bibinfo{year}{2017}).
\bibitem[{Millinger et~al.(2022)Millinger, Tafarte, Jordan, Musonda, Chan,
  Meisel, and Aliabadi}]{millinger2022model}
\bibinfo{author}{M.~Millinger}, \bibinfo{author}{P.~Tafarte},
  \bibinfo{author}{M.~Jordan}, \bibinfo{author}{F.~Musonda},
  \bibinfo{author}{K.~Chan}, \bibinfo{author}{K.~Meisel},
  \bibinfo{author}{D.~E. Aliabadi},
\newblock \bibinfo{title}{A model for cost-and greenhouse gas optimal material
  and energy allocation of biomass and hydrogen},
\newblock \bibinfo{journal}{SoftwareX} \bibinfo{volume}{20}
  (\bibinfo{year}{2022}) \bibinfo{pages}{101264}.
\bibitem[{Pfenninger and Pickering(2018)}]{pfenninger2018calliope}
\bibinfo{author}{S.~Pfenninger}, \bibinfo{author}{B.~Pickering},
\newblock \bibinfo{title}{Calliope: a multi-scale energy systems modelling
  framework},
\newblock \bibinfo{journal}{Journal of Open Source Software}
  \bibinfo{volume}{3} (\bibinfo{year}{2018}) \bibinfo{pages}{825}.
\bibitem[{Vaidyanathan(2021)}]{vaidyanathan2021integrated}
\bibinfo{author}{G.~Vaidyanathan},
\newblock \bibinfo{title}{Integrated assessment climate policy models have
  proven useful, with caveats},
\newblock \bibinfo{journal}{Proceedings of the National Academy of Sciences}
  \bibinfo{volume}{118} (\bibinfo{year}{2021}) \bibinfo{pages}{e2101899118}.
\bibitem[{Cointe(2024)}]{iiasa2022ar6}
\bibinfo{author}{B.~Cointe},
\newblock \bibinfo{title}{The ar6 scenario explorer and the history of ipcc
  scenarios databases: evolutions and challenges for transparency, pluralism
  and policy-relevance},
\newblock \bibinfo{journal}{npj Climate Action} \bibinfo{volume}{3}
  (\bibinfo{year}{2024}) \bibinfo{pages}{3}.
\bibitem[{Huppmann et~al.(2019)Huppmann, Kriegler, Krey, Riahi, Rogelj, Calvin,
  Humpenoeder, Popp, Rose, Weyant et~al.}]{huppmann2019iamc15c}
\bibinfo{author}{D.~Huppmann}, \bibinfo{author}{E.~Kriegler},
  \bibinfo{author}{V.~Krey}, \bibinfo{author}{K.~Riahi},
  \bibinfo{author}{J.~Rogelj}, \bibinfo{author}{K.~Calvin},
  \bibinfo{author}{F.~Humpenoeder}, \bibinfo{author}{A.~Popp},
  \bibinfo{author}{S.~K. Rose}, \bibinfo{author}{J.~Weyant}, et~al.,
\newblock \bibinfo{title}{Iamc 1.5 c scenario explorer and data hosted by
  iiasa},
\newblock \bibinfo{journal}{(No Title)}  (\bibinfo{year}{2019}).
\bibitem[{Rooney-Varga et~al.(2020)Rooney-Varga, Kapmeier, Sterman, Jones,
  Putko, and Rath}]{rooneyvarga2020climateaction}
\bibinfo{author}{J.~N. Rooney-Varga}, \bibinfo{author}{F.~Kapmeier},
  \bibinfo{author}{J.~D. Sterman}, \bibinfo{author}{A.~P. Jones},
  \bibinfo{author}{M.~Putko}, \bibinfo{author}{K.~Rath},
\newblock \bibinfo{title}{The climate action simulation},
\newblock \bibinfo{journal}{Simulation \& Gaming} \bibinfo{volume}{51}
  (\bibinfo{year}{2020}) \bibinfo{pages}{114--140}.
\bibitem[{van Ruijven and Min(2020)}]{van2020messageix}
\bibinfo{author}{B.~van Ruijven}, \bibinfo{author}{J.~Min},
\newblock \bibinfo{title}{The messageix-globiom model and scenarios for
  transition risk analysis}  (\bibinfo{year}{2020}).
\bibitem[{Rogelj et~al.(2018)Rogelj, Shindell, Jiang, Fifita, Forster,
  Ginzburg, Handa, Kheshgi, Kobayashi, Kriegler et~al.}]{rogelj2018mitigation}
\bibinfo{author}{J.~Rogelj}, \bibinfo{author}{D.~Shindell},
  \bibinfo{author}{K.~Jiang}, \bibinfo{author}{S.~Fifita},
  \bibinfo{author}{P.~Forster}, \bibinfo{author}{V.~Ginzburg},
  \bibinfo{author}{C.~Handa}, \bibinfo{author}{H.~Kheshgi},
  \bibinfo{author}{S.~Kobayashi}, \bibinfo{author}{E.~Kriegler}, et~al.,
\newblock \bibinfo{title}{Mitigation pathways compatible with 1.5 c in the
  context of sustainable development},
\newblock in: \bibinfo{booktitle}{Global warming of 1.5 C},
  \bibinfo{publisher}{Intergovernmental Panel on Climate Change},
  \bibinfo{year}{2018}, pp. \bibinfo{pages}{93--174}.
\bibitem[{Klein et~al.(2014)Klein, Luderer, Kriegler, Strefler, Bauer,
  Leimbach, Popp, Dietrich, Humpen{\"o}der, Lotze-Campen
  et~al.}]{klein2014value}
\bibinfo{author}{D.~Klein}, \bibinfo{author}{G.~Luderer},
  \bibinfo{author}{E.~Kriegler}, \bibinfo{author}{J.~Strefler},
  \bibinfo{author}{N.~Bauer}, \bibinfo{author}{M.~Leimbach},
  \bibinfo{author}{A.~Popp}, \bibinfo{author}{J.~P. Dietrich},
  \bibinfo{author}{F.~Humpen{\"o}der}, \bibinfo{author}{H.~Lotze-Campen},
  et~al.,
\newblock \bibinfo{title}{The value of bioenergy in low stabilization
  scenarios: an assessment using remind-magpie},
\newblock \bibinfo{journal}{Climatic change} \bibinfo{volume}{123}
  (\bibinfo{year}{2014}) \bibinfo{pages}{705--718}.
\bibitem[{Loulou et~al.(2005)Loulou, Remme, Kanudia, Lehtila, and
  Goldstein}]{loulou2005documentation}
\bibinfo{author}{R.~Loulou}, \bibinfo{author}{U.~Remme},
  \bibinfo{author}{A.~Kanudia}, \bibinfo{author}{A.~Lehtila},
  \bibinfo{author}{G.~Goldstein},
\newblock \bibinfo{title}{Documentation for the times model part ii},
\newblock \bibinfo{journal}{Energy technology systems analysis programme}
  \bibinfo{volume}{384} (\bibinfo{year}{2005}).
\bibitem[{McJeon et~al.(2014)McJeon, Edmonds, Bauer, Clarke, Fisher, Flannery,
  Hilaire, Krey, Marangoni, Mi et~al.}]{mcjeon2014limited}
\bibinfo{author}{H.~McJeon}, \bibinfo{author}{J.~Edmonds},
  \bibinfo{author}{N.~Bauer}, \bibinfo{author}{L.~Clarke},
  \bibinfo{author}{B.~Fisher}, \bibinfo{author}{B.~P. Flannery},
  \bibinfo{author}{J.~Hilaire}, \bibinfo{author}{V.~Krey},
  \bibinfo{author}{G.~Marangoni}, \bibinfo{author}{R.~Mi}, et~al.,
\newblock \bibinfo{title}{Limited impact on decadal-scale climate change from
  increased use of natural gas},
\newblock \bibinfo{journal}{Nature} \bibinfo{volume}{514}
  (\bibinfo{year}{2014}) \bibinfo{pages}{482--485}.
\bibitem[{Short et~al.(2011)Short, Sullivan, Mai, Mowers, Uriarte, Blair,
  Heimiller, and Martinez}]{short2011regional}
\bibinfo{author}{W.~Short}, \bibinfo{author}{P.~Sullivan},
  \bibinfo{author}{T.~Mai}, \bibinfo{author}{M.~Mowers},
  \bibinfo{author}{C.~Uriarte}, \bibinfo{author}{N.~Blair},
  \bibinfo{author}{D.~Heimiller}, \bibinfo{author}{A.~Martinez},
  \bibinfo{title}{Regional energy deployment system (ReEDS)},
  \bibinfo{type}{Technical Report}, National Renewable Energy Lab.(NREL),
  Golden, CO (United States), \bibinfo{year}{2011}.
\bibitem[{L{\"o}ffler et~al.(2017)L{\"o}ffler, Hainsch, Burandt, Oei, Kemfert,
  and Von~Hirschhausen}]{loffler2017designing}
\bibinfo{author}{K.~L{\"o}ffler}, \bibinfo{author}{K.~Hainsch},
  \bibinfo{author}{T.~Burandt}, \bibinfo{author}{P.-Y. Oei},
  \bibinfo{author}{C.~Kemfert}, \bibinfo{author}{C.~Von~Hirschhausen},
\newblock \bibinfo{title}{Designing a model for the global energy
  system—genesys-mod: an application of the open-source energy modeling
  system (osemosys)},
\newblock \bibinfo{journal}{Energies} \bibinfo{volume}{10}
  (\bibinfo{year}{2017}) \bibinfo{pages}{1468}.
\bibitem[{Brown et~al.(2022)Brown, Hampp, Hofmann, H{\"o}rsch, Maria, Neumann
  et~al.}]{brown2022pypsa}
\bibinfo{author}{T.~Brown}, \bibinfo{author}{J.~Hampp},
  \bibinfo{author}{F.~Hofmann}, \bibinfo{author}{J.~H{\"o}rsch},
  \bibinfo{author}{M.~Maria}, \bibinfo{author}{F.~Neumann}, et~al.,
\newblock \bibinfo{title}{Pypsa-eur-sec: A sector-coupled open optimisation
  model of the european energy system}  (\bibinfo{year}{2022}).
\bibitem[{Brown et~al.(2018)Brown, Schlachtberger, Kies, Schramm, and
  Greiner}]{brown2018synergies}
\bibinfo{author}{T.~Brown}, \bibinfo{author}{D.~Schlachtberger},
  \bibinfo{author}{A.~Kies}, \bibinfo{author}{S.~Schramm},
  \bibinfo{author}{M.~Greiner},
\newblock \bibinfo{title}{Synergies of sector coupling and transmission
  reinforcement in a cost-optimised, highly renewable european energy system},
\newblock \bibinfo{journal}{Energy} \bibinfo{volume}{160}
  (\bibinfo{year}{2018}) \bibinfo{pages}{720--739}.
\bibitem[{G{\o}tske et~al.(2024)G{\o}tske, Andresen, Neumann, and
  Victoria}]{gotske2024designing}
\bibinfo{author}{E.~K. G{\o}tske}, \bibinfo{author}{G.~B. Andresen},
  \bibinfo{author}{F.~Neumann}, \bibinfo{author}{M.~Victoria},
\newblock \bibinfo{title}{Designing a sector-coupled european energy system
  robust to 60 years of historical weather data},
\newblock \bibinfo{journal}{Nature Communications} \bibinfo{volume}{15}
  (\bibinfo{year}{2024}) \bibinfo{pages}{1--12}.
\bibitem[{Unnewehr et~al.(2022)Unnewehr, Sch{\"a}fer, and
  Weidlich}]{unnewehr2022value}
\bibinfo{author}{J.~F. Unnewehr}, \bibinfo{author}{M.~Sch{\"a}fer},
  \bibinfo{author}{A.~Weidlich},
\newblock \bibinfo{title}{The value of network resolution--a validation study
  of the european energy system model pypsa-eur},
\newblock in: \bibinfo{booktitle}{2022 Open Source Modelling and Simulation of
  Energy Systems (OSMSES)}, \bibinfo{organization}{IEEE}, \bibinfo{year}{2022},
  pp. \bibinfo{pages}{1--7}.
\bibitem[{Glaum et~al.(2024)Glaum, Neumann, and Brown}]{glaum2024offshore}
\bibinfo{author}{P.~Glaum}, \bibinfo{author}{F.~Neumann},
  \bibinfo{author}{T.~Brown},
\newblock \bibinfo{title}{Offshore power and hydrogen networks for europe’s
  north sea},
\newblock \bibinfo{journal}{Applied Energy} \bibinfo{volume}{369}
  (\bibinfo{year}{2024}) \bibinfo{pages}{123530}.
\bibitem[{Conejo and Wu(2022)}]{conejo2022robust}
\bibinfo{author}{A.~J. Conejo}, \bibinfo{author}{X.~Wu},
\newblock \bibinfo{title}{Robust optimization in power systems: a tutorial
  overview},
\newblock \bibinfo{journal}{Optimization and Engineering} \bibinfo{volume}{23}
  (\bibinfo{year}{2022}) \bibinfo{pages}{2051--2073}.
\bibitem[{Qiu et~al.(2016)Qiu, Xu, Wang, Dvorkin, and
  Kirschen}]{qiu2016stochastic}
\bibinfo{author}{T.~Qiu}, \bibinfo{author}{B.~Xu}, \bibinfo{author}{Y.~Wang},
  \bibinfo{author}{Y.~Dvorkin}, \bibinfo{author}{D.~S. Kirschen},
\newblock \bibinfo{title}{Stochastic multistage coplanning of transmission
  expansion and energy storage},
\newblock \bibinfo{journal}{IEEE Transactions on Power Systems}
  \bibinfo{volume}{32} (\bibinfo{year}{2016}) \bibinfo{pages}{643--651}.
\bibitem[{Lorca and Sun(2016)}]{lorca2016multistage}
\bibinfo{author}{A.~Lorca}, \bibinfo{author}{X.~A. Sun},
\newblock \bibinfo{title}{Multistage robust unit commitment with dynamic
  uncertainty sets and energy storage},
\newblock \bibinfo{journal}{IEEE Transactions on Power Systems}
  \bibinfo{volume}{32} (\bibinfo{year}{2016}) \bibinfo{pages}{1678--1688}.
\bibitem[{Moret et~al.(????)Moret, Babonneau, Bierlaire, and
  Mar{\'e}chal}]{moretdecision}
\bibinfo{author}{S.~Moret}, \bibinfo{author}{F.~Babonneau},
  \bibinfo{author}{M.~Bierlaire}, \bibinfo{author}{F.~Mar{\'e}chal},
  \bibinfo{title}{Decision support for strategic energy planning: a complete
  robust optimization framework}, ????
\bibitem[{Li and Grossmann(2021)}]{li2021review}
\bibinfo{author}{C.~Li}, \bibinfo{author}{I.~E. Grossmann},
\newblock \bibinfo{title}{A review of stochastic programming methods for
  optimization of process systems under uncertainty},
\newblock \bibinfo{journal}{Frontiers in Chemical Engineering}
  \bibinfo{volume}{2} (\bibinfo{year}{2021}) \bibinfo{pages}{622241}.
\bibitem[{Kayac{\i}k et~al.(2025)Kayac{\i}k, Basciftci, Schrotenboer, and
  Ursavas}]{kayacik2025partially}
\bibinfo{author}{S.~E. Kayac{\i}k}, \bibinfo{author}{B.~Basciftci},
  \bibinfo{author}{A.~H. Schrotenboer}, \bibinfo{author}{E.~Ursavas},
\newblock \bibinfo{title}{Partially adaptive multistage stochastic
  programming},
\newblock \bibinfo{journal}{European Journal of Operational Research}
  \bibinfo{volume}{321} (\bibinfo{year}{2025}) \bibinfo{pages}{192--207}.
\bibitem[{Hilpert et~al.(2018)Hilpert, Kaldemeyer, Krien, G{\"u}nther,
  Wingenbach, and Plessmann}]{hilpert2018open}
\bibinfo{author}{S.~Hilpert}, \bibinfo{author}{C.~Kaldemeyer},
  \bibinfo{author}{U.~Krien}, \bibinfo{author}{S.~G{\"u}nther},
  \bibinfo{author}{C.~Wingenbach}, \bibinfo{author}{G.~Plessmann},
\newblock \bibinfo{title}{The open energy modelling framework (oemof)-a new
  approach to facilitate open science in energy system modelling},
\newblock \bibinfo{journal}{Energy strategy reviews} \bibinfo{volume}{22}
  (\bibinfo{year}{2018}) \bibinfo{pages}{16--25}.
\bibitem[{Prina et~al.(2022)Prina, Nastasi, Groppi, Misconel, Garcia, and
  Sparber}]{prina2022comparison}
\bibinfo{author}{M.~G. Prina}, \bibinfo{author}{B.~Nastasi},
  \bibinfo{author}{D.~Groppi}, \bibinfo{author}{S.~Misconel},
  \bibinfo{author}{D.~A. Garcia}, \bibinfo{author}{W.~Sparber},
\newblock \bibinfo{title}{Comparison methods of energy system frameworks,
  models and scenario results},
\newblock \bibinfo{journal}{Renewable and Sustainable Energy Reviews}
  \bibinfo{volume}{167} (\bibinfo{year}{2022}) \bibinfo{pages}{112719}.
\bibitem[{Pfenninger et~al.(2018)Pfenninger, Hirth, Schlecht, Schmid, Wiese,
  Brown, Davis, Gidden, Heinrichs, Heuberger et~al.}]{pfenninger2018opening}
\bibinfo{author}{S.~Pfenninger}, \bibinfo{author}{L.~Hirth},
  \bibinfo{author}{I.~Schlecht}, \bibinfo{author}{E.~Schmid},
  \bibinfo{author}{F.~Wiese}, \bibinfo{author}{T.~Brown},
  \bibinfo{author}{C.~Davis}, \bibinfo{author}{M.~Gidden},
  \bibinfo{author}{H.~Heinrichs}, \bibinfo{author}{C.~Heuberger}, et~al.,
\newblock \bibinfo{title}{Opening the black box of energy modelling: Strategies
  and lessons learned},
\newblock \bibinfo{journal}{Energy Strategy Reviews} \bibinfo{volume}{19}
  (\bibinfo{year}{2018}) \bibinfo{pages}{63--71}.
\bibitem[{G{\"o}pfert et~al.(2025)G{\"o}pfert, Kuckertz, Weinand, and
  Stolten}]{gopfert2025wiki}
\bibinfo{author}{J.~G{\"o}pfert}, \bibinfo{author}{P.~Kuckertz},
  \bibinfo{author}{J.~M. Weinand}, \bibinfo{author}{D.~Stolten},
\newblock \bibinfo{title}{{Wiki-Quantities and Wiki-Measurements}: Datasets of
  quantities and their measurement context from {Wikipedia}},
\newblock \bibinfo{journal}{Scientific Data} \bibinfo{volume}{12}
  (\bibinfo{year}{2025}) \bibinfo{pages}{1277}.
\bibitem[{Pfenninger et~al.(2014)Pfenninger, Hawkes, and
  Keirstead}]{pfenninger2017energy}
\bibinfo{author}{S.~Pfenninger}, \bibinfo{author}{A.~Hawkes},
  \bibinfo{author}{J.~Keirstead},
\newblock \bibinfo{title}{Energy systems modeling for twenty-first century
  energy challenges},
\newblock \bibinfo{journal}{Renewable and sustainable energy reviews}
  \bibinfo{volume}{33} (\bibinfo{year}{2014}) \bibinfo{pages}{74--86}.
\bibitem[{Ramsebner et~al.(2021)Ramsebner, Haas, Ajanovic, and
  Wietschel}]{ueckerdt2019potential}
\bibinfo{author}{J.~Ramsebner}, \bibinfo{author}{R.~Haas},
  \bibinfo{author}{A.~Ajanovic}, \bibinfo{author}{M.~Wietschel},
\newblock \bibinfo{title}{The sector coupling concept: A critical review},
\newblock \bibinfo{journal}{Wiley interdisciplinary reviews: energy and
  environment} \bibinfo{volume}{10} (\bibinfo{year}{2021})
  \bibinfo{pages}{e396}.
\bibitem[{Yao et~al.(2023)Yao, Lum, Johnston, Mejia-Mendoza, Zhou, Wen,
  Aspuru-Guzik, Sargent, and Seh}]{yao2023machine}
\bibinfo{author}{Z.~Yao}, \bibinfo{author}{Y.~Lum},
  \bibinfo{author}{A.~Johnston}, \bibinfo{author}{L.~M. Mejia-Mendoza},
  \bibinfo{author}{X.~Zhou}, \bibinfo{author}{Y.~Wen},
  \bibinfo{author}{A.~Aspuru-Guzik}, \bibinfo{author}{E.~H. Sargent},
  \bibinfo{author}{Z.~W. Seh},
\newblock \bibinfo{title}{Machine learning for a sustainable energy future},
\newblock \bibinfo{journal}{Nature Reviews Materials} \bibinfo{volume}{8}
  (\bibinfo{year}{2023}) \bibinfo{pages}{202--215}.
\bibitem[{Grillone et~al.(2020)Grillone, Danov, Sumper, Cipriano, and
  Mor}]{teichgraeber2020methodology}
\bibinfo{author}{B.~Grillone}, \bibinfo{author}{S.~Danov},
  \bibinfo{author}{A.~Sumper}, \bibinfo{author}{J.~Cipriano},
  \bibinfo{author}{G.~Mor},
\newblock \bibinfo{title}{A review of deterministic and data-driven methods to
  quantify energy efficiency savings and to predict retrofitting scenarios in
  buildings},
\newblock \bibinfo{journal}{Renewable and Sustainable Energy Reviews}
  \bibinfo{volume}{131} (\bibinfo{year}{2020}) \bibinfo{pages}{110027}.
\bibitem[{Balduin et~al.(2020)Balduin, Westermann, and
  Puiutta}]{balduin2020lvgrid}
\bibinfo{author}{S.~Balduin}, \bibinfo{author}{T.~Westermann},
  \bibinfo{author}{E.~Puiutta},
\newblock \bibinfo{title}{Evaluating different machine learning techniques as
  surrogate for low voltage grids},
\newblock \bibinfo{journal}{Energy Informatics} \bibinfo{volume}{3}
  (\bibinfo{year}{2020}) \bibinfo{pages}{24}.
\bibitem[{Menear et~al.(2023)Menear, Duplyakin, Oliver, Shah, Martin, Martinek,
  Nithyanandam, and Ma}]{menear2023tes}
\bibinfo{author}{K.~Menear}, \bibinfo{author}{D.~Duplyakin},
  \bibinfo{author}{M.~C. Oliver}, \bibinfo{author}{M.~Shah},
  \bibinfo{author}{M.~J. Martin}, \bibinfo{author}{J.~Martinek},
  \bibinfo{author}{K.~Nithyanandam}, \bibinfo{author}{Z.~Ma},
  \bibinfo{title}{One System, Many Models: Designing a Surrogate Model for
  Sulfur Thermal Energy Storage}, \bibinfo{type}{Technical Report}, National
  Renewable Energy Laboratory (NREL), Golden, CO (United States),
  \bibinfo{year}{2023}.
\bibitem[{Herbinger et~al.(2023)Herbinger, Vandenhof, and
  Kummert}]{herbinger2023surrogate}
\bibinfo{author}{F.~Herbinger}, \bibinfo{author}{C.~Vandenhof},
  \bibinfo{author}{M.~Kummert},
\newblock \bibinfo{title}{Building energy model calibration using a surrogate
  neural network},
\newblock \bibinfo{journal}{Energy and buildings} \bibinfo{volume}{289}
  (\bibinfo{year}{2023}) \bibinfo{pages}{113057}.
\bibitem[{Bamdad et~al.(2020)Bamdad, Cholette, and Bell}]{bamdad2020soas}
\bibinfo{author}{K.~Bamdad}, \bibinfo{author}{M.~E. Cholette},
  \bibinfo{author}{J.~M. Bell},
\newblock \bibinfo{title}{Building energy optimization using surrogate model
  and active sampling},
\newblock \bibinfo{journal}{Journal of Building Performance Simulation}
  \bibinfo{volume}{13} (\bibinfo{year}{2020}) \bibinfo{pages}{760--776}.
\bibitem[{Ghafariasl et~al.(2024)Ghafariasl, Mahmoudan, Mohammadi, Nazarparvar,
  Hoseinzadeh, Fathali, Chang, Zeinalnezhad, and Garcia}]{ghafariasl2024mgs}
\bibinfo{author}{P.~Ghafariasl}, \bibinfo{author}{A.~Mahmoudan},
  \bibinfo{author}{M.~Mohammadi}, \bibinfo{author}{A.~Nazarparvar},
  \bibinfo{author}{S.~Hoseinzadeh}, \bibinfo{author}{M.~Fathali},
  \bibinfo{author}{S.~Chang}, \bibinfo{author}{M.~Zeinalnezhad},
  \bibinfo{author}{D.~A. Garcia},
\newblock \bibinfo{title}{Neural network-based surrogate modeling and
  optimization of a multigeneration system},
\newblock \bibinfo{journal}{Applied Energy} \bibinfo{volume}{364}
  (\bibinfo{year}{2024}) \bibinfo{pages}{123130}.
\bibitem[{Prina et~al.(2024)Prina, Dallapiccola, Moser, and
  Sparber}]{prina2024eplan}
\bibinfo{author}{M.~G. Prina}, \bibinfo{author}{M.~Dallapiccola},
  \bibinfo{author}{D.~Moser}, \bibinfo{author}{W.~Sparber},
\newblock \bibinfo{title}{Machine learning as a surrogate model for energyplan:
  Speeding up energy system optimization at the country level},
\newblock \bibinfo{journal}{Energy} \bibinfo{volume}{307}
  (\bibinfo{year}{2024}) \bibinfo{pages}{132735}.
\bibitem[{Bogensperger et~al.(2022)Bogensperger, Fabel, and
  Ferstl}]{bogensperger2022emulation}
\bibinfo{author}{A.~J. Bogensperger}, \bibinfo{author}{Y.~Fabel},
  \bibinfo{author}{J.~Ferstl},
\newblock \bibinfo{title}{Accelerating energy-economic simulation models via
  machine learning-based emulation and time series aggregation},
\newblock \bibinfo{journal}{Energies} \bibinfo{volume}{15}
  (\bibinfo{year}{2022}) \bibinfo{pages}{1239}.
\bibitem[{Mohammadi et~al.(2024)Mohammadi, Bui, Su, and
  Wang}]{mohammadi2024opf}
\bibinfo{author}{S.~Mohammadi}, \bibinfo{author}{V.-H. Bui},
  \bibinfo{author}{W.~Su}, \bibinfo{author}{B.~Wang},
\newblock \bibinfo{title}{Surrogate modeling for solving opf: A review},
\newblock \bibinfo{journal}{Sustainability} \bibinfo{volume}{16}
  (\bibinfo{year}{2024}) \bibinfo{pages}{9851}.
\bibitem[{Keramati and Hamdullahpur(2023)}]{keramati2023thermal}
\bibinfo{author}{H.~Keramati}, \bibinfo{author}{F.~Hamdullahpur},
\newblock \bibinfo{title}{Deep convolutional surrogates and freedom in thermal
  design},
\newblock \bibinfo{journal}{Energy and AI} \bibinfo{volume}{13}
  (\bibinfo{year}{2023}) \bibinfo{pages}{100248}.
\bibitem[{Weber et~al.(2020)Weber, Corotan, Hutchinson, Kravitz, and
  Link}]{weber2020precip}
\bibinfo{author}{T.~Weber}, \bibinfo{author}{A.~Corotan},
  \bibinfo{author}{B.~Hutchinson}, \bibinfo{author}{B.~Kravitz},
  \bibinfo{author}{R.~Link},
\newblock \bibinfo{title}{Deep learning for creating surrogate models of
  precipitation in earth system models},
\newblock \bibinfo{journal}{Atmospheric Chemistry and Physics}
  \bibinfo{volume}{20} (\bibinfo{year}{2020}) \bibinfo{pages}{2303--2317}.
\bibitem[{Wasserkrug et~al.(2025)Wasserkrug, Boussioux, Den~Hertog, Mirzazadeh,
  Birbil, Kurtz, and Maragno}]{wasserkrug2025enhancing}
\bibinfo{author}{S.~Wasserkrug}, \bibinfo{author}{L.~Boussioux},
  \bibinfo{author}{D.~Den~Hertog}, \bibinfo{author}{F.~Mirzazadeh},
  \bibinfo{author}{{\c{S}}.~I. Birbil}, \bibinfo{author}{J.~Kurtz},
  \bibinfo{author}{D.~Maragno},
\newblock \bibinfo{title}{Enhancing decision making through the integration of
  large language models and operations research optimization},
\newblock in: \bibinfo{booktitle}{Proceedings of the AAAI Conference on
  Artificial Intelligence}, volume~\bibinfo{volume}{39}, \bibinfo{year}{2025},
  pp. \bibinfo{pages}{28643--28650}.
\bibitem[{Buster(2023)}]{buster2023elm}
\bibinfo{author}{G.~Buster}, \bibinfo{title}{Large language models (llms) for
  energy systems research}, \bibinfo{year}{2023}.
\bibitem[{Weers et~al.(2024)Weers, Podgorny, Taverna, Anderson, Porse, and
  Buster}]{weers2025askgdr}
\bibinfo{author}{J.~Weers}, \bibinfo{author}{S.~Podgorny},
  \bibinfo{author}{N.~Taverna}, \bibinfo{author}{A.~Anderson},
  \bibinfo{author}{S.~Porse}, \bibinfo{author}{G.~Buster},
  \bibinfo{title}{Empowering Geothermal Research: The Geothermal Data
  Repository's New AI Research Assistant}, \bibinfo{type}{Technical Report},
  National Renewable Energy Laboratory (NREL), Golden, CO (United States),
  \bibinfo{year}{2024}.
\bibitem[{Buster et~al.(2024)Buster, Pinchuk, Barrons, McKeever, Levine, and
  Lopez}]{buster2024siting}
\bibinfo{author}{G.~Buster}, \bibinfo{author}{P.~Pinchuk},
  \bibinfo{author}{J.~Barrons}, \bibinfo{author}{R.~McKeever},
  \bibinfo{author}{A.~Levine}, \bibinfo{author}{A.~Lopez},
\newblock \bibinfo{title}{Supporting energy policy research with large language
  models: A case study in wind energy siting ordinances},
\newblock \bibinfo{journal}{Energy and AI} \bibinfo{volume}{18}
  (\bibinfo{year}{2024}) \bibinfo{pages}{100431}.
\bibitem[{Emami et~al.(2023)Emami, Li, and Grout}]{emami2023inventory}
\bibinfo{author}{P.~Emami}, \bibinfo{author}{Z.~Li},
  \bibinfo{author}{R.~Grout}, \bibinfo{title}{Language Model Inventory v1. 0
  [Slides]}, \bibinfo{type}{Technical Report}, National Renewable Energy
  Laboratory (NREL), Golden, CO (United States), \bibinfo{year}{2023}.
\bibitem[{Foucquier et~al.(2013)Foucquier, Robert, Suard, St{\'e}phan, and
  Jay}]{lombardi2020state}
\bibinfo{author}{A.~Foucquier}, \bibinfo{author}{S.~Robert},
  \bibinfo{author}{F.~Suard}, \bibinfo{author}{L.~St{\'e}phan},
  \bibinfo{author}{A.~Jay},
\newblock \bibinfo{title}{State of the art in building modelling and energy
  performances prediction: A review},
\newblock \bibinfo{journal}{Renewable and Sustainable Energy Reviews}
  \bibinfo{volume}{23} (\bibinfo{year}{2013}) \bibinfo{pages}{272--288}.
\bibitem[{Bevis et~al.(2017)Bevis, Schaab, Rautenbach, and
  Coetzee}]{pfenninger2014visualisation}
\bibinfo{author}{Y.~Bevis}, \bibinfo{author}{G.~Schaab},
  \bibinfo{author}{V.~Rautenbach}, \bibinfo{author}{S.~Coetzee},
\newblock \bibinfo{title}{Expert opinions on using the third dimension to
  visualise wind speed uncertainty in wind farm planning},
\newblock \bibinfo{journal}{International Journal of Cartography}
  \bibinfo{volume}{3} (\bibinfo{year}{2017}) \bibinfo{pages}{61--75}.
\bibitem[{Fontes et~al.(2018)Fontes, Freires et~al.}]{zeyringer2018visualising}
\bibinfo{author}{C.~H. d.~O. Fontes}, \bibinfo{author}{F.~G.~M. Freires},
  et~al.,
\newblock \bibinfo{title}{Sustainable and renewable energy supply chain: A
  system dynamics overview},
\newblock \bibinfo{journal}{Renewable and Sustainable Energy Reviews}
  \bibinfo{volume}{82} (\bibinfo{year}{2018}) \bibinfo{pages}{247--259}.
\bibitem[{Bouckaert et~al.(2021)Bouckaert, Pales, McGlade, Remme, Wanner,
  Varro, D'Ambrosio, and Spencer}]{IEA2021netzero}
\bibinfo{author}{S.~Bouckaert}, \bibinfo{author}{A.~F. Pales},
  \bibinfo{author}{C.~McGlade}, \bibinfo{author}{U.~Remme},
  \bibinfo{author}{B.~Wanner}, \bibinfo{author}{L.~Varro},
  \bibinfo{author}{D.~D'Ambrosio}, \bibinfo{author}{T.~Spencer},
  \bibinfo{title}{Net zero by 2050: A roadmap for the global energy sector},
  \bibinfo{year}{2021}.
\bibitem[{L’Heureux et~al.(2022)L’Heureux, Grolinger, and
  Capretz}]{zhang2023transformerenergy}
\bibinfo{author}{A.~L’Heureux}, \bibinfo{author}{K.~Grolinger},
  \bibinfo{author}{M.~A. Capretz},
\newblock \bibinfo{title}{Transformer-based model for electrical load
  forecasting},
\newblock \bibinfo{journal}{Energies} \bibinfo{volume}{15}
  (\bibinfo{year}{2022}) \bibinfo{pages}{4993}.
\bibitem[{Floudas and Lin(2005)}]{floudas2005mixed}
\bibinfo{author}{C.~A. Floudas}, \bibinfo{author}{X.~Lin},
\newblock \bibinfo{title}{Mixed integer linear programming in process
  scheduling: Modeling, algorithms, and applications},
\newblock \bibinfo{journal}{Annals of Operations Research}
  \bibinfo{volume}{139} (\bibinfo{year}{2005}) \bibinfo{pages}{131--162}.
\bibitem[{Forootani et~al.(2020)Forootani, Tipaldi, Zarch, Liuzza, and
  Glielmo}]{forootani2020least}
\bibinfo{author}{A.~Forootani}, \bibinfo{author}{M.~Tipaldi},
  \bibinfo{author}{M.~G. Zarch}, \bibinfo{author}{D.~Liuzza},
  \bibinfo{author}{L.~Glielmo},
\newblock \bibinfo{title}{A least-squares temporal difference based method for
  solving resource allocation problems},
\newblock \bibinfo{journal}{IFAC Journal of Systems and Control}
  \bibinfo{volume}{13} (\bibinfo{year}{2020}) \bibinfo{pages}{100106}.
\bibitem[{Sobol(2001)}]{sobol2001global}
\bibinfo{author}{I.~M. Sobol},
\newblock \bibinfo{title}{Global sensitivity indices for nonlinear mathematical
  models and their monte carlo estimates},
\newblock \bibinfo{journal}{Mathematics and computers in simulation}
  \bibinfo{volume}{55} (\bibinfo{year}{2001}) \bibinfo{pages}{271--280}.
\bibitem[{McKay et~al.(2000)McKay, Beckman, and Conover}]{mckay2000comparison}
\bibinfo{author}{M.~D. McKay}, \bibinfo{author}{R.~J. Beckman},
  \bibinfo{author}{W.~J. Conover},
\newblock \bibinfo{title}{A comparison of three methods for selecting values of
  input variables in the analysis of output from a computer code},
\newblock \bibinfo{journal}{Technometrics} \bibinfo{volume}{42}
  (\bibinfo{year}{2000}) \bibinfo{pages}{55--61}.
\bibitem[{Jain et~al.(2000)Jain, Duin, and Mao}]{jain2000statistical}
\bibinfo{author}{A.~K. Jain}, \bibinfo{author}{R.~P.~W. Duin},
  \bibinfo{author}{J.~Mao},
\newblock \bibinfo{title}{Statistical pattern recognition: A review},
\newblock \bibinfo{journal}{IEEE Transactions on pattern analysis and machine
  intelligence} \bibinfo{volume}{22} (\bibinfo{year}{2000})
  \bibinfo{pages}{4--37}.
\bibitem[{Pearson(1895)}]{pearson1895vii}
\bibinfo{author}{K.~Pearson},
\newblock \bibinfo{title}{Vii. note on regression and inheritance in the case
  of two parents},
\newblock \bibinfo{journal}{proceedings of the royal society of London}
  \bibinfo{volume}{58} (\bibinfo{year}{1895}) \bibinfo{pages}{240--242}.
\bibitem[{Lundberg and Lee(2017)}]{lundberg2017unified}
\bibinfo{author}{S.~M. Lundberg}, \bibinfo{author}{S.-I. Lee},
\newblock \bibinfo{title}{A unified approach to interpreting model
  predictions},
\newblock \bibinfo{journal}{Advances in neural information processing systems}
  \bibinfo{volume}{30} (\bibinfo{year}{2017}).
\bibitem[{Breiman(2001)}]{breiman2001random}
\bibinfo{author}{L.~Breiman},
\newblock \bibinfo{title}{Random forests},
\newblock \bibinfo{journal}{Machine learning} \bibinfo{volume}{45}
  (\bibinfo{year}{2001}) \bibinfo{pages}{5--32}.
\bibitem[{Pedregosa et~al.(2011)Pedregosa, Varoquaux, Gramfort, Michel,
  Thirion, Grisel, Blondel, Prettenhofer, Weiss, Dubourg
  et~al.}]{pedregosa2011scikit}
\bibinfo{author}{F.~Pedregosa}, \bibinfo{author}{G.~Varoquaux},
  \bibinfo{author}{A.~Gramfort}, \bibinfo{author}{V.~Michel},
  \bibinfo{author}{B.~Thirion}, \bibinfo{author}{O.~Grisel},
  \bibinfo{author}{M.~Blondel}, \bibinfo{author}{P.~Prettenhofer},
  \bibinfo{author}{R.~Weiss}, \bibinfo{author}{V.~Dubourg}, et~al.,
\newblock \bibinfo{title}{Scikit-learn: Machine learning in python},
\newblock \bibinfo{journal}{the Journal of machine Learning research}
  \bibinfo{volume}{12} (\bibinfo{year}{2011}) \bibinfo{pages}{2825--2830}.
\bibitem[{ets(2016)}]{etsap:times}
\bibinfo{title}{The integrated markal–efom system (times) model},
  \bibinfo{howpublished}{IEA‑ETSAP}, \bibinfo{year}{2016}.
  \bibinfo{note}{TIMES is a bottom‑up, technology‑rich model generator
  using LP over medium‑long horizons}.
\bibitem[{wik(2025)}]{wikipedia:times}
\bibinfo{title}{Times (energy model)}, \bibinfo{howpublished}{Wikipedia},
  \bibinfo{year}{2025}. \bibinfo{note}{TIMES succeeded MARKAL in 2008; used in
  177 institutions worldwide}.
\bibitem[{mes(2025)}]{messageix:docs}
\bibinfo{title}{Messageix documentation}, \bibinfo{howpublished}{MESSAGE‑ix
  project (IIASA)}, \bibinfo{year}{2025}. \bibinfo{note}{Open‑source
  Python/GAMS framework for energy–economy–environment modelling}.
\bibitem[{mes(2021)}]{messageix:openmod}
\bibinfo{title}{The messageix modelling framework},
  \bibinfo{howpublished}{OpenMODE Initiative / workshop}, \bibinfo{year}{2021}.
  \bibinfo{note}{All contents are open‑source and designed for integrated
  assessment energy systems}.
\bibitem[{Brown et~al.(2017)Brown, H{\"o}rsch, and
  Schlachtberger}]{brown2017pypsa}
\bibinfo{author}{T.~Brown}, \bibinfo{author}{J.~H{\"o}rsch},
  \bibinfo{author}{D.~Schlachtberger},
\newblock \bibinfo{title}{{PyPSA}: Python for power system analysis},
\newblock \bibinfo{journal}{arXiv preprint arXiv:1707.09913}
  (\bibinfo{year}{2017}).
\bibitem[{pyp(2025)}]{pypsa:wikipedia}
\bibinfo{title}{Pypsa}, \bibinfo{howpublished}{Wikipedia},
  \bibinfo{year}{2025}. \bibinfo{note}{Python library for power systems,
  supports electricity, heating, transport, hydrogen; MILP framework}.
\bibitem[{Hopkins et~al.(2018)Hopkins, Sun, DeWitt, and
  Knappe}]{hopkins2018recently}
\bibinfo{author}{Z.~R. Hopkins}, \bibinfo{author}{M.~Sun},
  \bibinfo{author}{J.~C. DeWitt}, \bibinfo{author}{D.~R. Knappe},
\newblock \bibinfo{title}{Recently detected drinking water contaminants: Genx
  and other per-and polyfluoroalkyl ether acids},
\newblock \bibinfo{journal}{Journal-American Water Works Association}
  \bibinfo{volume}{110} (\bibinfo{year}{2018}) \bibinfo{pages}{13--28}.
\bibitem[{Ho et~al.(2021)Ho, Becker, Brown, Brown, Chernyakhovskiy, Cohen,
  Cole, Corcoran, Eurek, Frazier et~al.}]{ho2021regional}
\bibinfo{author}{J.~Ho}, \bibinfo{author}{J.~Becker},
  \bibinfo{author}{M.~Brown}, \bibinfo{author}{P.~Brown},
  \bibinfo{author}{I.~Chernyakhovskiy}, \bibinfo{author}{S.~Cohen},
  \bibinfo{author}{W.~Cole}, \bibinfo{author}{S.~Corcoran},
  \bibinfo{author}{K.~Eurek}, \bibinfo{author}{W.~Frazier}, et~al.,
  \bibinfo{title}{Regional energy deployment system (reeds) model documentation
  (version 2020)}, \bibinfo{type}{Technical Report}, National Renewable Energy
  Lab.(NREL), Golden, CO (United States), \bibinfo{year}{2021}.
\bibitem[{Lund et~al.(2021)Lund, Thellufsen, {\O}stergaard, Sorkn{\ae}s, Skov,
  and Mathiesen}]{lund2021energyplan}
\bibinfo{author}{H.~Lund}, \bibinfo{author}{J.~Z. Thellufsen},
  \bibinfo{author}{P.~A. {\O}stergaard}, \bibinfo{author}{P.~Sorkn{\ae}s},
  \bibinfo{author}{I.~R. Skov}, \bibinfo{author}{B.~V. Mathiesen},
\newblock \bibinfo{title}{{EnergyPLAN}--advanced analysis of smart energy
  systems},
\newblock \bibinfo{journal}{Smart Energy} \bibinfo{volume}{1}
  (\bibinfo{year}{2021}) \bibinfo{pages}{100007}.
\bibitem[{Calvin et~al.(2019)Calvin, Patel, Clarke, Asrar, Bond-Lamberty, Cui,
  Di~Vittorio, Dorheim, Edmonds, Hartin et~al.}]{calvin2019gcam}
\bibinfo{author}{K.~Calvin}, \bibinfo{author}{P.~Patel},
  \bibinfo{author}{L.~Clarke}, \bibinfo{author}{G.~Asrar},
  \bibinfo{author}{B.~Bond-Lamberty}, \bibinfo{author}{R.~Y. Cui},
  \bibinfo{author}{A.~Di~Vittorio}, \bibinfo{author}{K.~Dorheim},
  \bibinfo{author}{J.~Edmonds}, \bibinfo{author}{C.~Hartin}, et~al.,
\newblock \bibinfo{title}{{GCAM} v5. 1: representing the linkages between
  energy, water, land, climate, and economic systems},
\newblock \bibinfo{journal}{Geoscientific Model Development}
  \bibinfo{volume}{12} (\bibinfo{year}{2019}) \bibinfo{pages}{677--698}.
\bibitem[{{Energy Exemplar LLC}(2025)}]{plexos:energyexemplar}
\bibinfo{author}{{Energy Exemplar LLC}}, \bibinfo{title}{Plexos: Energy market
  simulation software},
  \bibinfo{howpublished}{\url{https://www.energyexemplar.com/plexos}},
  \bibinfo{year}{2025}. \bibinfo{note}{Commercial, supports sub‑hourly
  LP/MIP, zonal \& nodal analysis across power, gas, water, and renewables}.
\bibitem[{Dorfner(2023)}]{github:urbs}
\bibinfo{author}{J.~Dorfner}, \bibinfo{title}{urbs: A linear optimisation model
  for distributed energy systems},
  \bibinfo{howpublished}{\url{https://github.com/tum-ens/urbs}},
  \bibinfo{year}{2023}. \bibinfo{note}{Accessed: 2025-10-10, release v1.0.1,
  GPL-3.0 license}.

\end{thebibliography}

\vfill

\end{document}